\pdfoutput=1 

\documentclass[11pt]{article}

\usepackage[]{acl}

\usepackage{times}
\usepackage{latexsym}

\usepackage[T1]{fontenc}

\usepackage[utf8]{inputenc}

\usepackage{microtype}

\usepackage{amsmath}
\usepackage[linesnumbered,ruled,algo2e]{algorithm2e}
\usepackage{todonotes}
\usepackage{enumerate}
\usepackage{amssymb}
\usepackage{amsmath}
\usepackage{latexsym}
\usepackage{url}
\usepackage{mathtools}
\usepackage{multirow}
\usepackage{xspace}
\usepackage{comment}
\usepackage{flowchart}
\usetikzlibrary{arrows}
\usepackage{booktabs}
\usepackage{subcaption}
\usepackage{hyperref}
\usepackage{enumitem}
\usepackage{multirow}
\usepackage{tikz}
\usepackage{pgfplots}
\usepackage{ifthen}

\usepackage{float}
\usepackage{multicol}
\restylefloat{table}
\usepackage{amsmath}
\usepackage{graphicx}
\usepackage{color}
 
\usepackage{comment}
\usepackage{enumerate}
\usepackage{enumitem}

\usepackage{float}
\restylefloat{table}
\usepackage{placeins}

\usepackage{algorithm}
\usepackage{algpseudocode}
\usepackage{enumerate}

\usepackage{xcolor}
\usepackage{soul}
\definecolor{hlcolor}{rgb}{0.92,1,0.91}
\sethlcolor{hlcolor}
 
\DeclareMathAlphabet{\pazocal}{OMS}{zplm}{m}{n}
\DeclareMathAlphabet{\pazocal}{OMS}{zplm}{m}{n}

\usepackage{cleveref}

\crefformat{section}{\S#2#1#3} 
\crefformat{subsection}{\S#2#1#3}
\crefformat{subsubsection}{\S#2#1#3}

\newcommand{\red}[1]{\textcolor{red}{#1}}

\newcommand{\ie}{{\em i.e.,}\xspace}

\newcommand{\Na}{({\em a})~}
\newcommand{\Nb}{({\em b})~}
\newcommand{\Nc}{({\em c})~}
\newcommand{\Nd}{({\em d})~}

\usepackage{enumitem}

\title{Data Selection Curriculum for Neural Machine Translation}

\author{
Tasnim Mohiuddin$^{1,2}$\thanks{\ \ Work done while Tasnim was interning at Meta AI}\hspace{3mm} 
Philipp Koehn$^2$\hspace{3mm} 
Vishrav Chaudhary$^2$\\
{\bf James Cross}$^2$\hspace{3mm}
{\bf Shruti Bhosale}$^2$\hspace{3mm}
{\bf Shafiq Joty}$^1$\\ 
Nanyang Technological University$^1$ \hspace{1em} Meta  AI$^2$\\
\texttt{\{mohi0004,srjoty\}@ntu.edu.sg}\\
\texttt{\{pkoehn,vishrav,jcross,shru\}@fb.com}
}

\begin{document}
\maketitle
\begin{abstract}
Neural Machine Translation (NMT) models are typically trained on heterogeneous data that are concatenated and randomly shuffled. {However, not all of the training data are equally useful to the model.} Curriculum training aims to present the data to the NMT models in a meaningful order. In this work, we introduce a two-stage curriculum training framework for NMT where we fine-tune a base NMT model on subsets of data, selected by both deterministic scoring using pre-trained methods and online scoring that considers prediction scores of the emerging NMT model. Through comprehensive experiments on six language pairs comprising low- and high-resource languages from WMT'21, we have shown that our curriculum strategies consistently demonstrate better quality (up to +2.2 BLEU improvement) and faster convergence (approximately 50\% fewer updates).

\end{abstract}

\section{Introduction}
\label{sec:intro}

The notion of a curriculum came from the human learning experience; we learn better {and faster} when the learnable examples are presented in a meaningful sequence rather than a random order \cite{NEWPORT199011}. In the case of machine learning, curriculum training hypothesizes presenting the data samples in a meaningful order to machine learners during training such that it imposes structure in the task of learning \cite{bengio-ccl}.

In recent years, Neural Machine Translation (NMT) has shown impressive performance in high-resource settings \cite{Hassan2018AchievingHP,popel2020transforming}. {Typically, training data of the NMT systems are a heterogeneous collection from different domains, sources, topics, styles, and modalities. The quality of the training data also varies a lot, so as their linguistic difficulty levels. The usual practice of training NMT systems is to concatenate all available data into a single pool and randomly sample training examples. However, not all of them may be useful, some examples may be redundant, and some data might even be noisy and detrimental to the final NMT system performance \cite{khayrallah-koehn-2018-impact}.} 
So, NMT systems have the potential to benefit greatly from curriculum training in terms of both speed and quality.

{There have been several attempts to extend the success of curriculum training to NMT \cite{zhang2018empirical, platanios-etal-2019-competence}. To our knowledge,  \citet{kocmi-bojar-2017-curriculum} were the first to explore the impact of several curriculum heuristics on training an NMT system, in their case Czech-English. They ensure that samples within a mini-batch have similar linguistic properties, and order mini-batches based on some heuristics like sentence length and vocabulary frequency -- which improves the translation quality.
Another successful line of research in NMT is domain-specific fine-tuning \cite{luong-manning-2015-stanford}, where NMT models are first trained on a large general-domain data and then fine-tuned on small in-domain data.}

In this work, we propose a \textit{two-stage} curriculum training framework for NMT --- \textit{model warm-up} and \textit{model fine-tuning}, {where we apply the data-selection curriculum in the later stage.} We initially train a base NMT model in the warm-up stage on all available data. In the fine-tuning stage, we adapt the base model on {selected subsets of the data. The subset selection is performed by considering data quality and/or usefulness at the current state of the model.} We explore two sets of data-selection curriculum strategies --- {\textit{deterministic} and \textit{online}}. The deterministic curriculum uses external measures which require pretrained models for selecting the data subset at the beginning of the model fine-tuning stage and continues training on the selected subset. In contrast, the online curriculum dynamically selects a subset of the data for each epoch without requiring any external measure. Specifically, it leverages the prediction scores of the emerging NMT model which are the by-product of the training.

For picking the data subset in the online curriculum, we investigate two approaches of \textit{data-selection window}  --- static and dynamic. Even though the size of the data-selection window is \textit{constant} throughout the training in the static approach, the samples in the selected subset \textit{vary} from epoch-to-epoch due to the change in their prediction scores. In contrast, we \textit{change} the data-selection window size in the dynamic approach by either expanding or shrinking.

Comprehensive experiments on six language pairs (12 translation directions) comprising low- and high-resource languages from WMT'21 \cite{akhbardeh-etal-2021-findings}
reveal that our curriculum strategies consistently 
demonstrate better performance compared to the baseline trained on all the data (up to +2.2 BLEU). We observe bigger gains in the high-resource pairs compared to the low-resource ones. Interestingly, we find that the online curriculum approaches perform on par with the deterministic approaches while not using any external pretrained models. Our proposed curriculum training approaches not only exhibit better performance but also converge much faster requiring approximately 50\% fewer updates. 







\section{Proposed Framework}

Let $s$ and $t$ denote the source and target language respectively, and $\mathcal{D}_g = \{(x_i, y_i)\}_{i=1}^N$ denote the general-domain parallel training data containing $N$ sentence pairs with $x_i$ and $y_i$ coming from $s$ and $t$ languages, respectively. Also, let $\mathcal{D}_{d} \subseteq \mathcal{D}_g$ be the in-domain parallel training data  and  $\mathcal{M}$ is an NMT model that can translate sentences from $s$ to $t$. The overall training objective of the NMT model is to minimize the total loss of the training data:

\begin{equation}
\small
    \mathcal{J}(\theta) = \sum_{i=1}^N \mathcal{L}(x_i, y_i, \theta) = \sum_{i=1}^N -\log P_{\theta}(y_i|x_i)
\end{equation}
\normalsize

\noindent where $P_{\theta}(y_i|x_i)$ is the sentence-level  translation probability of the target sentence $y_i$ for the source sentence $x_i$ with $\theta$ being the parameters of $\mathcal{M}$.

We propose a \textit{two-stage} training curriculum where in  \textit{model warm-up} stage we train $\mathcal{M}$ on general domain parallel data $\mathcal{D}_g$ for $K$ number of gradient updates; $K$ is generally smaller than the total number of updates $\mathcal{M}$ requires for convergence. Then in \textit{model fine-tuning} stage, we adapt $\mathcal{M}$ on selected subsets of in-domain parallel data $\mathcal{D}_d$. Based on the intuition: \textit{``not all of the training data are useful or non-redundant, some samples might be irrelevant or even detrimental to the model''}, we hypothesize that there exists a $\mathcal{D}_s \subset \mathcal{D}_d$, fine-tuning on which $\mathcal{M}$ will exhibit improved performance. 

Our goal is to design a {ranking} of the training samples which will eventually help us to extract $\mathcal{D}_s$ from $\mathcal{D}_d$. For this, we investigate two sets of data-selection curriculum strategies --- {\textit{deterministic} and \textit{online}}. Both strategies require a measure of data quality and/or usefulness at the current state of the model to extract $\mathcal{D}_s$. While the deterministic curriculum uses external measures that require pretrained models, the online curriculum leverages the prediction scores of the emerging NMT models.

\subsection{{Deterministic Curriculum}}
\label{subsec:deterministic-curriculum}
In this strategy, we select a $\mathcal{D}_s  \subset \mathcal{D}_d$ initially and do not change it during the \textit{model fine-tuning} stage. We first score each parallel sentence pair $(x_i, y_i) \in \mathcal{D}_d$ using an {external bitext scoring method}. We experiment with three scoring methods as described below.

\paragraph{$\bullet$ LASER} This approach utilizes the Language-Agnostic SEntence Representations ({LASER}) toolkit \cite{artetxe-schwenk-2019}, which gives multilingual sentence representations using an encoder-decoder architecture trained on a parallel corpus. We use the sentence representations to \textit{score the similarity} of a parallel sentence pair using the Cross-Domain Similarity Local Scaling (CSLS) measure, which performs better {than other similarity metrics} in reducing the hubness problem \cite{conneau2017word}.
\begin{equation}
\small
    \mathcal{S}{core}_{\text{laser}}(x_i, y_i) = \text{CSLS}(\texttt{LASER}(x_i), \texttt{LASER}(y_i))
\end{equation}
\citet{chaudhary-etal-2019-low} showed benefits of {LASER}-based ranking for low-resource corpus filtering.

\paragraph{$\bullet$ Dual Conditional Cross-Entropy (DCCE)} \citet{junczys-dowmunt-2018-dual} proposed this method, which requires two inverse translation models -- one forward model ($f$) and one backward model ($b$), trained on the same parallel corpus. It then finds the score of a sentence pair $(x_i, y_i)$ by taking the maximal symmetric agreement of the two models which exploits the conditional cross-entropy ($H$).
\begin{equation}
\small
\begin{split}
    \hspace{-1em} \mathcal{S}core_{\text{dcce}}(x_i, y_i) = |H_{f} - H_{b}| + \frac{1}{2}(H_{f} + H_{b}) \\
     \hspace{-0.8em} \text{where~}  H_{f} = -\log P_{\theta_f}(y_i|x_i); \hspace{0.1em}  H_{b} = -\log P_{\theta_b}(x_i|y_i) \label{eq:dcce} 
\end{split}
\end{equation}
The absolute difference between the conditional cross-entropy in Eq. \ref{eq:dcce} measures the agreement between the two conditional probability distributions. If the sentences in a bitext are equally probable (good) or equally improbable (bad/noisy), this part of the equation will have a low score. To differentiate between these two scenarios, we need the average cross-entropy score which scores higher for improbable sentence pairs.

\begin{algorithm}[t]
    \SetKwInOut{Input}{Input}
    \SetKwInOut{Output}{Output}
    \SetAlgoLined
    \SetNoFillComment
    \LinesNotNumbered 
    \SetNlSkip{0em}
    \SetKwRepeat{Do}{do}{while}

    \Input{General domain corpus $\mathcal{D}_g$, in-domain corpus $\mathcal{D}_{d} \subseteq \mathcal{D}_g$, {external pretrained bitext scorer $\mathcal{S}$}} 
    \Output{A trained translation model }
    
    1. \tcp{\hl{model warm-up stage}} 
    \hspace{.5em}\underline{Train} a \textit{base model} $\mathcal{M}$ on general domain corpus $\mathcal{D}_g$ for $K$ number of updates \\
    \vspace{-1.em}
    
    2. \tcp{\hl{model fine-tuning stage}}  
    \hspace{.5em}\Na \textbf{Score} each $(x_i, y_i) \in \mathcal{D}_{d}$ using $\mathcal{S}$  \\
    \Nb \textbf{Rank} $(x_i, y_i) \in \mathcal{D}_{d}$ based on these scores \\
    \Nc \textbf{Find} $\mathcal{D}_s \subset \mathcal{D}_d$ by selecting top $p\%$ of $\mathcal{D}_{d}$ \\
    \Nd  \For{n\_epochs}{ 
           \hspace{2em}  \underline{Fine-tune} $\mathcal{M}$  on $\mathcal{D}_s$
      }
    \caption{Deterministic Curriculum Strategy}
    \label{alg:det}
\end{algorithm}

\paragraph{$\bullet$ Modified Moore-Lewis (MML)} 
MML ranks the sentence pairs based on domain relevance by calculating \textit{cross-entropy difference} scores \cite{moore-lewis-2010-intelligent, axelrod-etal-2011-domain}. For this, we need to train four language models (LM): \textbf{in}- and \textbf{gen}eral-domain LMs in both \textbf{s}ource and \textbf{t}arget languages. 
Then we find the MML score of a parallel sentence pair $(x_i, y_i)$ as follows:
\vspace{-1.em}

\begin{equation}
\small
\begin{split}
    \hspace{-1em} \mathcal{S}core_{\text{mml}}(x_i, y_i) &=  (H_{s,in}(x_i) - H_{s,gen}(x_i)) \\
    & + (H_{t,in}(y_i) - H_{t,gen}(y_i))  \\
    \text{where~} H_{b,C} (z) &= -\log P^{\text{LM}}_{b,C}(z) \label{eq:ced}
\end{split}
\end{equation}
\normalsize

\noindent Here, $b \in \{s,t\}$ refers to the bitext side and $C \in \{in,gen\}$ refers to the corpus domain. In our experiments, we use the \textit{newscrawl} data as \textbf{in}-domain and \textit{commoncrawl} data combined with newscrawl as \textbf{gen}eral-domain for training the LMs.

After scoring each parallel sentence pair $(x_i, y_i) \in \mathcal{D}_d$ by any of the above methods, we rank $\mathcal{D}_d$ based on the scores. We then pick the better subset $\mathcal{D}_s$ by selecting top $p\%$ pairs from the ranked $\mathcal{D}_d$. Finally, we fine-tune the base model $\mathcal{M}$ on $\mathcal{D}_s$. LASER and DCCE performs \textit{denoising} curriculum {(\ie\ higher rank for good translation and lower rank for noisy ones)} while MML performs \textit{domain similarity} curriculum on the given data.  Algorithm \ref{alg:det} presents a pseudo-code of the deterministic curriculum strategy.

\subsection{{Online} Curriculum}
\label{subsec:online}

\begin{algorithm}[t]
    \SetKwInOut{Input}{Input}
    \SetKwInOut{Output}{Output}
    \SetAlgoLined
    \SetNoFillComment
    \LinesNotNumbered 
    \SetNlSkip{0em}
    \SetKwRepeat{Do}{do}{while}

    \Input{General corpus $\mathcal{D}_g$, in-domain corpus $\mathcal{D}_{d} \subseteq \mathcal{D}_g$}
    \Output{A trained translation model }
    
    1. \tcp{\hl{model warm-up stage}} 
    \hspace{.5em}\underline{Train} a \textit{base model} $\mathcal{M}$ on general domain corpus $\mathcal{D}_g$ for $K$ number of updates \\
     \vspace{-1.em}
    
    2.  \tcp{\hl{model fine-tuning stage}}  
    \For{n\_epochs}{
        \hspace{-.2em} \Na \textbf{Get} pred. score for each $(x_i, y_i) \in \mathcal{D}_{d}$ \\
        \hspace{-1em} \Nb \textbf{Rank} $\mathcal{D}_d$  based on these scores \\
        \hspace{-1em} \Nc \textbf{Find} $\mathcal{D}_s \subset \mathcal{D}_d$ by picking a \textit{data-selection window} \\
        \hspace{-1em} \Nd \underline{Fine-tune} $\mathcal{M}$  on $\mathcal{D}_s$
      }
    \caption{Online Curriculum Strategy}
    \label{alg:online}
\end{algorithm}

Unlike deterministic curriculum, in this strategy the selected subset $\mathcal{D}_s$ changes dynamically in each epoch of the \textit{model fine-tuning} stage through instantaneous feedback from the current NMT model. Specifically, in each epoch, we rank $(x_i, y_i) \in \mathcal{D}_d$ by leveraging the prediction scores from the emerging NMT model which assigns a probability to each token in the target sentence $y_i$. We then take the average of the token-level probabilities to get the sentence-level probability score $P_{\theta}(y_i|x_i)$ which is regarded as the \textit{prediction score} for the sentence pair $(x_i, y_i)$. Formally,

\begin{equation}
\small
    P_{\theta}(y_i|x_i) = \frac{1}{\ell} \sum_{t=1}^\ell  p_{\theta}(y_{i,t}|y_{i,<t}, x_i)
\end{equation}

\noindent This prediction score indicates the \textit{confidence} of the emerging NMT model to generate the target sentence $y_i$ from the source sentence $x_i$. Intuitively, if the model can predict the target sentence of a training data sample $(x_i, y_i)$ with higher confidence, it indicates that the sample is \textit{too easy} for the model and might not contain useful information to improve the NMT model further at that state. On the other hand, if a target sentence is predicted with lower confidence, it indicates that the training data sample might be \textit{too hard} for the model at that state or it might be a noisy sample. Subsequently, including such hard or noisy samples in training at that state might degrade the NMT model performance.

\begin{figure}[t!]
  \centering
\scalebox{0.8}{
  \includegraphics[width=1\linewidth,trim=1 1 1 1,clip]{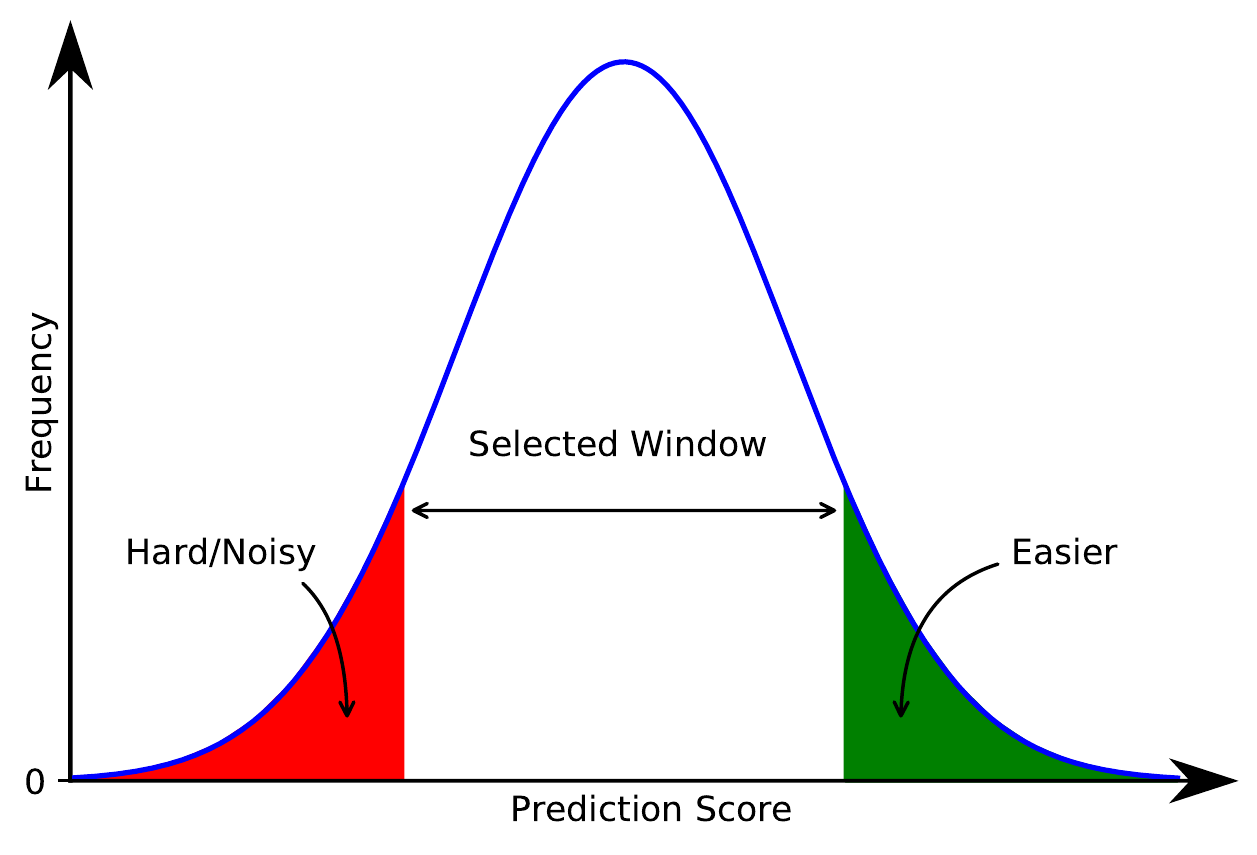}}
\vspace{-0.5em}
\caption{\small {Conceptual demonstration of online curriculum. We rank the  bitext pairs based on the prediction scores of the emerging NMT model and pick a \textit{data-selection window} which discards easy and hard/noisy ones.}}
\label{fig:online-curriculum}
\vspace{-.5em}
\end{figure}

Algorithm \ref{alg:online} presents the pseudo-code of our online data-selection curriculum strategy. After the \textit{model warm-up} stage, we fine-tune $\mathcal{M}$ for $n\_epochs$ on data subset $\mathcal{D}_s$ which is selected in every epoch {based on the emerging NMT models' confidence. Specifically, in the beginning of each epoch in the \textit{model fine-tuning} stage,} we find the \textit{prediction score} $P_{\theta}(y_i|x_i)$ of each sample $(x_i, y_i) \in \mathcal{D}_d$. We then rank $\mathcal{D}_d$ based on these scores and select $\mathcal{D}_s \subset \mathcal{D}_d$ by picking a \textit{data-selection window} in the ranked data. Finally, we fine-tune $\mathcal{M}$ on $\mathcal{D}_s$ for that epoch. We present the conceptual demonstration of our online curriculum strategy in Figure \ref{fig:online-curriculum}. For picking the data-selection window in ranked $\mathcal{D}_d$, we investigate two methods:

\paragraph{$\bullet$ Static Data-selection Window} Here in each epoch, we discard a \textit{constant} amount {(\%)} of easy and hard/noisy samples from $\mathcal{D}_d$ based on the prediction scores and select the rests as $\mathcal{D}_s$. Even though in this method the size of the selected data subset ($\mathcal{D}_s$) is {constant} through out the model fine-tuning stage, unlike deterministic strategy the samples in $\mathcal{D}_s$ \textit{varies} from epoch-to-epoch due to the change in their prediction scores by the emerging NMT model $\mathcal{M}$.\footnote{We present an illustrative example of this phenomenon in Appendix \ref{app:illust-static}.}

\paragraph{$\bullet$ Dynamic Data-selection Window} Unlike the static approach, here we \textit{change} the data-selection window size in subsequent epochs. This can be done in two ways:

\begin{enumerate}[label=(\roman*),leftmargin=*]
\vspace{-.5em}
    \item \textit{Expansion:} Begin with a smaller window ($|\mathcal{D}_s| << |\mathcal{D}_d|$) and gradually increase the window to a maximum size $\lambda_{max}$.
    \item \textit{Shrink:} Begin with a larger window ($|\mathcal{D}_s| \sim |\mathcal{D}_d|$) and gradually decrease the window to a minimum size $\lambda_{min}$.
\end{enumerate} 

\noindent To change the data-selection window size, we use \textit{linear scheduler}\footnote{We experiment with other schedulers (Appendix \ref{app:sched-dynamic}).} which can be regarded as a function $\lambda(t)$ to map the current training epoch $t$ to a scalar value. This value is regarded as the data-selection window size at epoch $t$. Formally, 

\vspace{-.8em}
\begin{equation}
\small
\begin{split}
    \lambda_{\text{exp}}(t) &= 
    \begin{cases}
    \lambda_{init} + {l_{inc}}*{t},& \text{if } \lambda_{\text{exp}}(t) < \lambda_{max}\\
    \lambda_{max},              & \text{otherwise}
    \end{cases} \\
    \lambda_{\text{shr}}(t) &= 
    \begin{cases}
    \lambda_{init} - {l_{dec}}*{t},& \text{if } \lambda_{\text{shr}}(t) > \lambda_{min}\\
    \lambda_{min},              & \text{otherwise}
    \end{cases}
\end{split}
\label{eq:lin-scheduler}
\end{equation}
\normalsize
\noindent where $\lambda_{init}$ is the initial window size which is smaller for \textit{expansion} and larger for \textit{shrink}, and ${l_{inc}}$, ${l_{dec}}$ are the hyperparameters of the  schedulers.

\section{Experimental Setup}

\paragraph{Datasets}

We conduct experiments on six language pairs: three \textit{high-resource} including English (En) to/from German (De), Hungarian (Hu), and Estonian (Et); and three \textit{low-resource} including English (En) to/from Hausa (Ha), Tamil (Ta), and Malay (Ms). We use the dataset provided in WMT 2021\footnote{\href{http://www.statmt.org/wmt21/}{http://www.statmt.org/wmt21/}} --- De and Ha are from \textit{News shared task}, while the remaining four pairs are from \textit{Large-Scale Multilingual MT shared task}. For En$\leftrightarrow$De, we use newstest2019 as validation set and report test results on newstest2020. For En$\leftrightarrow$Ha, we randomly split the provided dev set into validation and test set. For the other language pairs, we use the official evaluation data (dev and devtest) as validation and test sets. Table~\ref{tab:dataset-stats} presents the dataset statistics after cleaning and deduplication. For high-resource language pairs, we consider {formal texts} parallel data corpora sources as in-domain ($\mathcal{D}_{d} \subset \mathcal{D}_g$), while for low-resource pairs, we do not differentiate between general-domain and in-domain corpus ($\mathcal{D}_{d} \coloneqq \mathcal{D}_g$). 
Table \ref{tab:indomain-corp} shows the in-domain corpora sources for high-resource language pairs.

\paragraph{Model Settings}
We use the Transformer \cite{NIPS2017_3f5ee243} implementation in Fairseq \cite{ott-etal-2019-fairseq}; details of our model architecture settings are given in Appendix \ref{app:arch-settings}. We use sentencepiece library\footnote{\href{https://github.com/google/sentencepiece}{https://github.com/google/sentencepiece}} to learn joint Byte-Pair-Encoding (BPE) of size 32,000 and 16,000 for En$\leftrightarrow$De and En$\leftrightarrow$Ha, respectively. For other language pairs, we use the official sentencepiece model provided in \textit{Large-Scale Multilingual MT shared task}.
We filter out parallel data with a length longer than 250 tokens during training. All experiments are evaluated using SacreBLEU \cite{post-2018-call}.

For LM training in the modified Moore-Lewis method (\S \ref{subsec:deterministic-curriculum}), we use the implementation in Fairseq. For in-domain LM training, we use 5M sentences from newscrawl, while we combine 10M commoncrawl data with newscrawl totaling 15M sentences to train the general-domain LM.

\begin{table}[t!]
\centering
\scalebox{0.78}{
\begin{tabular}{l|cc|c|c}
\toprule
{\textbf{Pair}} & \multicolumn{2}{c}{\textbf{Train}} & \textbf{Validation}  & \textbf{Test}  \\ 
& All-data & In-domain & & \\
\midrule
En-De     & 89,893,260 & 2,152,577 &  1997 & 1418  \\
De-En     & 89,893,260 & 2,152,577 &  2000 & 785  \\
En-Hu     &  53,219,023 & 647,106 & 997 & 1012   \\ 
En-Et     &  19,685,308 & 869,537 & 997 & 1012   \\ 
\midrule
En-Ms     & 1,694,311 & -- & 997 & 1012   \\ 
En-Ta     & 1,064,032   & -- & 997 & 1012 \\ 
En-Ha     &  685,780 & -- & 1000 & 1000   \\ 
\bottomrule
\end{tabular}
}
\caption{\small Dataset statistics after cleaning and deduplication.}
\label{tab:dataset-stats}
\end{table}

\begin{table}[t!]
\centering
\scalebox{0.8}{
\begin{tabular}{l|l}
\toprule
{\textbf{Pair}} & \textbf{In-domain Corpora} \\
\midrule
En-De     & Europarl, News Commentary  \\
En-Hu     & EUconst, Europarl, GlobalVoices, Wikipedia, \\
& WikiMatrix, WMT-News    \\ 
En-Et     &  EUconst, Europarl, WikiMatrix, WMT-News   \\ 
 
\bottomrule
\end{tabular}
}
\caption{In-domain corpora sources for high-resource language pairs.}
\label{tab:indomain-corp}
\end{table}


\makeatletter
\def\ifPositive#1{%
    \@ifnextchar{-}%
      {\expandafter\@secondoftwo\remove@to@nnil}%
      {\expandafter\@firstoftwo\remove@to@nnil}%
        #1\@nnil
}
\makeatother
\definecolor{darkgreen}{rgb}{0,.5,0}
\newcommand{\diffscore}[1]{%
    \ifPositive{#1}%
        {\textcolor{darkgreen}{\tiny{$#1$}}}%
        {\textcolor{red}{\tiny{$#1$}}}
}

\begin{table*}[!t]
\centering
\small
\scalebox{0.86}{
\begin{tabular}{llc|cc|cc|cc}
\toprule
\textbf{Type} & \textbf{Setting} & \textbf{\%data-used} & \multicolumn{2}{c}{\textbf{En-Ha}}  &        \multicolumn{2}{c}{\textbf{En-Ms}} &\multicolumn{2}{c}{\textbf{En-Ta}}
\\

& & \textbf{in each ep.} & \textbf{$\rightarrow$} & \textbf{$\leftarrow$} & \textbf{$\rightarrow$} & \textbf{$\leftarrow$} & 
\textbf{$\rightarrow$} & \textbf{$\leftarrow$}  
\\   
\toprule
Warm-up Model   & All Data    & 100\%      & 13.5 & 14.7 & 30.8 & 27.3 & 8.5 & 15.4 \\
\midrule
Converged Model  & All Data  & 100\%      & 14.3 & 15.3 & 31.4 & 27.9 & 8.8 & 15.7 \\
\midrule

\multicolumn{9}{c}{\textit{Warm-up Stage Model Fine-tuning (Ft.)}} \\

\midrule
\multirow{1}{*}{Traditional Ft.} & 
All Data & 100\%    &  14.4 \diffscore{+0.1} & 15.6 \diffscore{+0.3} & 31.5 \diffscore{+0.1} & 28.0 \diffscore{+0.1} & 8.7 \diffscore{-0.1} & 15.7 \diffscore{+0.0}\\

\midrule
\multirow{3}{*}{Det. Curricula} & 
LASER & 40\%      & 14.6 \diffscore{+0.3} & \textbf{17.5} \diffscore{+2.2} & 31.7 \diffscore{+0.3} & 28.2 \diffscore{+0.3} & 8.8 \diffscore{+0.0} & 15.9 \diffscore{+0.2} \\
& 
Dual Cond. CE (\text{DCCE}) & 40\%      & 14.3 \diffscore{+0.0} & 16.3 \diffscore{+1.1}& 31.4 \diffscore{+0.0} & 28.2 \diffscore{+0.3} & 8.6 \diffscore{-0.2} & 16.0 \diffscore{+0.3} \\
& 
Mod. Moore-Lewis (\text{MML}) & 40\%      &  {14.8} \diffscore{+0.5} & 15.6 \diffscore{+0.3} & 31.6 \diffscore{+0.2} & 28.1 \diffscore{+0.2} & 9.0 \diffscore{+0.2} & 15.6 \diffscore{-0.1} \\

\midrule
\multirow{2}{*}{Online Curricula} 
&  Static Window  & 40\%      & 14.7 \diffscore{+0.4} & 16.1 \diffscore{+0.8} & 31.6 \diffscore{+0.2} & 28.3 \diffscore{+0.4} & {9.1} \diffscore{+0.3} & \textbf{16.2} \diffscore{+0.5} \\
&  Dynamic Window  \\
& \quad Expansion & <40\%      &  \textbf{14.9} \diffscore{+0.6} & 16.6 \diffscore{+1.3} & \textbf{31.8} \diffscore{+0.4} & \textbf{28.4} \diffscore{+0.5} & \textbf{9.2} \diffscore{+0.4} & 16.1 \diffscore{+0.4} \\

& \quad Shrink & <40\%      &   14.7 \diffscore{+0.4} & 15.9 \diffscore{+0.6} & 31.4 \diffscore{+0.0} & 28.3 \diffscore{+0.4} & 8.8 \diffscore{+0.0} & 16.0 \diffscore{+0.3} \\

\midrule

Det. + Online & Hybrid & 15-20\%      &   14.7 \diffscore{+0.4} & 16.4 \diffscore{+1.1} & 31.5 \diffscore{+0.1} & 28.2 \diffscore{+0.3} & {9.1} \diffscore{+0.2} & 15.9 \diffscore{+0.2} \\

\bottomrule
\end{tabular}}
\vspace{-0.3em}
\caption{\small Main results for \textbf{low-resource} languages. Here, the data-percentage represents \textit{general-domain} data ($\mathcal{D}_g$) and we do not differentiate between general-domain and in-domain corpus ($\mathcal{D}_{d} \coloneqq \mathcal{D}_g$). Subscript values denote the BLEU score differences from the respective converged model.}
\label{tab:low-res-main-results} 
\vspace{-.5em}
\end{table*}


\paragraph{Baselines}
We compare our methods with the \textbf{converged model}, which is a standard NMT model trained on all the general-domain data ($\mathcal{D}_g$) until convergence. Additionally, we compare both the deterministic and online curriculum approaches with the \textbf{traditional fine-tuning} where we fine-tune the base model from the warm-up stage with all the in-domain train data ($\mathcal{D}_d$) until convergence.

\section{Results }
\label{sec:res}

The main results for the low- and high-resource languages are shown in Tables~\ref{tab:low-res-main-results} and~\ref{tab:high-res-main-results}, respectively. For low-resource languages, we train the warm-up stage models for 20K updates, while the converged models are trained for 50K updates. For high-resource languages,  we train for 50K and 100K updates for the warm-up and converged models, respectively. In traditional fine-tuning (\textit{Traditional Ft.} row in the Tables), we use all the available in-domain data ($\mathcal{D}_d$) in each fine-tuning epoch. On the other hand, for both deterministic and online curricula, we use \textit{at most} 40\% of the available in-domain data ($\mathcal{D}_s \subset \mathcal{D}_d$) in each fine-tuning epoch.

Comparing the performance of traditional fine-tuning with the \textit{Converged Model} on low-resource languages (Table \ref{tab:low-res-main-results}), we see that both of these perform on par. This is not surprising as both approaches use all the data ($\mathcal{D}_g$) during the whole training (for low-resource languages $\mathcal{D}_{d} \coloneqq \mathcal{D}_g$). The only difference between the two approaches is -- while the converged model continues to train the base model from the warm-up stage, the traditional fine-tuning approach resets the base model's meta-parameters (\textit{e.g.,} learning-rate, lr-scheduler, data-loader, optimizer) and continue the training.

For high-resource languages (Table \ref{tab:high-res-main-results}), we fine-tune the base model only on the in-domain training data ($\mathcal{D}_d \subset \mathcal{D}_g$) in traditional fine-tuning, while the converged model continues to train the base model on all the general-domain data ($\mathcal{D}_g$). Here, traditional fine-tuning performs better than the converged model on En-De (+0.4) and En-Et (+0.9) but exhibits poor performance on the other four directions by 0.7 BLEU score on an average.

In the following, we discuss the performance of our proposed curriculum approaches:

\vspace{-.3em}
\subsection{Performance of Deterministic Curricula}
\vspace{-.3em}

First, we consider the performance of deterministic curriculum approaches 
on low-resource languages. From Table \ref{tab:low-res-main-results}, we see that fine-tuning the base model on the data subset ($\mathcal{D}_s$) selected by \text{LASER} outperforms the baseline (\textit{Converged Model}) on five out of six translation tasks with a +2.2 BLEU gain in Ha-En. For the other two scoring methods, dual conditional cross-entropy (\text{DCCE}) and modified Moore-Lewis (\text{MML}), we also see a better or similar performance on 5/6 translation tasks. Compared to the traditional fine-tuning, the deterministic approaches perform better in most of the tasks -- on average +0.5, +0.4, +0.2 BLEU gains for \text{LASER}, \text{DCCE}, and \text{MML}, respectively.


\begin{table*}[t!]
\centering
\small
\scalebox{0.86}{
\begin{tabular}{llc|cc|cc|cc}
\toprule
\textbf{Type} & \textbf{Setting} & \textbf{\%data-used} & \multicolumn{2}{c}{\textbf{En-De}}  &        \multicolumn{2}{c}{\textbf{En-Hu}} &\multicolumn{2}{c}{\textbf{En-Et}}
\\

& & \textbf{in each ep.} & \textbf{$\rightarrow$} & \textbf{$\leftarrow$} & \textbf{$\rightarrow$} & \textbf{$\leftarrow$} & 
\textbf{$\rightarrow$} & \textbf{$\leftarrow$}  
\\   
\toprule
Warm-up Model   & All Data   & 100\%+OOD      & 34.9 & 40.8 & 33.9 & 36.0 & 35.7 & 37.1 \\
\midrule
Converged Model  & All Data  & 100\%+OOD     & 36.1 & 41.2 & 35.9 & \textbf{36.7} & 36.7 & \textbf{38.2} \\
\midrule

\multicolumn{9}{c}{\textit{Warm-up Stage Model Fine-tuning (Ft.)}} \\

\midrule
\multirow{1}{*}{Traditional Ft.} & All In-domain Data & 100\%    &  36.5 \diffscore{+0.4} & 40.5 \diffscore{-0.3} & 35.4 \diffscore{-0.5} & 35.5 \diffscore{-1.2} & 37.6 \diffscore{+0.9} & 37.4 \diffscore{-0.8} \\

\midrule
\multirow{3}{*}{Det. Curricula} & 
LASER & 40\%      & 37.6 \diffscore{+1.5} & 42.4 \diffscore{+1.2} & 36.0 \diffscore{+0.1} & 35.9 \diffscore{-0.8} & 37.6 \diffscore{+0.9} & 37.8 \diffscore{-0.4} \\
& 
Dual Cond. CE (\text{DCCE}) & 40\%      & 37.9 \diffscore{+1.8} & 43.0 \diffscore{+1.8} & {36.2} \diffscore{+0.3} & 35.4 \diffscore{-1.3} & {38.0} \diffscore{+1.3} & 37.3 \diffscore{-0.9} \\

& 
Mod. Moore-Lewis (\text{MML}) & 40\%      &  37.1 \diffscore{+1.0} & 41.7 \diffscore{+0.5} & 35.8 \diffscore{-0.1} & 35.2 \diffscore{-1.5} & 37.3 \diffscore{+0.6} & 37.4 \diffscore{-0.8} \\

\midrule

\multirow{2}{*}{Online Curricula} 
&  Static Window  & 40\%      & 37.3 \diffscore{+1.2} & 41.4 \diffscore{+0.2} & 36.1 \diffscore{+0.2} & 35.4 \diffscore{-1.3} & 37.9 \diffscore{+1.2} & 37.7 \diffscore{-0.5} \\
&  Dynamic Window  \\
& \quad Expansion & <40\%     & 37.3 \diffscore{+1.2} & 41.6 \diffscore{+0.4} & \textbf{36.4} \diffscore{+0.5} & 35.6 \diffscore{-1.1} & \textbf{38.1} \diffscore{+1.4} & 37.8 \diffscore{-0.4} \\
& \quad Shrink & <40\%      &   37.0 \diffscore{+0.9} & 41.2 \diffscore{+0.0} & 36.0 \diffscore{+0.1} & 35.7 \diffscore{-1.0} & {38.0}\diffscore{+1.3} & 37.6 \diffscore{-0.6} \\

\midrule

Det. + Online & Hybrid & 15-20\%      & \textbf{38.1} \diffscore{+2.0} & \textbf{43.3} \diffscore{+2.1} & 36.1 \diffscore{+0.2} & 35.6 \diffscore{-1.1} & 37.9\diffscore{+1.2} & 37.3 \diffscore{-0.9} \\

\bottomrule
\end{tabular}}
\vspace{-0.3em}
\caption{\small Main results for \textbf{high-resource} languages. Here, the data-percentage represents only \textit{In-domain} data ($\mathcal{D}_d$) from Table \ref{tab:dataset-stats} and \textit{100\%+OOD} denotes \textit{All-data} ($\mathcal{D}_g$). Subscript values denote the BLEU score differences from respective converged model.}
\label{tab:high-res-main-results} 
\vspace{-0.5em}
\end{table*}



In Table \ref{tab:high-res-main-results}, we see a similar trend of better performance of the  deterministic curricula over the converged model on high-resource languages. Specifically, fine-tuning on the data subset selected by utilizing the scoring of both \text{LASER} and \text{DCCE} performs better on four out of six translation tasks, while the \text{MML}-based method achieves better performances on three tasks. The margins of improved performances for the high-resource languages are \textit{higher} compared to the low-resource languages: +1.4, +0.9, +0.7 BLEU gains on average for \text{DCCE}, \text{LASER}, and \text{MML}, respectively over the baseline. If we compare with the traditional fine-tuning, the deterministic curriculum approaches perform better in most of the tasks -- on average +1.2, +0.8, +0.4 BLEU scores better for \text{DCCE}, \text{LASER}, and \text{MML}, respectively.



To observe the better performance of the deterministic curriculum approaches more clearly, we fine-tune the base model from the warm-up stage with different percentages of ranked data selected by the bitext scoring methods. 
Figure \ref{fig:det-cur-diff-perc} shows the results. We observe that there exist multiple subsets of data ($\mathcal{D}_s \subset \mathcal{D}_d$), fine-tuning the base model on which demonstrates better performance compared to the \textit{Converged Model} and \textit{traditional fine-tuning}. For De-En, traditional fine-tuning (on 100\% data) reduces the BLEU score by 0.3 from the base model, while fine-tuning on most of the subsets selected by the deterministic curricula leads to improved performances. For Hu-En, traditional fine-tuning diminishes the performance of the base model by 0.5 BLEU. Unlike De-En, here we could not find a subset by the deterministic curricula fine-tuning on which improves the performance of the base model.

\subsection{Performance of Online Curricula}
\vspace{-.3em}

Our online curriculum approaches perform on par with the deterministic curricula for both low- and high-resource languages as shown in Tables \ref{tab:low-res-main-results} and \ref{tab:high-res-main-results}, respectively. Unlike deterministic approaches, here we leverage the emerging models' prediction scores without using any external pretrained scoring methods. In our static window approach, we {discard the top 30\% and bottom 30\%} sentence pairs from the ranked $\mathcal{D}_d$ and fine-tune the base model from the warm-up stage on the remaining 40\% data ($\mathcal{D}_s$). The selected data in $\mathcal{D}_s$ \textit{vary} dynamically from epoch-to-epoch due to the change in the prediction scores of the emerging NMT models. From the results (Tables \ref{tab:low-res-main-results}, \ref{tab:high-res-main-results}), we notice that the data-selection by \textit{Static Window} method outperforms the \textit{Converged Model} on ten out of twelve translation tasks and the BLEU scores are comparable to the deterministic curriculum approaches.

In our dynamic window approach, we either expand or shrink the window size, where the selected window is confined to the range of 30\% to 70\% of the ranked $\mathcal{D}_d$, \textit{i.e.,} $\mathcal{D}_s$ can be at most 40\% of $\mathcal{D}_d$. In window \textit{expansion}, we start $\mathcal{D}_s$ with 10\% of $\mathcal{D}_d$ and linearly increase it to 40\% in the subsequent epochs, while in the window \textit{shrink} method we start $\mathcal{D}_s$ with 40\% and linearly decrease to 10\% of $\mathcal{D}_d$. With dynamic window \textit{expansion}, we achieve slightly better (up to +0.5 BLEU) performance on 10/12 translation tasks compared to the static window method. On the other hand, the dynamic window \textit{shrink} method performs slightly lower than window expansion in most of the translation tasks.


\begin{figure*}[t!]
\centering

\scalebox{0.99}{
\begin{subfigure}{.25\textwidth}
  \centering
  \includegraphics[scale=0.28]{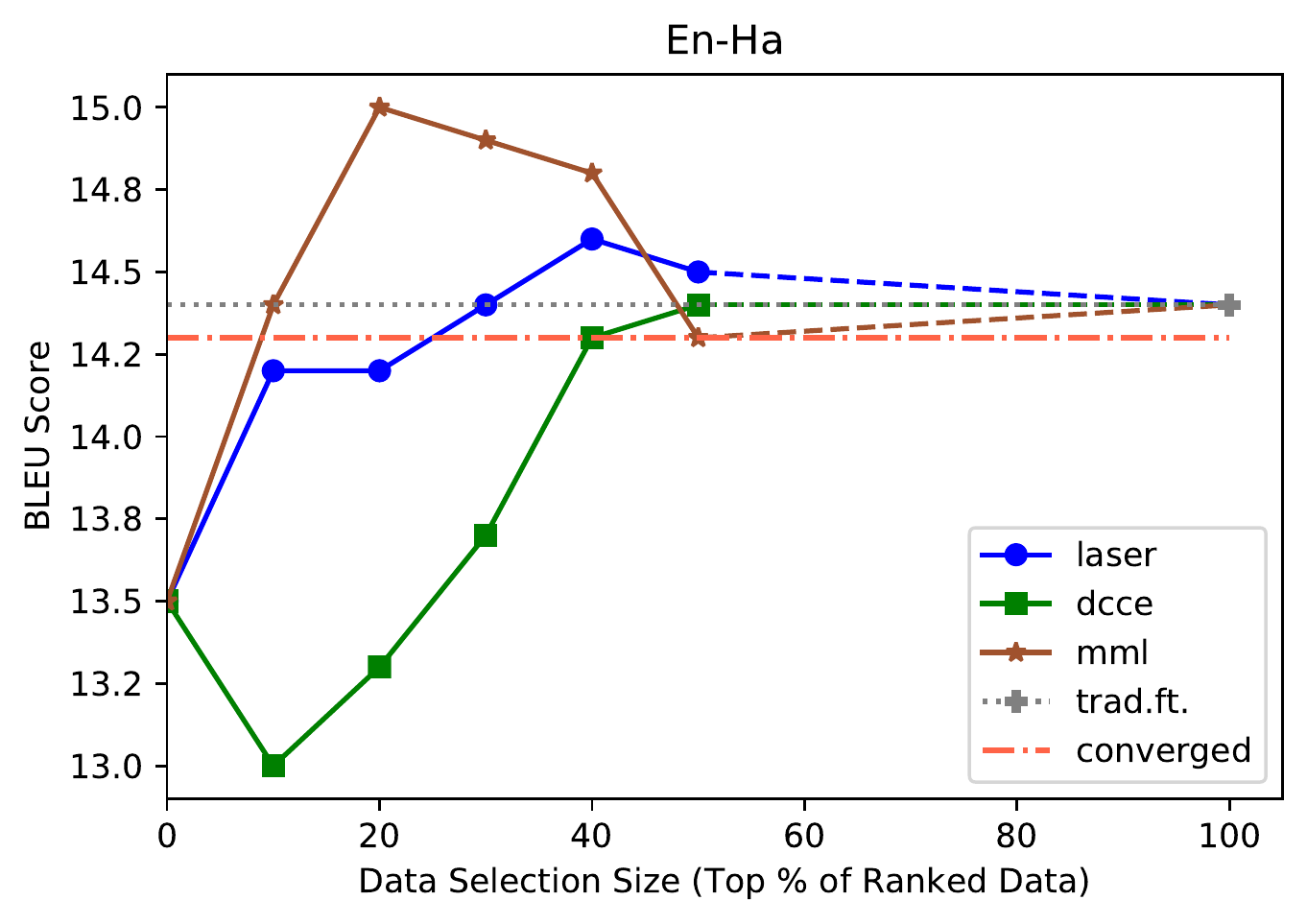}
\end{subfigure}%
\begin{subfigure}{.25\textwidth}
  \centering
  \includegraphics[scale=0.28]{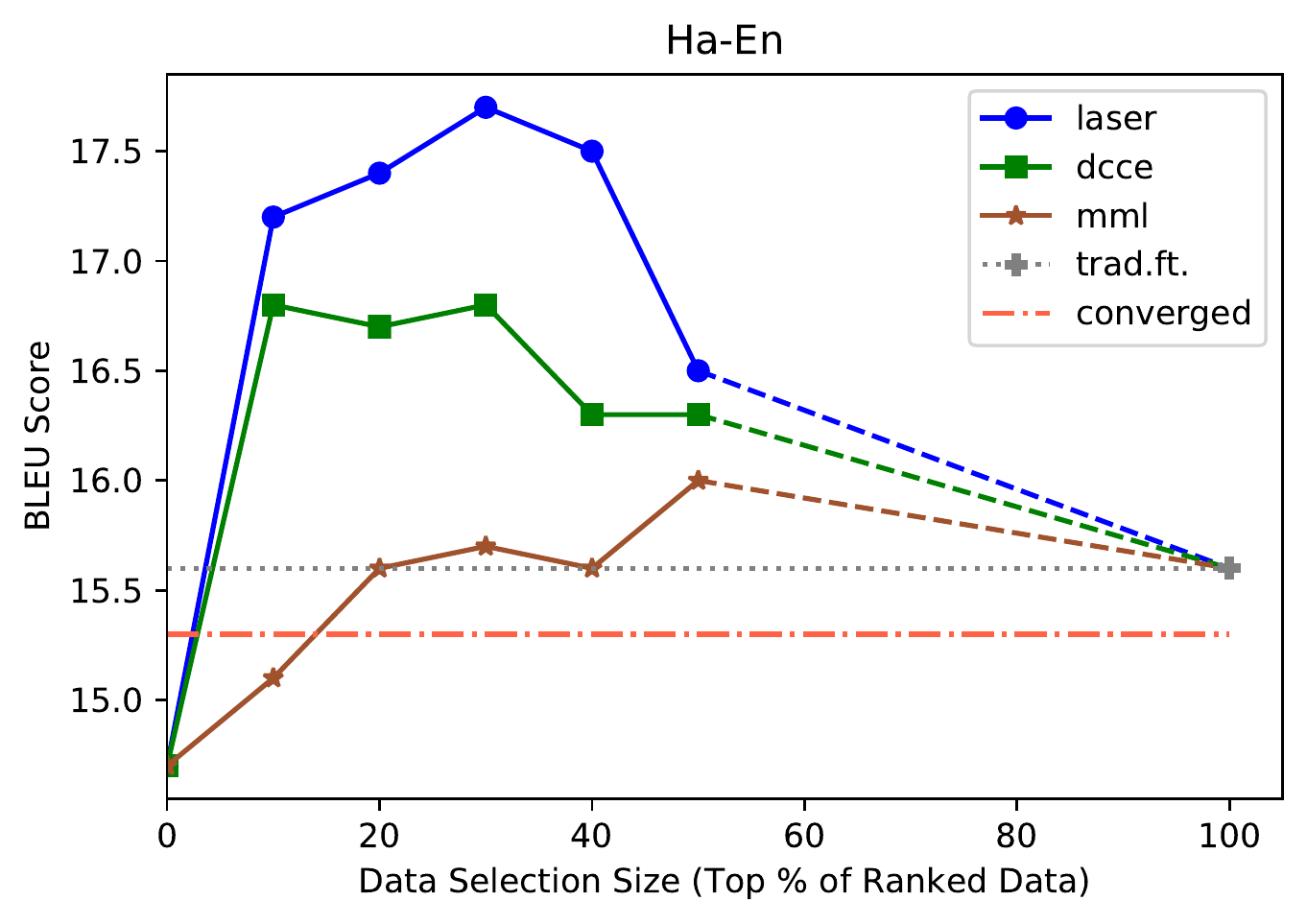}
\end{subfigure}

\begin{subfigure}{.25\textwidth}
  \centering
  \includegraphics[scale=0.28]{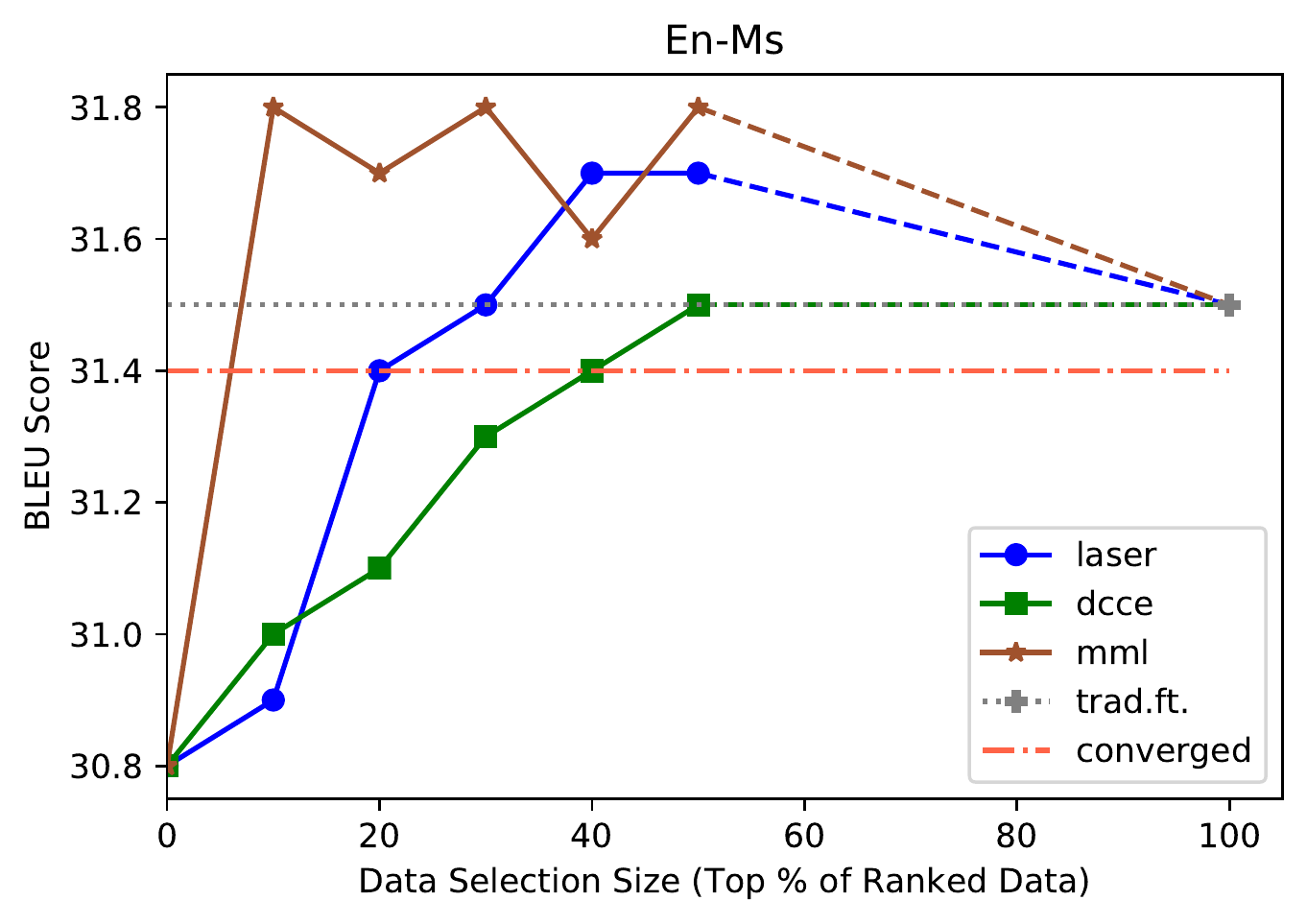}
\end{subfigure}%

\begin{subfigure}{.25\textwidth}
  \centering
  \includegraphics[scale=0.28]{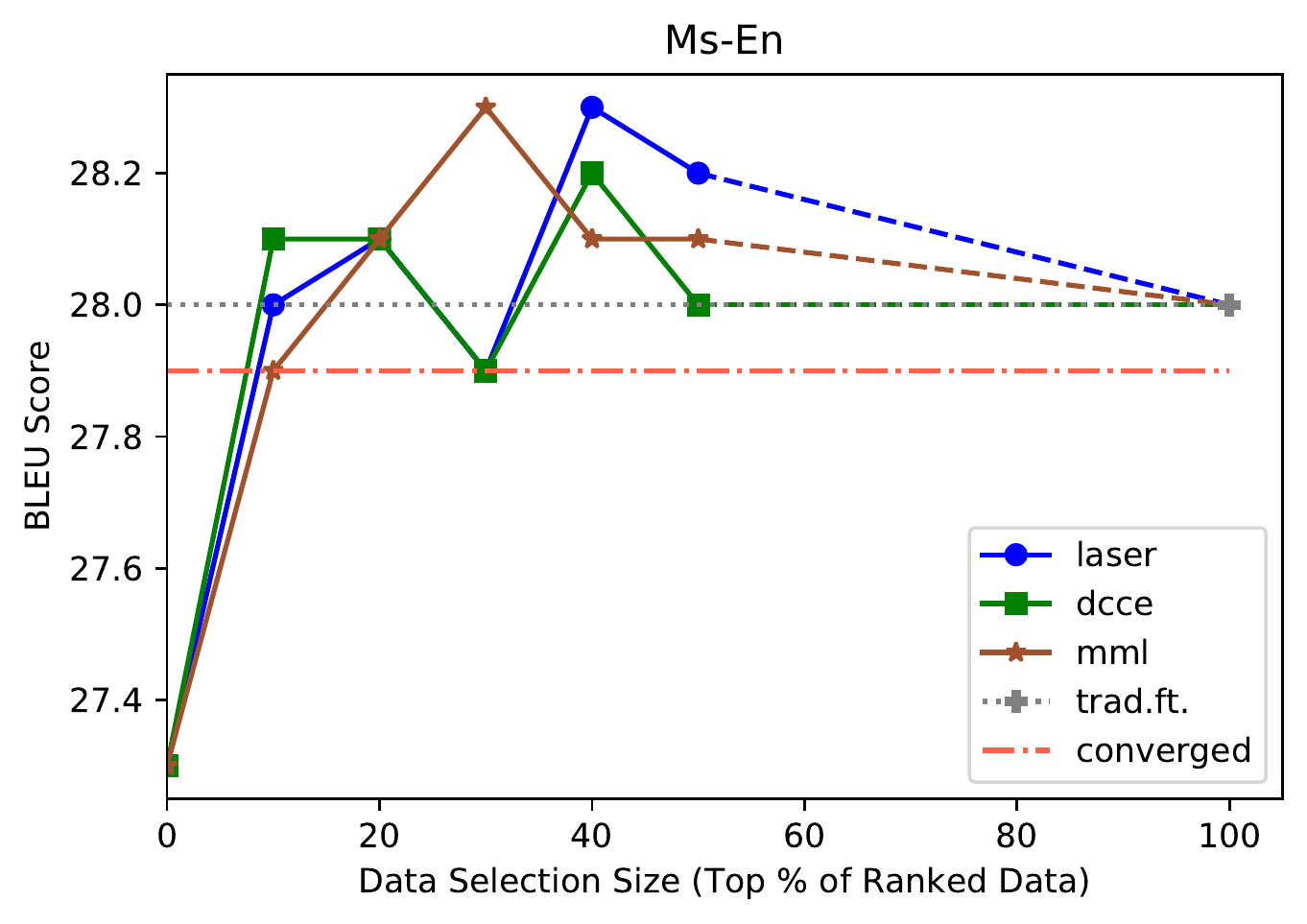}
\end{subfigure}
}

\scalebox{0.99}{
\begin{subfigure}{.25\textwidth}
  \centering
  \includegraphics[scale=0.28]{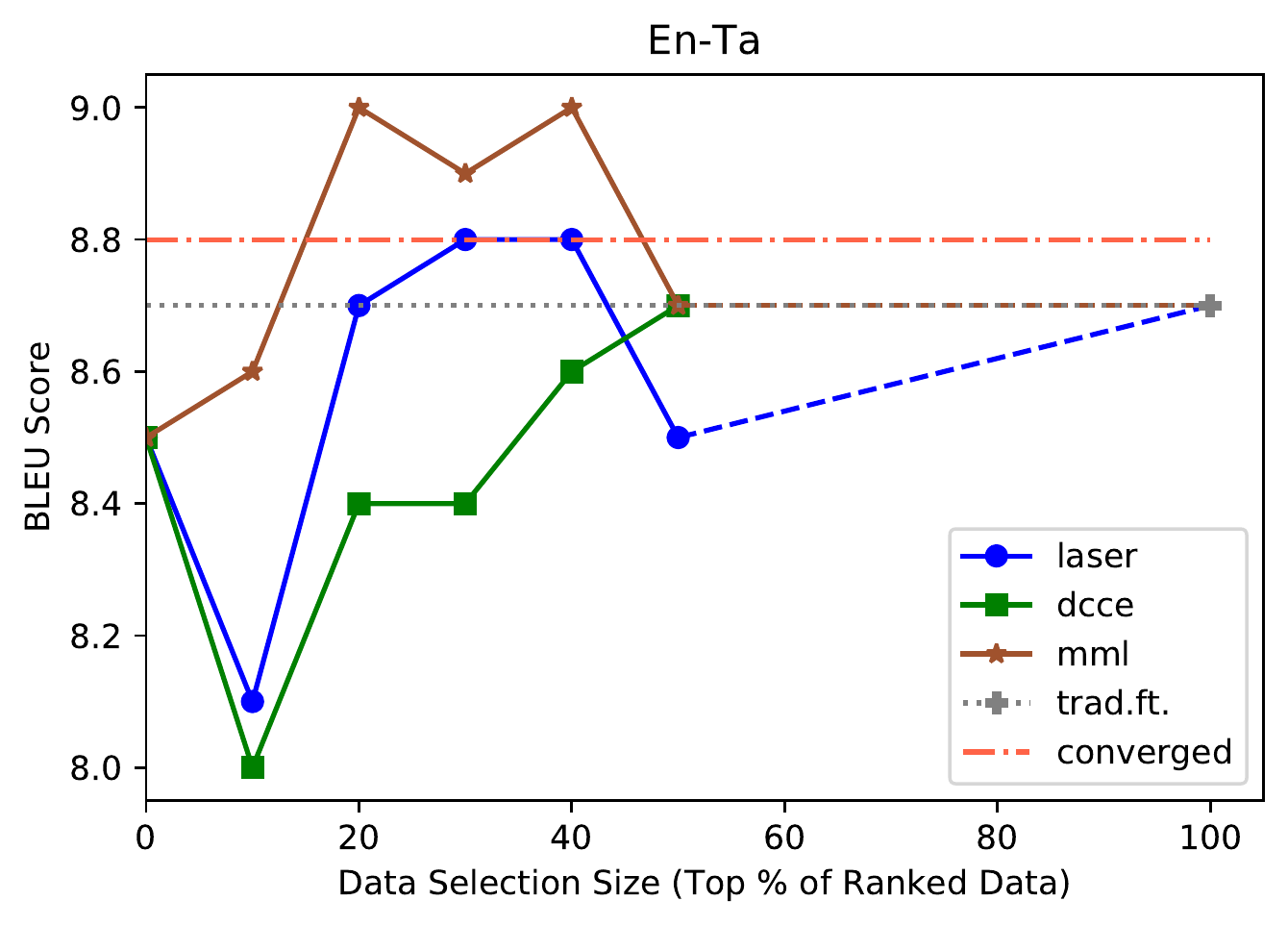}
\end{subfigure}%
\begin{subfigure}{.25\textwidth}
  \centering
  \includegraphics[scale=0.28]{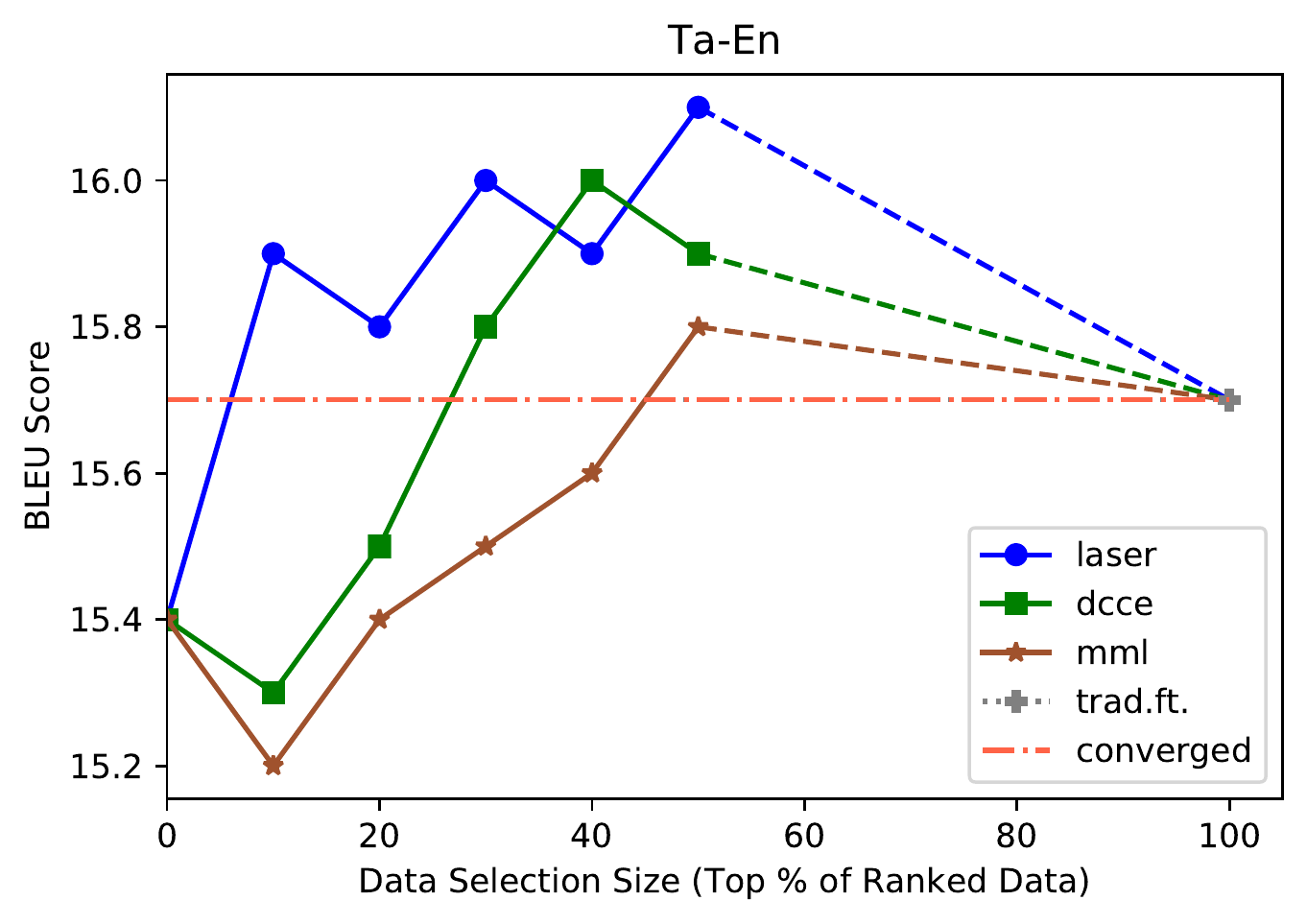}
\end{subfigure}

\begin{subfigure}{.25\textwidth}
  \centering
  \includegraphics[scale=0.28]{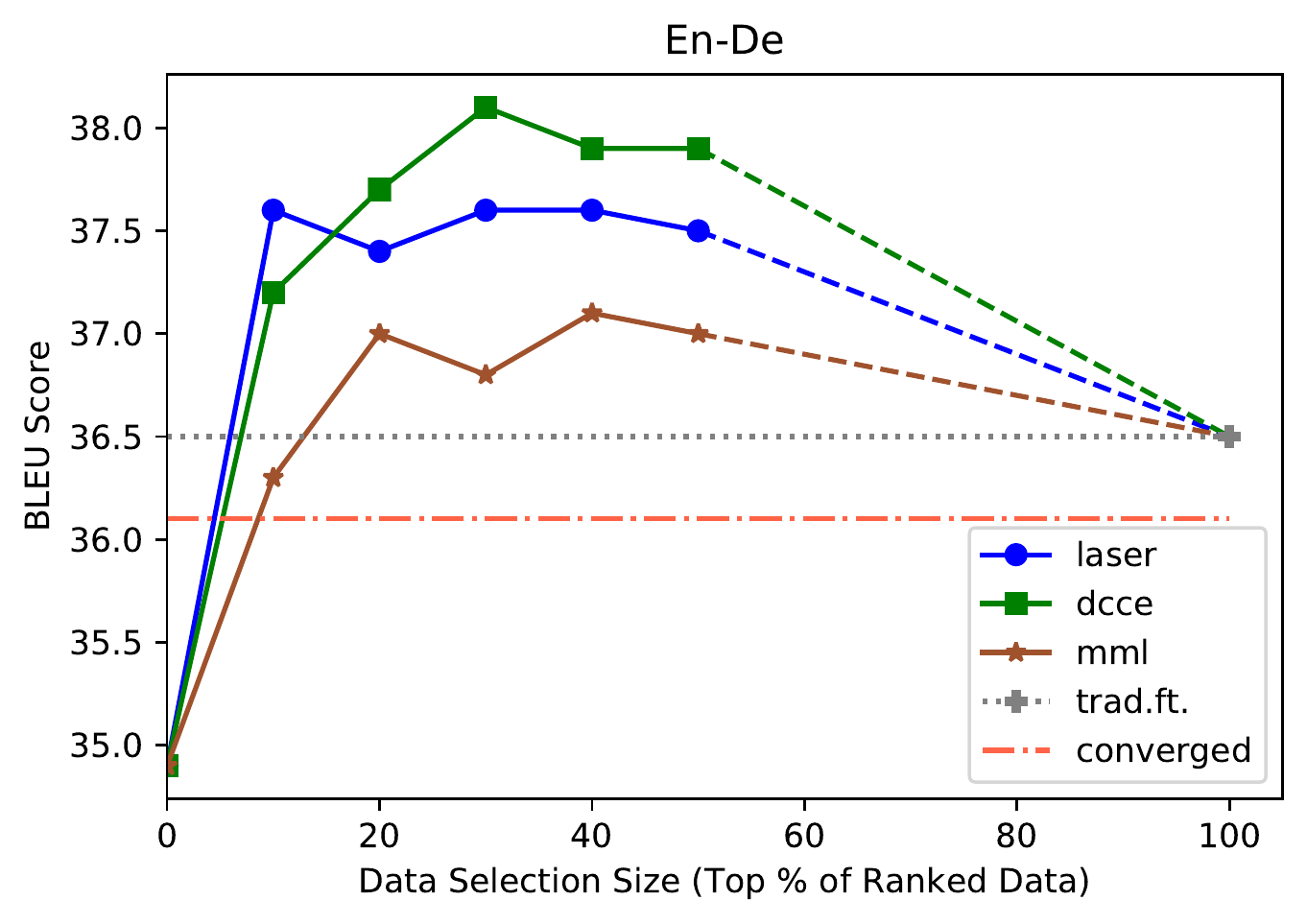}
\end{subfigure}%

\begin{subfigure}{.25\textwidth}
  \centering
  \includegraphics[scale=0.28]{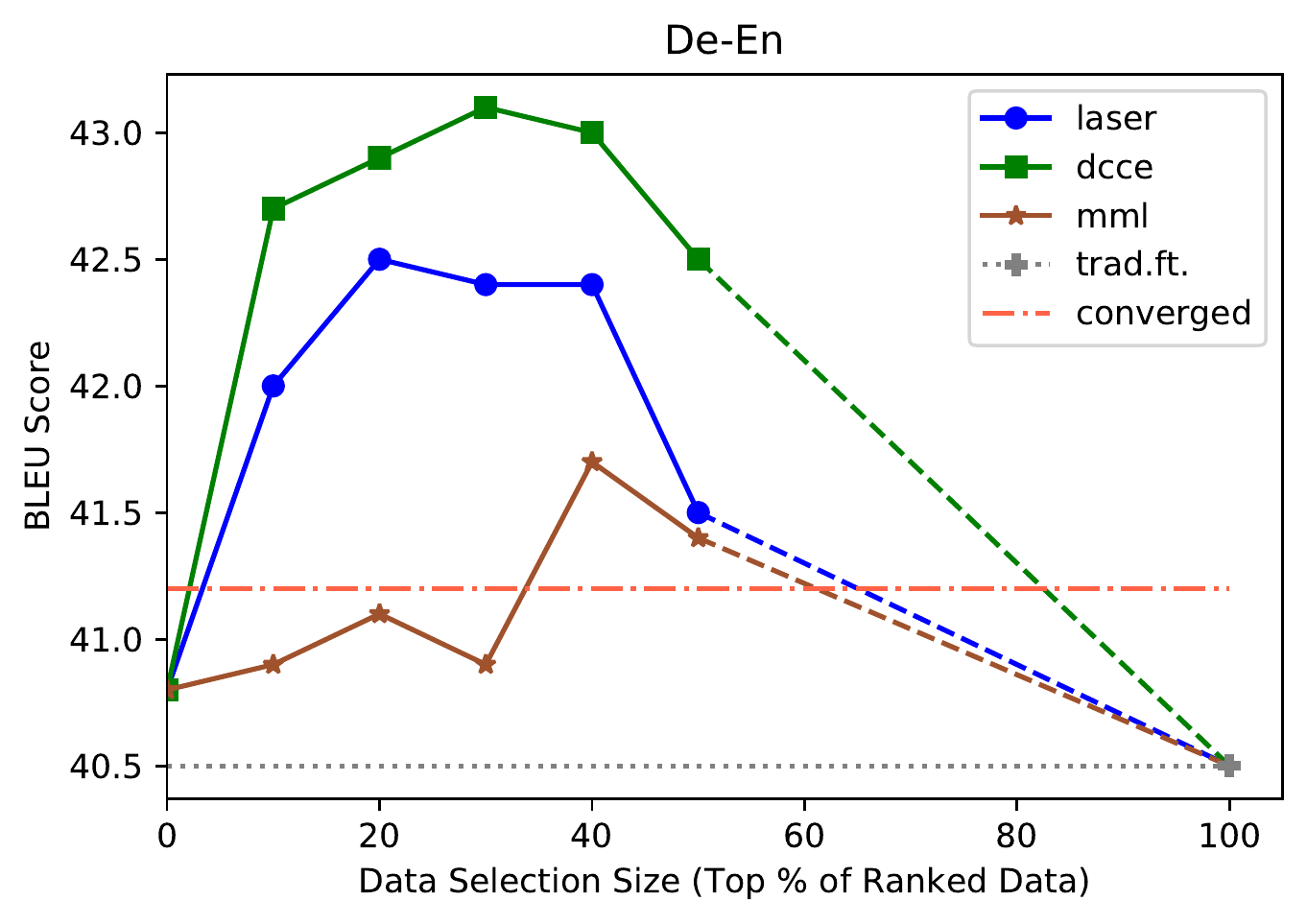}
\end{subfigure}
}

\scalebox{0.99}{
\begin{subfigure}{.25\textwidth}
  \centering
  \includegraphics[scale=0.28]{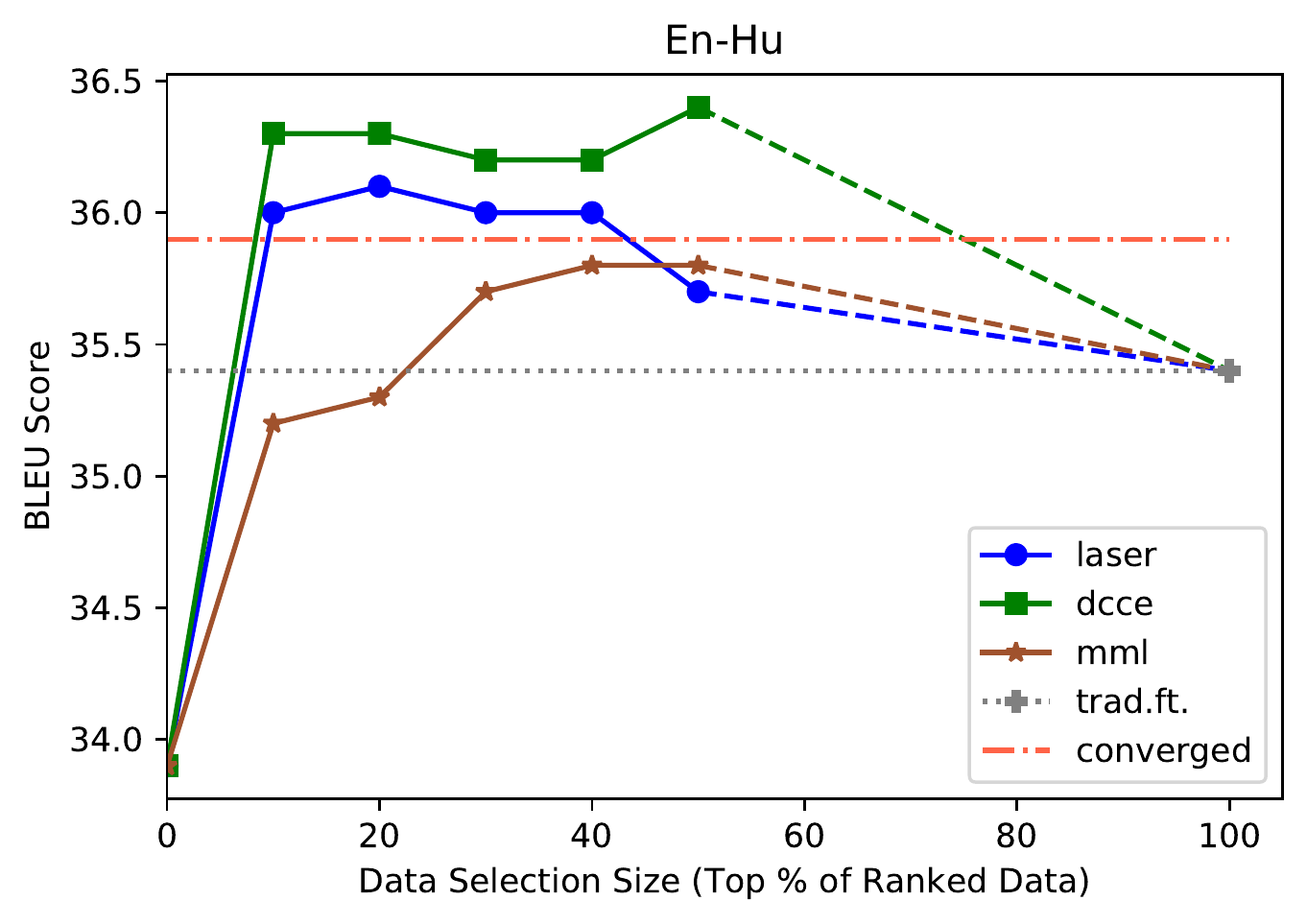}
\end{subfigure}%
\begin{subfigure}{.25\textwidth}
  \centering
  \includegraphics[scale=0.28]{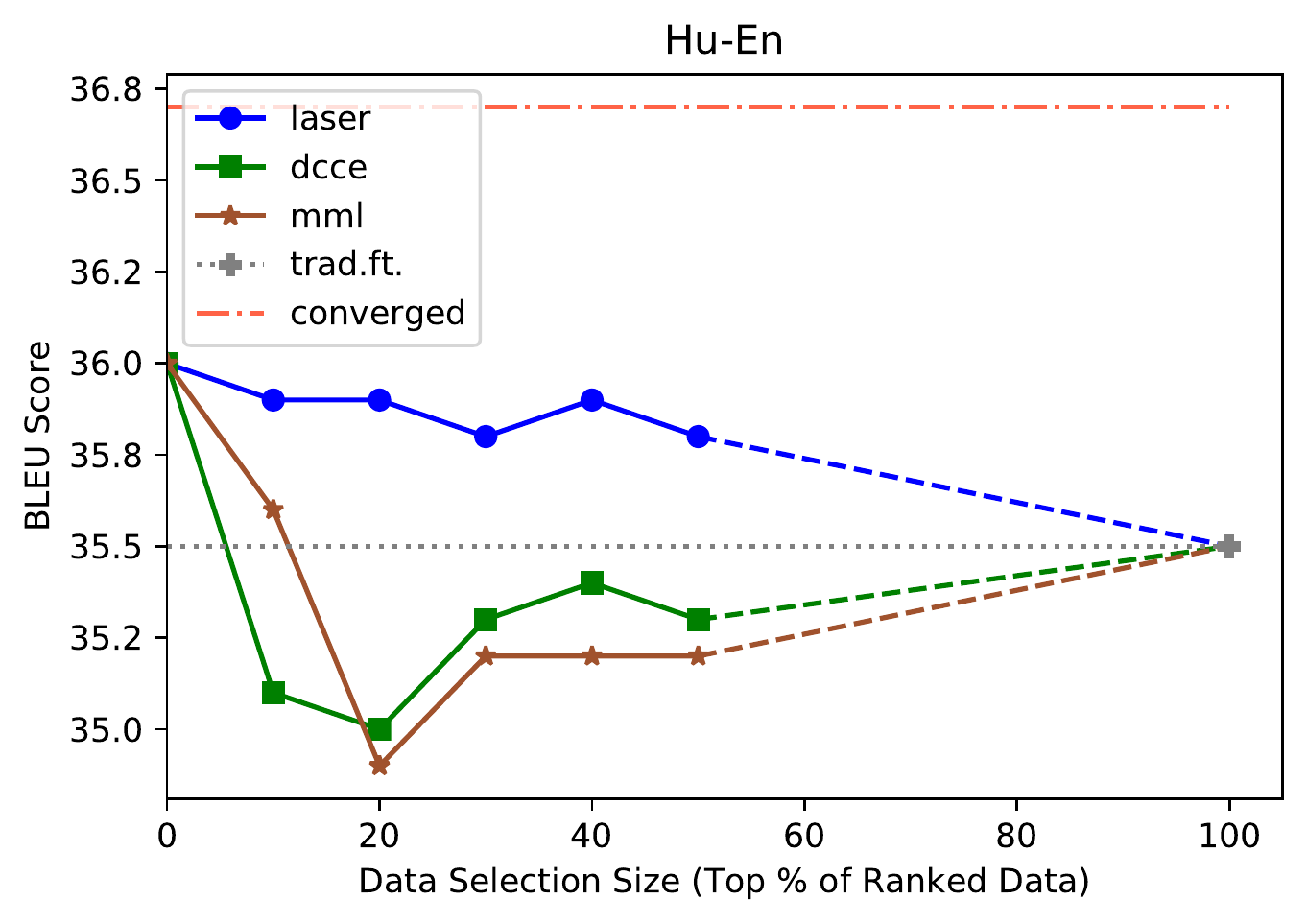}
\end{subfigure}

\begin{subfigure}{.25\textwidth}
  \centering
  \includegraphics[scale=0.28]{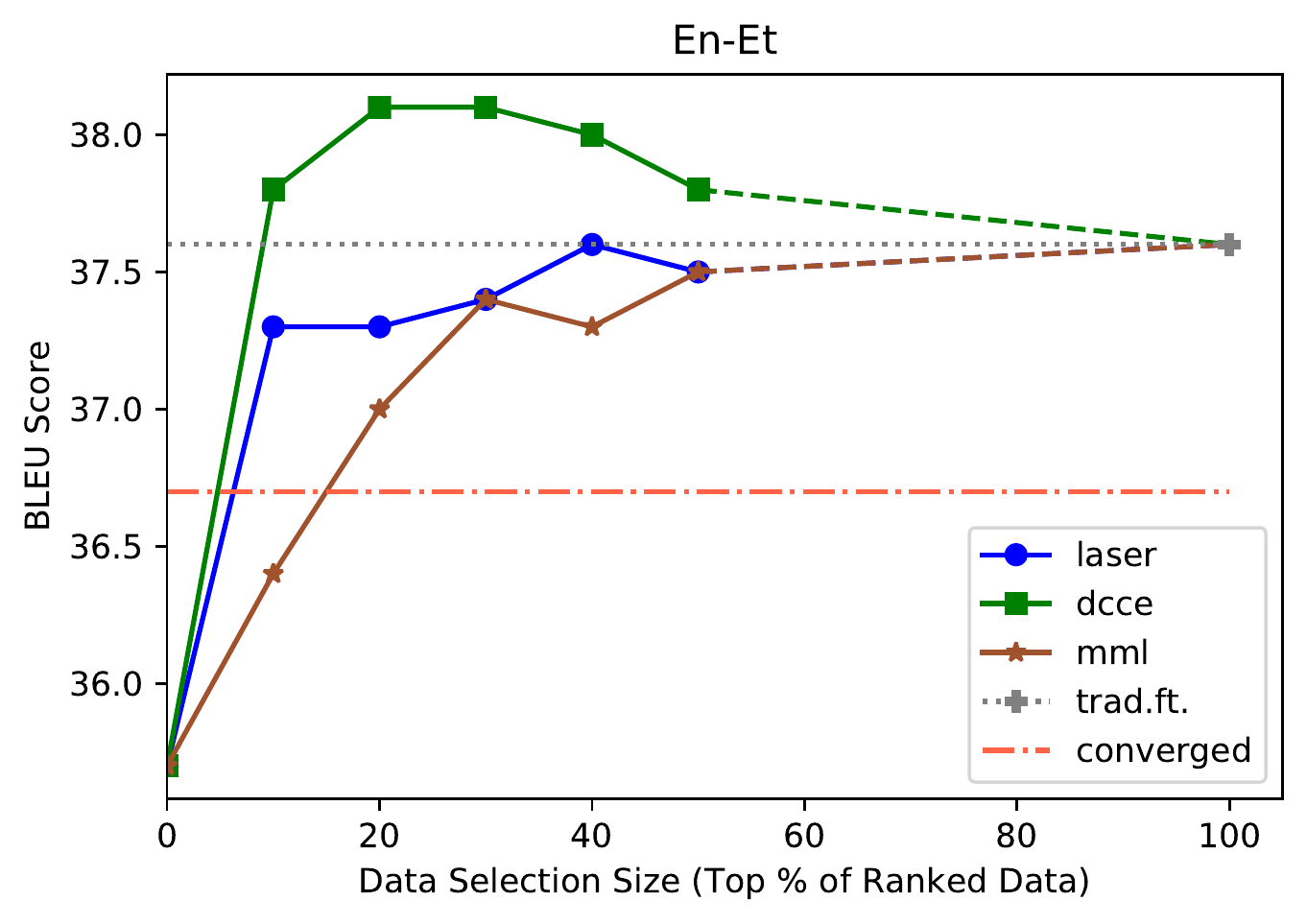}
\end{subfigure}%

\begin{subfigure}{.25\textwidth}
  \centering
  \includegraphics[scale=0.28]{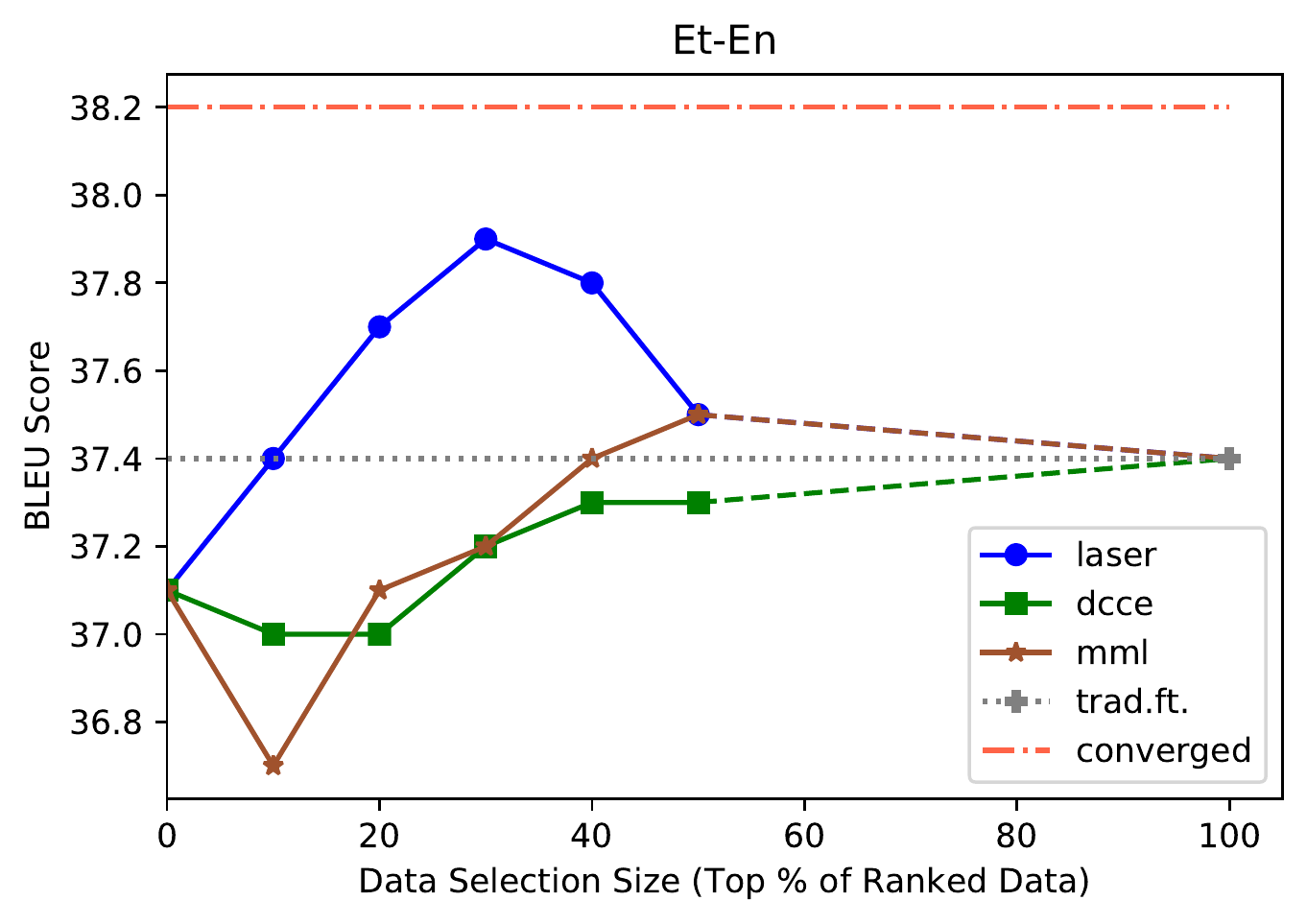}
\end{subfigure}
}
\vspace{-0.5em}
\caption{{Fine-tuned }\textit{warm-up stage model} on different sizes of ranked data (deterministic curricula).}
\label{fig:det-cur-diff-perc}
\end{figure*}

\section{Discussion and Analysis}


\vspace{-.2em}
\subsection{Hybrid Curriculum}
\vspace{-.2em}
To benefit from both deterministic and online curricula, we combine the two strategies. Specifically, we consider three subsets of data comprising of the top 50\% of $\mathcal{D}_d$ ranked by each of the three bitext scoring methods in \S \ref{subsec:deterministic-curriculum} and keep the common bitext pairs (intersection of three subsets). We then apply the static window data-selection curriculum on these bitext pairs, where we {discard the top 10\%  and bottom 10\% } pairs (ranked by the emerging model's prediction scores) and fine-tune the base model from the warm-up stage on the remaining bitext. Depending on the language pairs, the data percentage for the fine-tuning stage ($\mathcal{D}_s$) becomes 15-20\% of $\mathcal{D}_d$. Despite being a smaller subset of data for fine-tuning, performances of the hybrid curriculum strategy are better on 10 out of 12 translation tasks compared to the baseline (Table \ref{tab:low-res-main-results}, \ref{tab:high-res-main-results}). Notably, for En-De and De-En, the hybrid curriculum attains +2.0 and +2.1 BLEU scores compared to the converged model.

\vspace{-.2em}
\subsection{Are All Data Useful Always?}
\label{subsec:are-all-data-useful-always}
\vspace{-.2em}

Our proposed curriculum training framework uses all the data ($\mathcal{D}_g$) in the model warm-up stage and then utilizes subsets of in-domain data ($\mathcal{D}_s$) in the model fine-tuning stage. This resembles the ``\textit{formal education system}'' where students first learn the general subjects with the same weights and later concentrate more on a selected subset of specialized subjects. The first stage teaches them the base knowledge which is useful in the ensuing stage. We observe a similar phenomenon in our experiments. From Table \ref{tab:high-res-all-vs-indomain-results}, we see that the performance of the NMT model using only the in-domain data is worse than using all general-domain data (-8.1 BLEU on average). Moreover, our curriculum training framework outperforms the converged model that uses all the data throughout the training in most of the translation tasks by a sizable margin. This indicates that \textit{not all data are useful all the time}. Additionally, Figure \ref{fig:det-cur-diff-perc} shows that in most scenarios, fine-tuning on selected data subsets $\mathcal{D}_s$ outperform the traditional fine-tuning that uses all the data. This observation validates our intuition that some data samples are not only redundant but also detrimental to the NMT model's performance.


\begin{table}
\centering
\small
\scalebox{0.9}{
\begin{tabular}{l|cc|cc|cc}
\toprule
\textbf{Corpus} & \multicolumn{2}{c}{\textbf{En-De}}  &        \multicolumn{2}{c}{\textbf{En-Hu}} &\multicolumn{2}{c}{\textbf{En-Et}}
\\

& \textbf{$\rightarrow$} & \textbf{$\leftarrow$} & \textbf{$\rightarrow$} & \textbf{$\leftarrow$} & 
\textbf{$\rightarrow$} & \textbf{$\leftarrow$}  
\\   
\toprule
All-data & 36.1 & 41.2 & 35.9 & 36.7 & 36.7 & 38.2 \\
\midrule
In-domain & 32.6 & 33.5 & 25.5 & 23.6 & 30.6 & 30.3 \\

\bottomrule
\end{tabular}}
\vspace{-0.5em}
\caption{Results for high-resource languages on \textbf{all-data ($\mathcal{D}_g$) vs. in-domain data ($\mathcal{D}_d$)} when trained from a random state until convergence.}
\label{tab:high-res-all-vs-indomain-results} 
\end{table}


\begin{figure*}[t!]
\centering
\scalebox{0.99}{
\begin{subfigure}{.17\textwidth}
  \centering
  \includegraphics[scale=0.2]{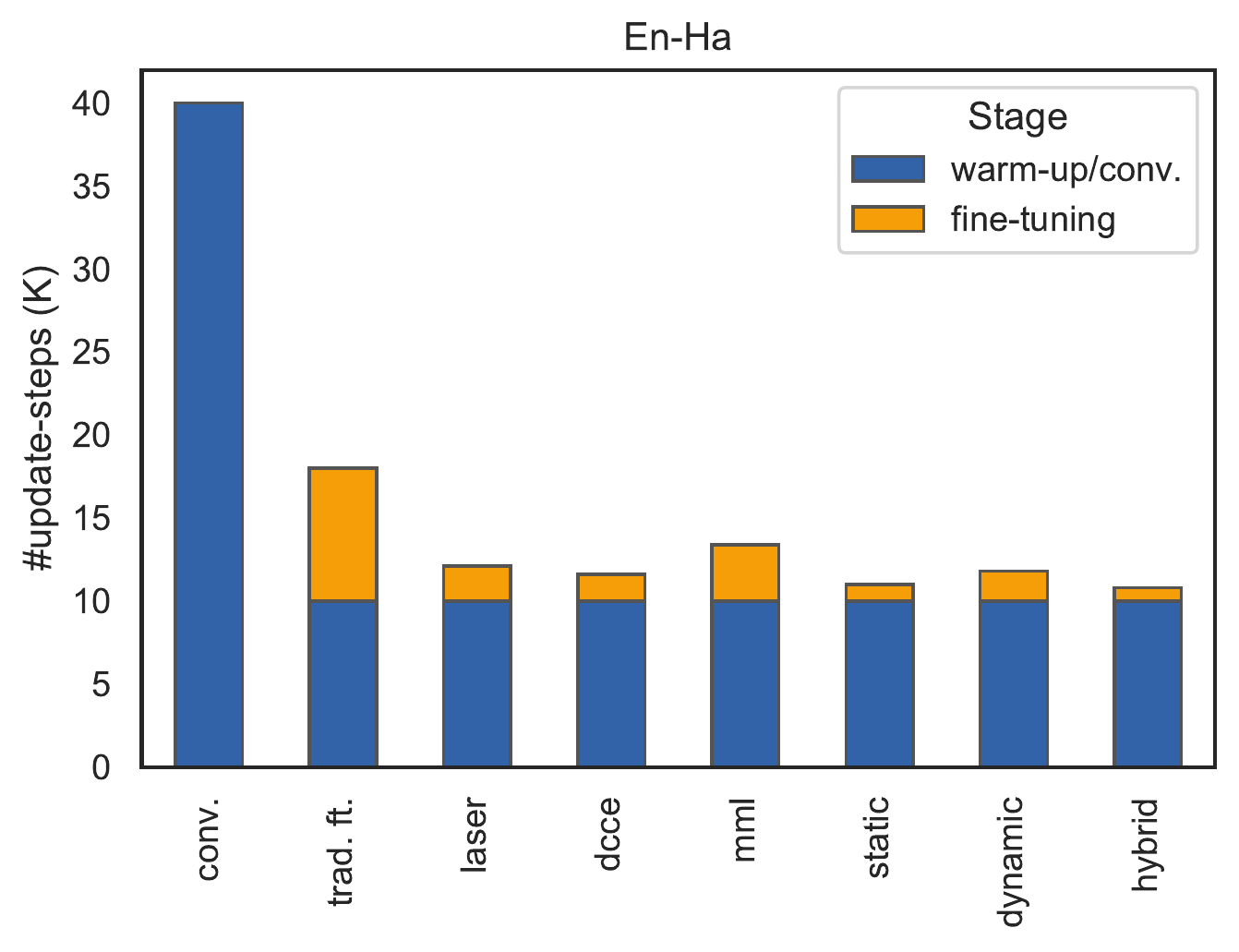}
\end{subfigure}%
\begin{subfigure}{.17\textwidth}
  \centering
  \includegraphics[scale=0.2]{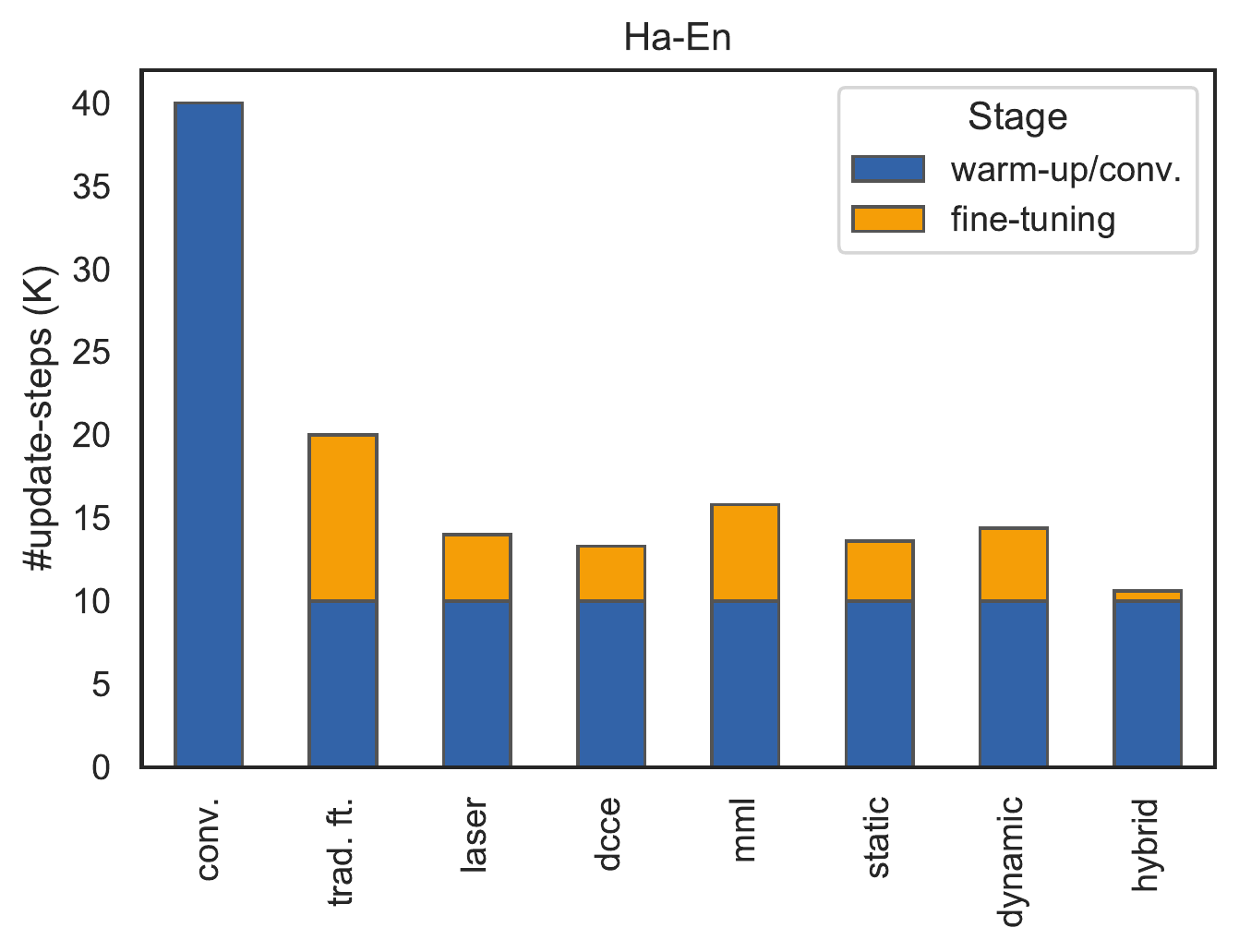}
\end{subfigure}

\begin{subfigure}{.17\textwidth}
  \centering
  \includegraphics[scale=0.2]{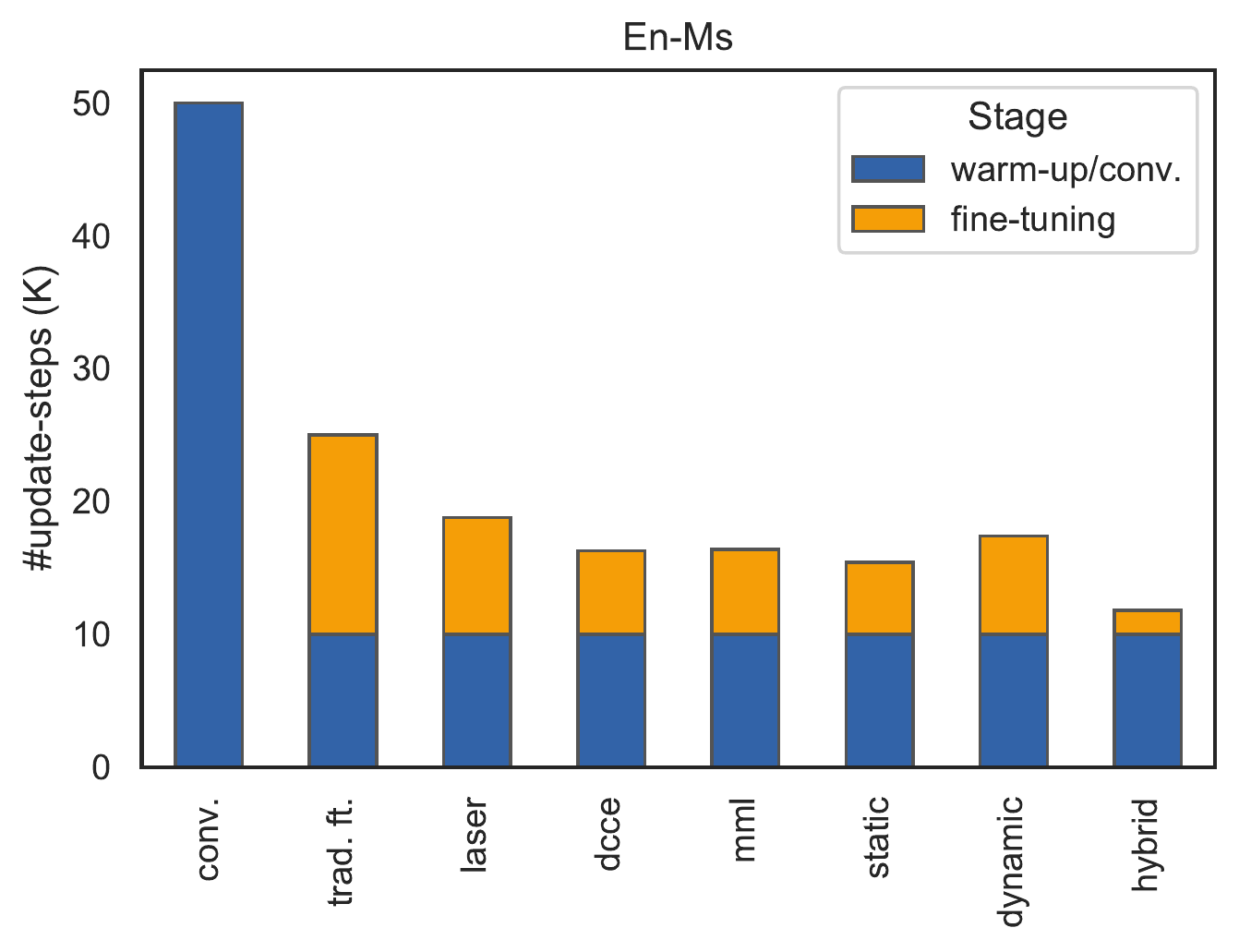}
\end{subfigure}%
\begin{subfigure}{.17\textwidth}
  \centering
  \includegraphics[scale=0.2]{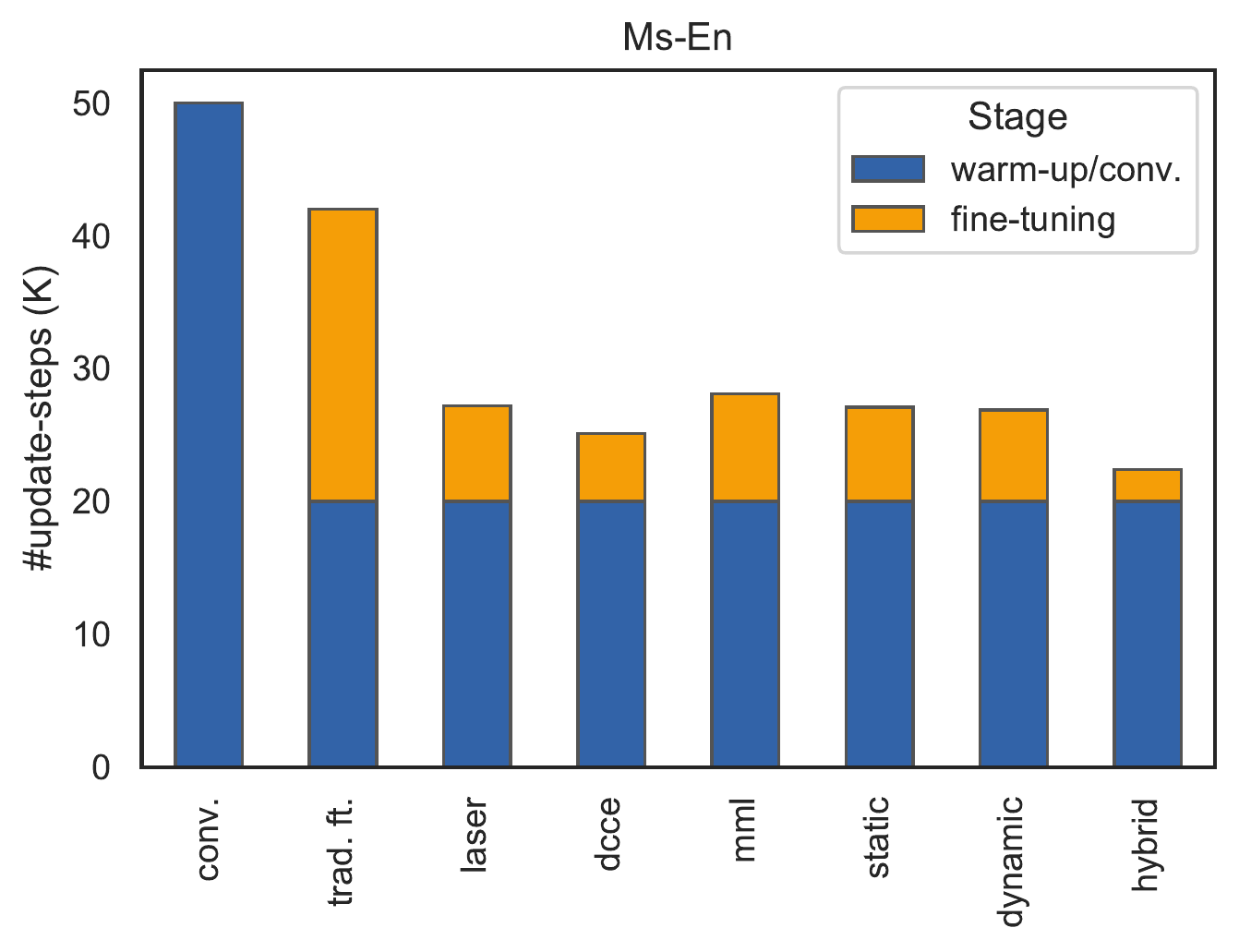}
\end{subfigure}

\begin{subfigure}{.17\textwidth}
  \centering
  \includegraphics[scale=0.2]{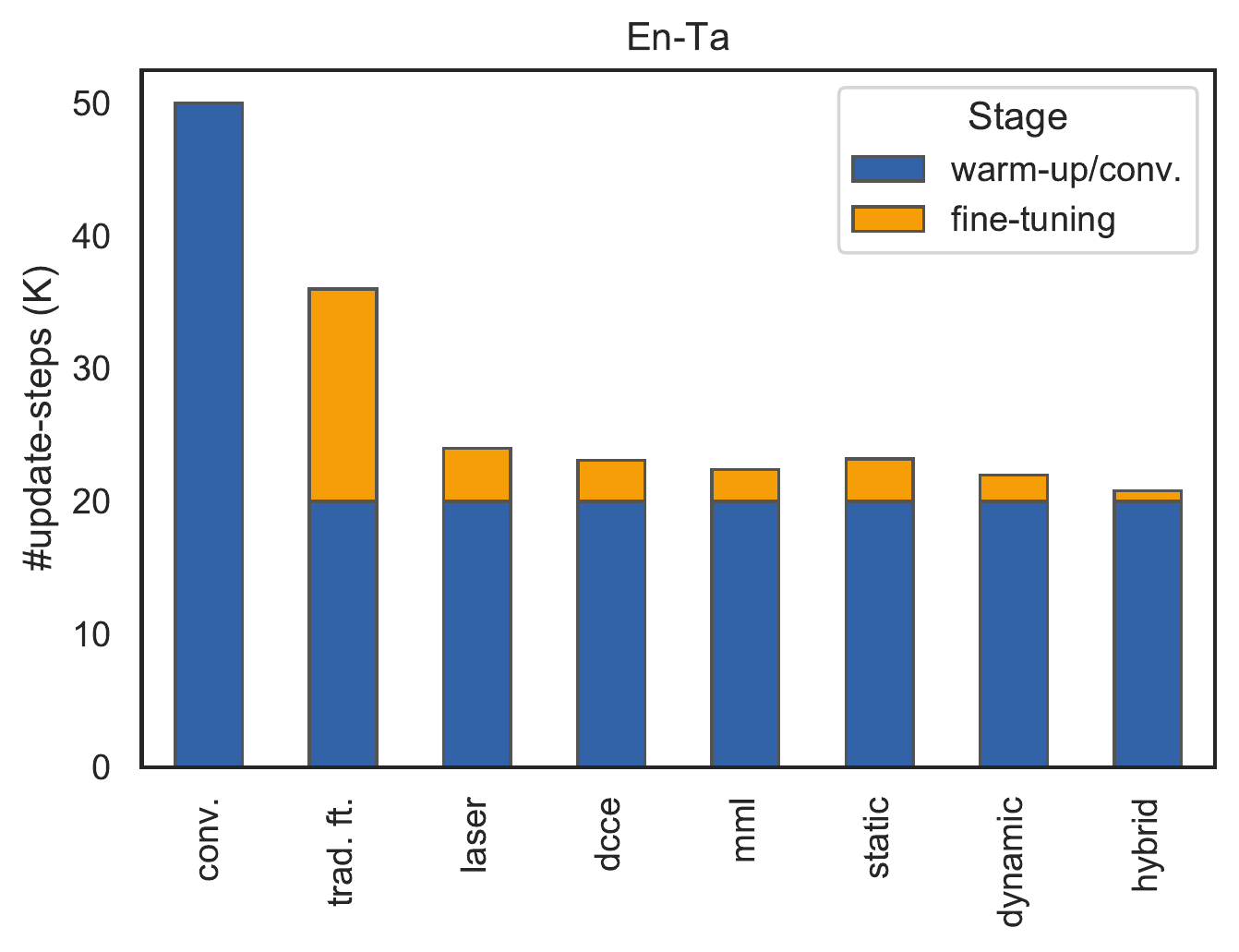}
\end{subfigure}%
\begin{subfigure}{.17\textwidth}
  \centering
  \includegraphics[scale=0.2]{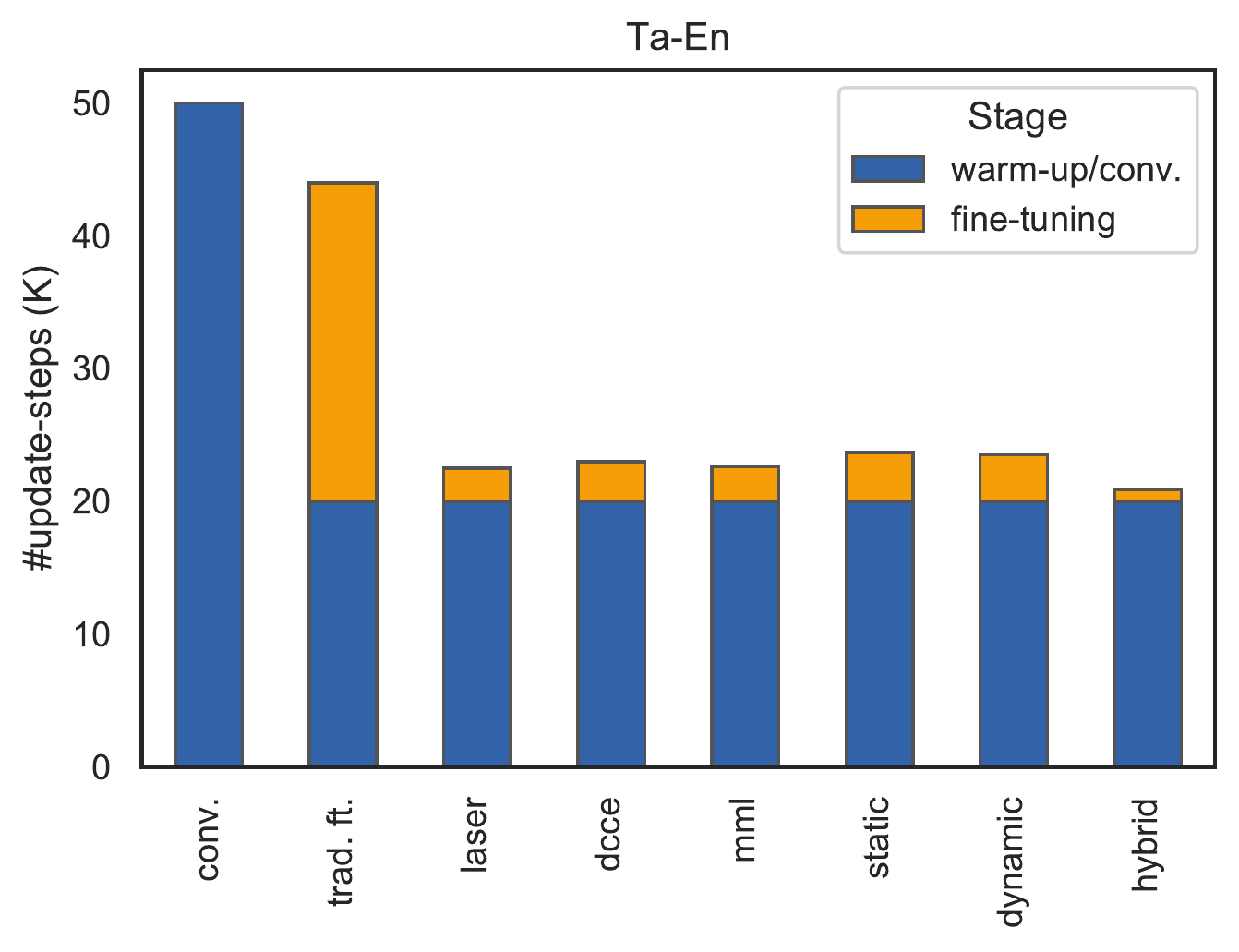}
\end{subfigure}

}

\scalebox{0.99}{
\begin{subfigure}{.17\textwidth}
  \centering
  \includegraphics[scale=0.2]{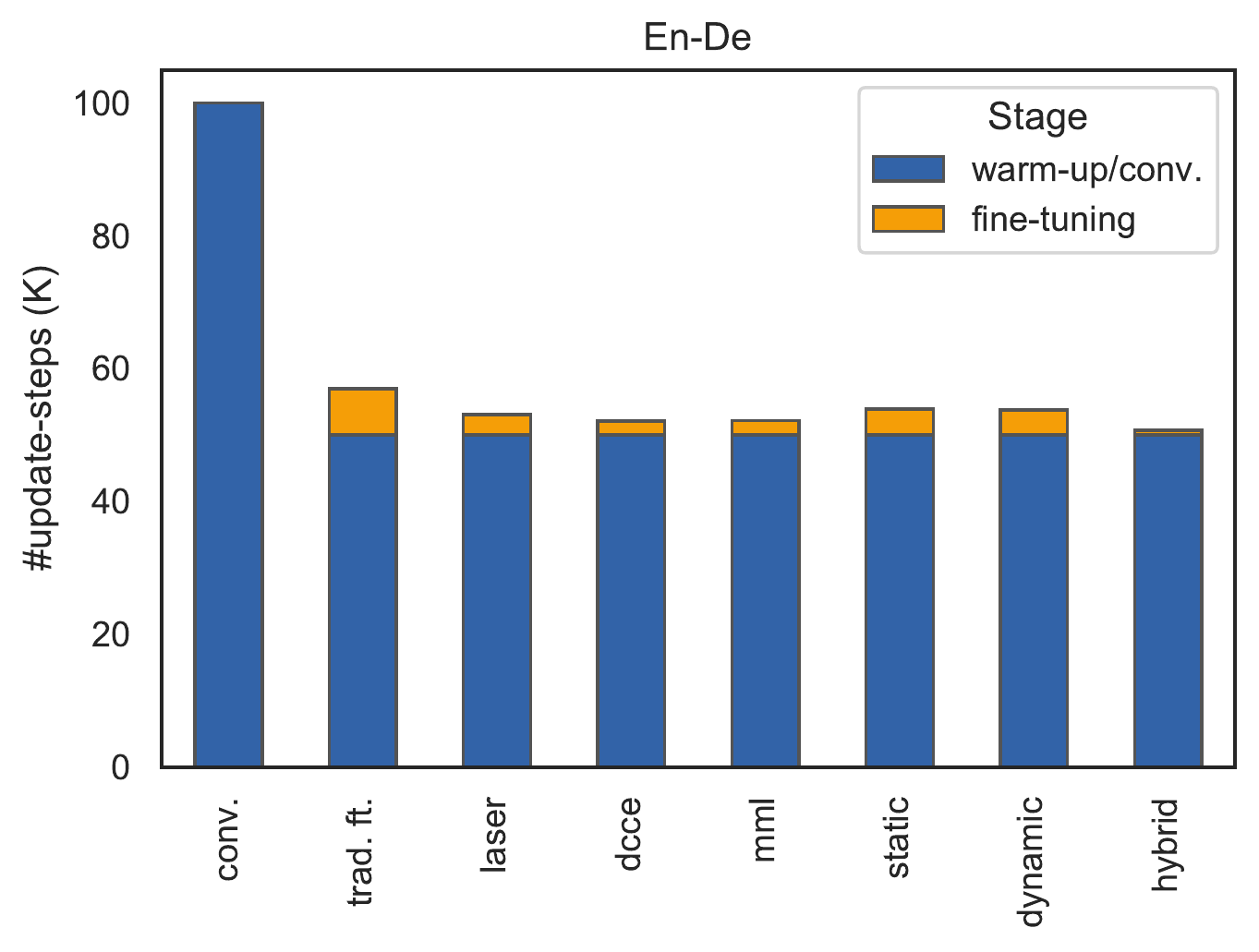}
\end{subfigure}%
\begin{subfigure}{.17\textwidth}
  \centering
  \includegraphics[scale=0.2]{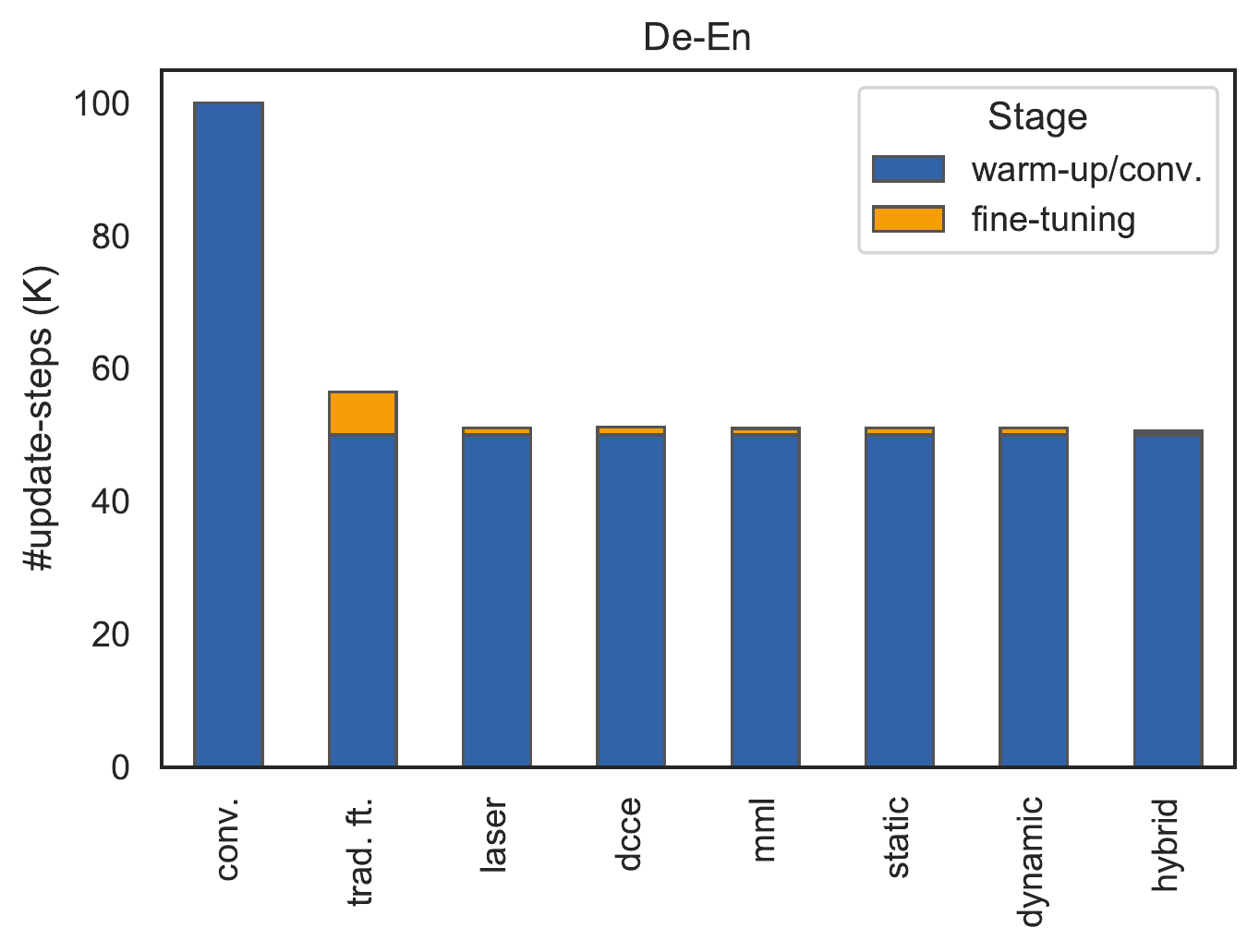}
\end{subfigure}

\begin{subfigure}{.17\textwidth}
  \centering
  \includegraphics[scale=0.2]{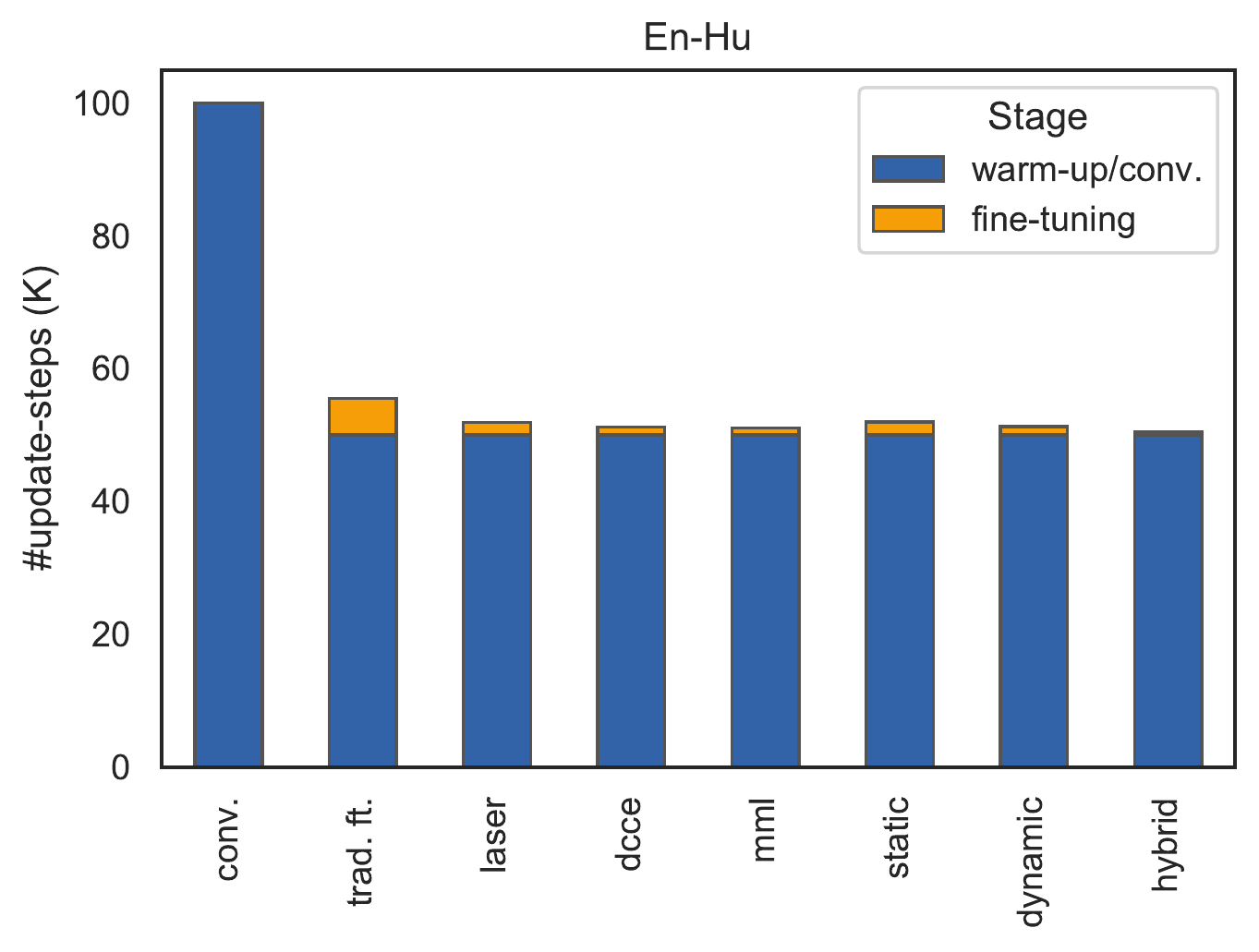}
\end{subfigure}%
\begin{subfigure}{.17\textwidth}
  \centering
  \includegraphics[scale=0.2]{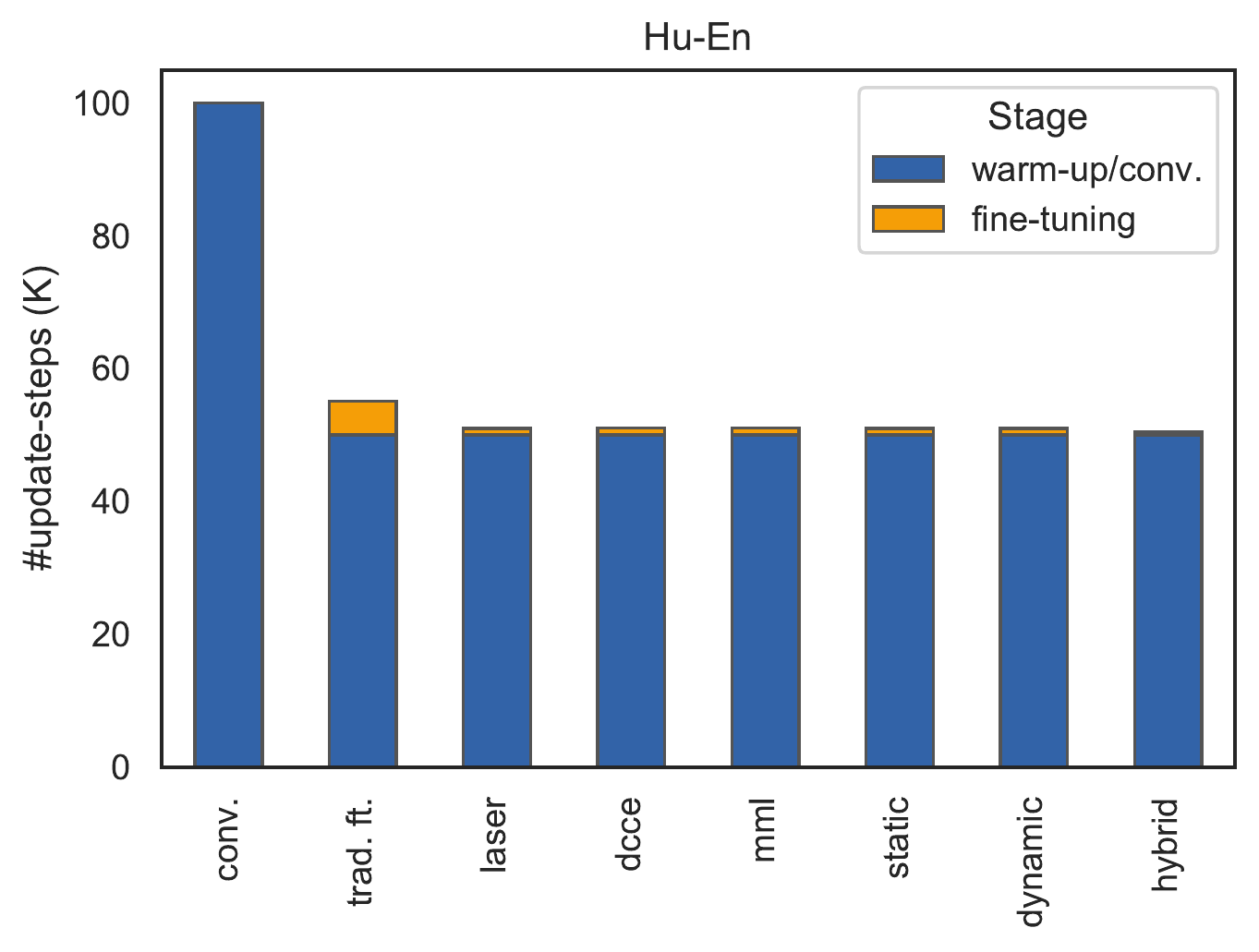}
\end{subfigure}

\begin{subfigure}{.17\textwidth}
  \centering
  \includegraphics[scale=0.2]{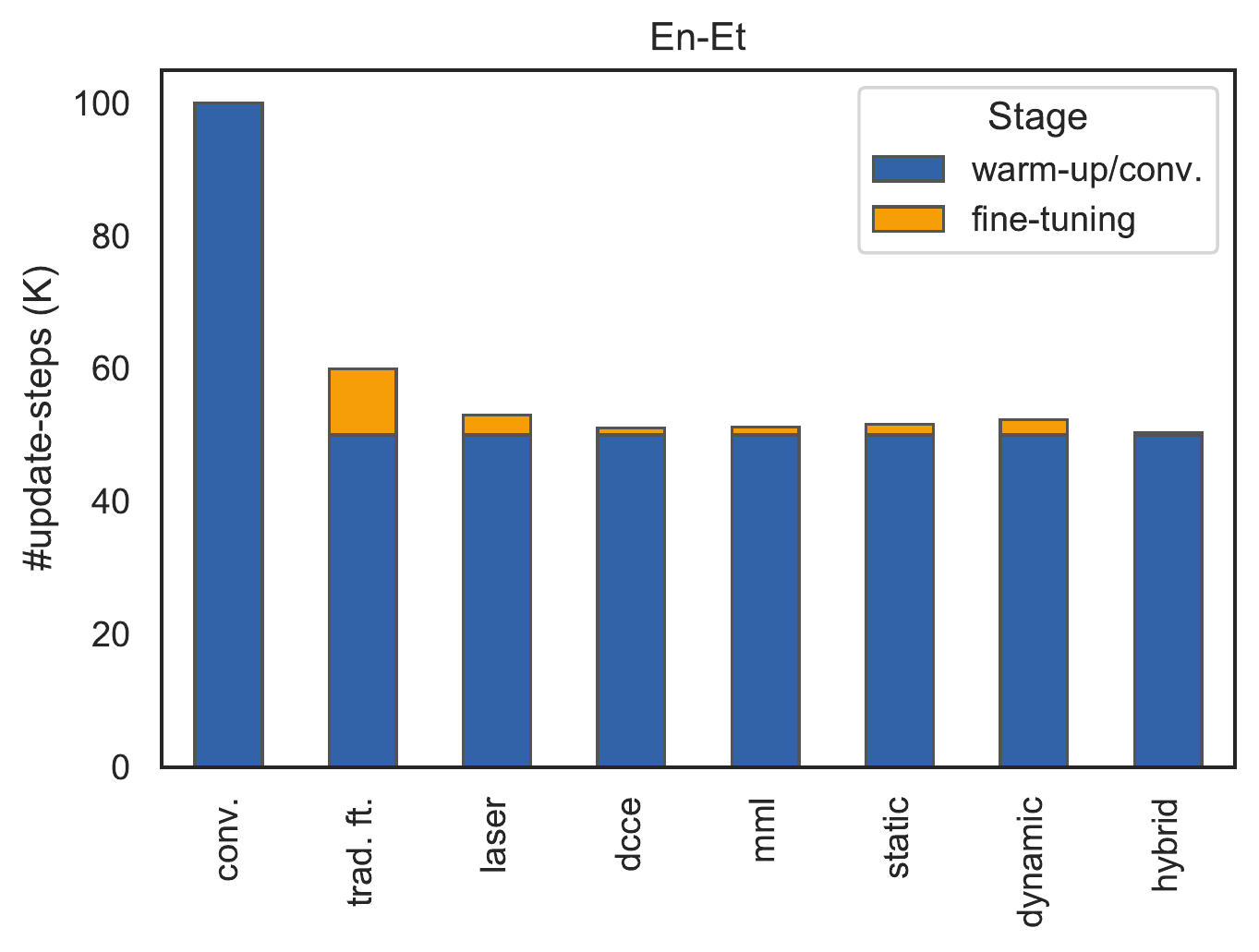}
\end{subfigure}%
\begin{subfigure}{.17\textwidth}
  \centering
  \includegraphics[scale=0.2]{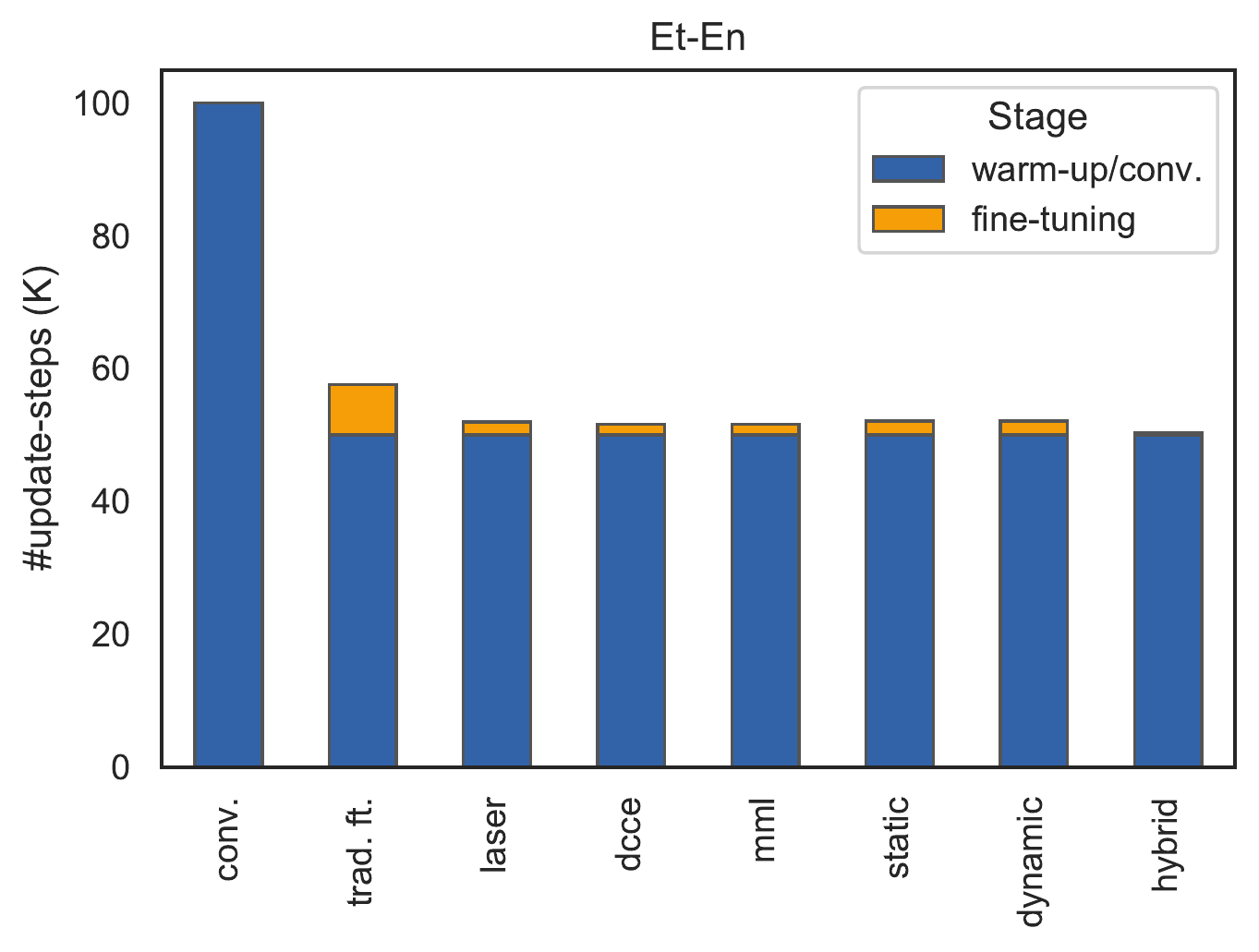}
\end{subfigure}

}
\vspace{-.5em}
\caption{Number of {update steps} required for each setting of Tables \ref{tab:low-res-main-results}, \ref{tab:high-res-main-results}. We keep batch size same in each setting.}
\label{fig:num-updates}
\end{figure*}

\begin{table}[t!]
\centering
\small
\scalebox{0.65}{
\begin{tabular}{lc|cc|cc|cc}
\toprule
\textbf{Scoring} & \textbf{Top} & \multicolumn{2}{c}{\textbf{En-Ha}}  &        \multicolumn{2}{c}{\textbf{En-Ms}} &\multicolumn{2}{c}{\textbf{En-Ta}}
\\

\textbf{Method}& \textbf{data\%} &  \textbf{$\rightarrow$} & \textbf{$\leftarrow$} & \textbf{$\rightarrow$} & \textbf{$\leftarrow$} & 
\textbf{$\rightarrow$} & \textbf{$\leftarrow$}  
\\   
\toprule

\multirow{2}{*}{} 
LASER 
& 10\%    & 14.1\textsubscript{\red{8.3}}  & 17.3\textsubscript{\red{10.1}}  & 30.9\textsubscript{\red{18.9}} & 27.9\textsubscript{\red{15.1}} & 8.1\textsubscript{\red{0.7}} & 15.8\textsubscript{\red{1.6}} \\

& 40\%    & 14.6\textsubscript{\red{13.1}}  & 17.5\textsubscript{\red{16.5}}  & 31.7\textsubscript{\red{30.2}} & 28.2\textsubscript{\red{25.2}} & 8.8\textsubscript{\red{5.9}} & 15.9\textsubscript{\red{10.7}} \\
\midrule

Dual  
& 10\%    &  13.0\textsubscript{\red{1.3}} & 16.3\textsubscript{\red{8.0}}  & 31.0\textsubscript{\red{18.4}} & 28.0\textsubscript{\red{15.5}} & 8.0\textsubscript{\red{0.0}} & 15.2\textsubscript{\red{0.2}} \\

Cond. CE
& 40\%    &  14.3\textsubscript{\red{12.9}} & 16.3\textsubscript{\red{15.3}} & 31.4\textsubscript{\red{29.5}} & 28.2\textsubscript{\red{25.0}} & 8.6\textsubscript{\red{5.3}} & 16.0\textsubscript{\red{11.0}} \\
\midrule

Modified  
& 10\%    & 14.4\textsubscript{\red{5.9}} & 15.1\textsubscript{\red{4.7}} & 31.8\textsubscript{\red{19.6}} & 27.9\textsubscript{\red{15.3}} & 8.5\textsubscript{\red{0.0}} & 15.2\textsubscript{\red{0.6}} \\

Moore-Lewis
& 40\%    & 14.8\textsubscript{\red{13.3}} & 15.6\textsubscript{\red{13.6}} & 31.6\textsubscript{\red{30.8}} & 28.1\textsubscript{\red{24.9}} & 9.0\textsubscript{\red{5.9}} & 15.6\textsubscript{\red{10.5}} \\

\bottomrule
\end{tabular}}
\caption{Results for \textbf{two-stage curriculum training framework vs. training without warm-up stage} on top 10\% and 40\% of selected data ranked by three scoring methods (\S \ref{subsec:deterministic-curriculum}). Main values denote the results of our two-stage framework utilizing warm-up stage, while {subscript} values represent results when model is trained on the same data subset without warm-up stage.  }
\label{tab:low-res-ft-vs-frm-scratch-results} 
\end{table}

\vspace{-.2em}
\subsection{Do We Need the Two Stages?}
\vspace{-.2em}
For the online curricula, we leverage the model $\mathcal{M}$ for selecting $\mathcal{D}_s$ based on the prediction scores, while in the deterministic curricula, we do not use the emerging model for selecting the data subset. One might ask -- \textit{do we need a base model in the deterministic curricula? Can we get rid of the warm-up stage?} To answer these questions, we perform another set of experiments where we train $\mathcal{M}$ from a randomly initialized state on the top $p$\% of the selected data ($p=$\{10, 40\}) ranked by the three bitext scoring methods (\S \ref{subsec:deterministic-curriculum}) and compare the results with our two-stage curriculum training framework where we fine-tune the base model from the warm-up stage on the same data subset. From the results in Table \ref{tab:low-res-ft-vs-frm-scratch-results}, it is evident that our proposed curriculum training framework utilizing the warm-up stage outperforms the approach not using any warm-up stage by a sizable margin in all the tasks.

\vspace{-.2em}
\subsection{Comparing Required Update Steps}
\vspace{-.2em}

Our proposed curriculum training approaches not only exhibit better performance but also converge faster compared to the baseline and traditional fine-tuning method. In Figure \ref{fig:num-updates}, we plot the number of update steps required by each of the settings in Table~\ref{tab:low-res-main-results} and~\ref{tab:high-res-main-results}. On average, we need about 50\% fewer updates compared to the converged model. For high-resource languages, we need much fewer updates in the model fine-tuning stage. For all the language pairs, the hybrid curriculum strategy requires the fewest updates as the size of selected subsets is much lower compared to other approaches.



\begin{table}[t!]
\centering
\small
\scalebox{0.65}{
\begin{tabular}{llc|cc}
\toprule
\textbf{Type} & \textbf{Setting} & \textbf{\%data-used} & \multicolumn{2}{c}{\textbf{En-De}}  
\\

& & \textbf{in each ep.} & \textbf{$\rightarrow$} & \textbf{$\leftarrow$} 
\\   
\toprule
Warm-up Model & All Data    & 100\%     & 33.3 & 39.1  \\
\midrule
Converged Model & All Data  & 100\%     & 34.6 & 40.0  \\
\midrule

\multicolumn{5}{c}{\textit{Warm-Up Model Fine-tuning (Ft.)}} \\

\midrule
\multirow{1}{*}{Traditional Ft.} & All data & 100\%    &  
 34.0 \diffscore{-0.6} & 41.6 \diffscore{+1.6}  
\\

\midrule
\multirow{3}{*}{Det. Curricula} & 
LASER & 40\%      & 
34.4 \diffscore{-0.2} & 43.2 \diffscore{+3.2}  
\\
& 
Dual Cond. CE (\text{DCCE}) & 40\%      & 
\textbf{35.1} \diffscore{+0.5} & \textbf{44.4} \diffscore{+4.4}  
\\

& 
Mod. Moore-Lewis (\text{MML}) & 40\%      & 34.5 \diffscore{-0.1} & 41.6 \diffscore{+1.6}  
\\

\midrule

\multirow{2}{*}{Online Curricula} 
&  Static Window  & 40\%      & 
34.1 \diffscore{-0.5} & 41.9 \diffscore{+1.9}  
\\
&  Dynamic Window  \\
& \quad Expansion & <40\%      & 
34.4 \diffscore{-0.2} & 42.2 \diffscore{+2.2}  
\\
& \quad Shrink & <40\%      &  
34.3 \diffscore{-0.3} & 42.0 \diffscore{+2.0}    
\\

\bottomrule
\end{tabular}}
\vspace{-0.5em}
\caption{Results for En$\leftrightarrow$De on \textbf{noisy ParaCrawl corpus} of 10M bitext pairs. Here, the data-percentage corresponds to all 10M bitext ($\mathcal{D}_g$) and $\mathcal{D}_{d} \coloneqq \mathcal{D}_g$.
Subscript values denote the BLEU score difference from the respective converged model.
} 
\label{tab:ende-paracrawl} 
\end{table}

\vspace{-.2em}
\subsection{Performance on Noisy Data}
\vspace{-.2em}
We further evaluate our framework on noisy data. We randomly selected 10M bitext pairs from the En-De ParaCrawl corpus \cite{banon-etal-2020-paracrawl}. We keep the experimental settings similar to \S \ref{sec:res} and present the results in Table~\ref{tab:ende-paracrawl}. Fine-tuning on the data subset ($\mathcal{D}_s$) selected by \text{DCCE} method outperforms the baseline (\textit{Converged Model}) on both directions with a +4.4 BLEU gain in De-En. All the other deterministic and online curriculum methods perform better than the converged model on the De-En direction with a sizable margin. Compared to the traditional fine-tuning, all the curriculum methods perform better in both En to/from De.

\section{Related Work}

\paragraph{Curriculum Learning}
Inspired by human learners, \citet{ELMAN199371} argues that optimization of neural network training can be accelerated by gradually increasing the difficulty of the concepts. \citet{bengio-ccl} were the first to use the term ``curriculum learning'' to refer to the easy-to-hard training strategies in the context of machine learning. Using an easy-to-hard curriculum based on increasing vocabulary size in language model training, they achieved performance improvement. Recent work \cite{sp-ccl,pmlr-v97-hacohen19a,NEURIPS2020_62000dee} shows that manoeuvring the sequence of training data can improve both training efficiency and model accuracy. Several studies show the effectiveness of the difficulty-based curriculum learning in a wide range of NLP tasks including task-specific word representation learning \cite{tsvetkov-etal-2016-learning}, natural language understanding tasks \cite{sachan-xing-2016-easy, xu-etal-2020-curriculum}, reading comprehension \cite{tay-etal-2019-simple}, and language modeling \cite{campos_21}. 

\paragraph{Curriculum Learning in NMT}
The difficulty-based curriculum in NMT was first explored by \citet{kocmi-bojar-2017-curriculum}. Later, \citet{zhang2018empirical} adopt a probabilistic view of curriculum learning and investigate a variety of difficulty criteria based on human intuition, \textit{e.g.,} sentence length and word rarity. \citet{platanios-etal-2019-competence} connect the appearance of difficult samples with NMT model competence. \citet{liu-etal-2020-norm} propose a norm-based curriculum learning method based on the norm of word embedding. \citet{zhou-etal-2020-uncertainty} use a pre-trained language model to measure the word-level uncertainty. \citet{zhan2021metacl} propose meta-curriculum learning for domain adaptation in NMT. Most curriculum learning methods in NMT focus on addressing the batch selection issue from the beginning of the training by using hand-designed heuristics \cite{Zhao_Wu_Niu_Wang_2020}. In contrast, our proposed two-stage curriculum training framework for NMT fine-tunes the base model from the warm-up stage on \textit{selected} subsets of data. Our curriculum training framework is more realistic, resembling the formal education system as discussed in \S \ref{subsec:are-all-data-useful-always}.



\paragraph{Self-paced Learning in NMT} Here,
the model itself measures the difficulty of the training samples to adjust the learning pace \cite{NIPS2010_e57c6b95}. \citet{wan-etal-2020-self} first train the NMT model for $M$ passes on the data and cache the translation probabilities to find the variance. The lower variance of the translation probabilities of a sample reflects higher confidence. Later, they use the confidence scores as factors to weight the loss to control the model updates. For low-resource NMT, \citet{xu-etal-2020-dynamic} utilize the declination of the loss of a sample as the difficulty measure and train the model on easier samples (higher loss drop). In our online curriculum, we leverage the prediction scores of the emerging model in the model fine-tuning stage. However, after ranking the samples based on the prediction scores, we employ a \textit{variety} of {data-selection methods} to select the better data subset (\S\ref{subsec:online}).  


\paragraph{Domain Specific Fine-tuning in NMT}
{Here, converged NMT models trained on large general-domain parallel data are fine-tuned on in-domain data  \cite{luong-manning-2015-stanford,zoph-etal-2016-transfer, mfreitag-et-al}. In contrast, in our framework, we adapt a base NMT model (non-converged) on \textit{selected} subsets of the in-domain data considering the data usefulness and quality.}


\paragraph{Data-selection Strategy in NMT}
We apply the data-selection curriculum in the model fine-tuning stage in our framework. \citet{joty-etal-2015-avoid} use domain adaptation by penalizing sequences similar to the out-domain data in their training data-selection. \citet{wang-etal-2018-denoising} propose a curriculum-based data-selection strategy by using an \textit{additional} trusted clean dataset to calculate the noise level of a sample. \citet{kumar-etal-2019-reinforcement} use reinforcement learning to learn a denoising curriculum jointly with the NMT system. \citet{jiao-etal-2020-data} identify the inactive samples during training and re-label them for later use. \citet{xinyi-wang-glmask}  find gradient alignments between a clean dataset and the training data to mask out noisy data.




\section{Conclusion}
We have presented a two-stage curriculum training framework for NMT where we apply a data-selection curriculum in the model fine-tuning stage. Our novel online curriculum strategy utilizes the emerging models' prediction scores for the selection of a better data subset. Experiments on six low- and high-resource language pairs show the efficacy of our proposed framework. Our curriculum training approaches exhibit better performance as well as converge much faster by requiring fewer updates compared to the baselines.

\bibliography{anthology}
\bibliographystyle{acl_natbib}

\clearpage
\appendix

\section*{Appendix}


\section{Model Architecture Settings}
\label{app:arch-settings}


For En$\leftrightarrow$Ha, we use a smaller Transformer architecture with five layers, while for the other language pairs we use larger Transformer architecture with six encoder and decoder layers. We present the number of attention heads, embedding dimension, and the inner-layer dimension of both settings in Table \ref{tab:arch-setting}. 

\begin{table}[H]
\centering
\scalebox{0.9}{
\begin{tabular}{l|c c}
\toprule
{\textbf{Settings}} & \textbf{En$\leftrightarrow$Ha} & \textbf{Other Pairs} \\
\midrule
Transformer Layers & 5 & 6 \\
\#Attention Heads & 8 & 16 \\
Embedding Dimension & 512 & 1024 \\
Inner-layer Dimension & 2048 & 4096 \\
 
\bottomrule
\end{tabular}
}
\caption{Model architecture settings.}
\label{tab:arch-setting}
\end{table}


\section{Variety of Data Samples in Static Data-selection Window}
\label{app:illust-static}

In the beginning of each epoch in the \textit{model fine-tuning} stage in static data-selection window approach (\S\ref{subsec:online}), we rank $\mathcal{D}_d$ based on the prediction scores of each sentence pair $(x_i, y_i)$. We then pick a \textit{fixed} data-selection window (confined to a range of data-percentage in the ranked $\mathcal{D}_d$ \textit{e.g.,} 30\% to 70\%) by discarding too easy and too hard/noisy samples. Even though in this approach the size of the selected data subset ($\mathcal{D}_s$) remains the {same} throughout the model fine-tuning stage, the samples in $\mathcal{D}_s$ \textit{changes} from epoch-to-epoch due to the change in their prediction scores by the current model. We present an illustrative example of this phenomenon in Figure \ref{fig:illust-static}. 

In the \textit{current} epoch of the fine-tuning stage (Figure \ref{fig:illust-static}(a)), samples 2, 3, 4, and 5 are selected to train the model while samples 1 and 6 are discarded -- 1 is too hard/noisy and 6 is too easy for the current model. In the \textit{next} epoch (Figure \ref{fig:illust-static}(b)), some samples might be selected again (samples 3 and 5), while some earlier selected samples might have lower prediction scores and not be selected due to the hardness to the current model (sample 2). Again, some previously selected samples might have higher prediction scores and not be selected due to the easiness (sample 4). And some samples not selected in the previous epoch can now be selected (samples 1 and 6).

\begin{figure*}[t!]
\centering
\scalebox{0.99}{
\begin{subfigure}{.49\textwidth}
  \centering
  \includegraphics[scale=0.16]{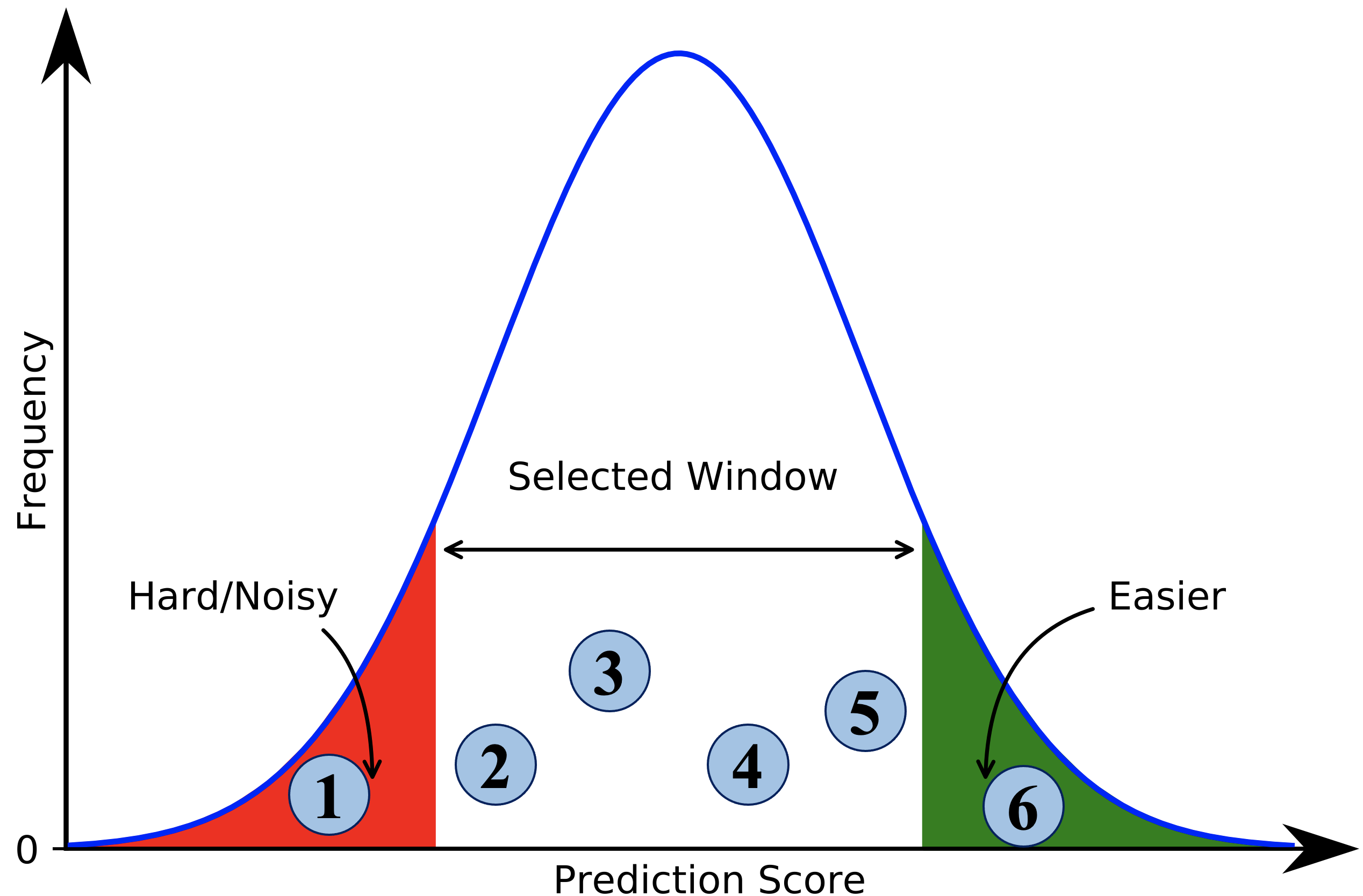}
  \caption{Current epoch.}
\end{subfigure}%
\begin{subfigure}{.49\textwidth}
  \centering
  \includegraphics[scale=0.16]{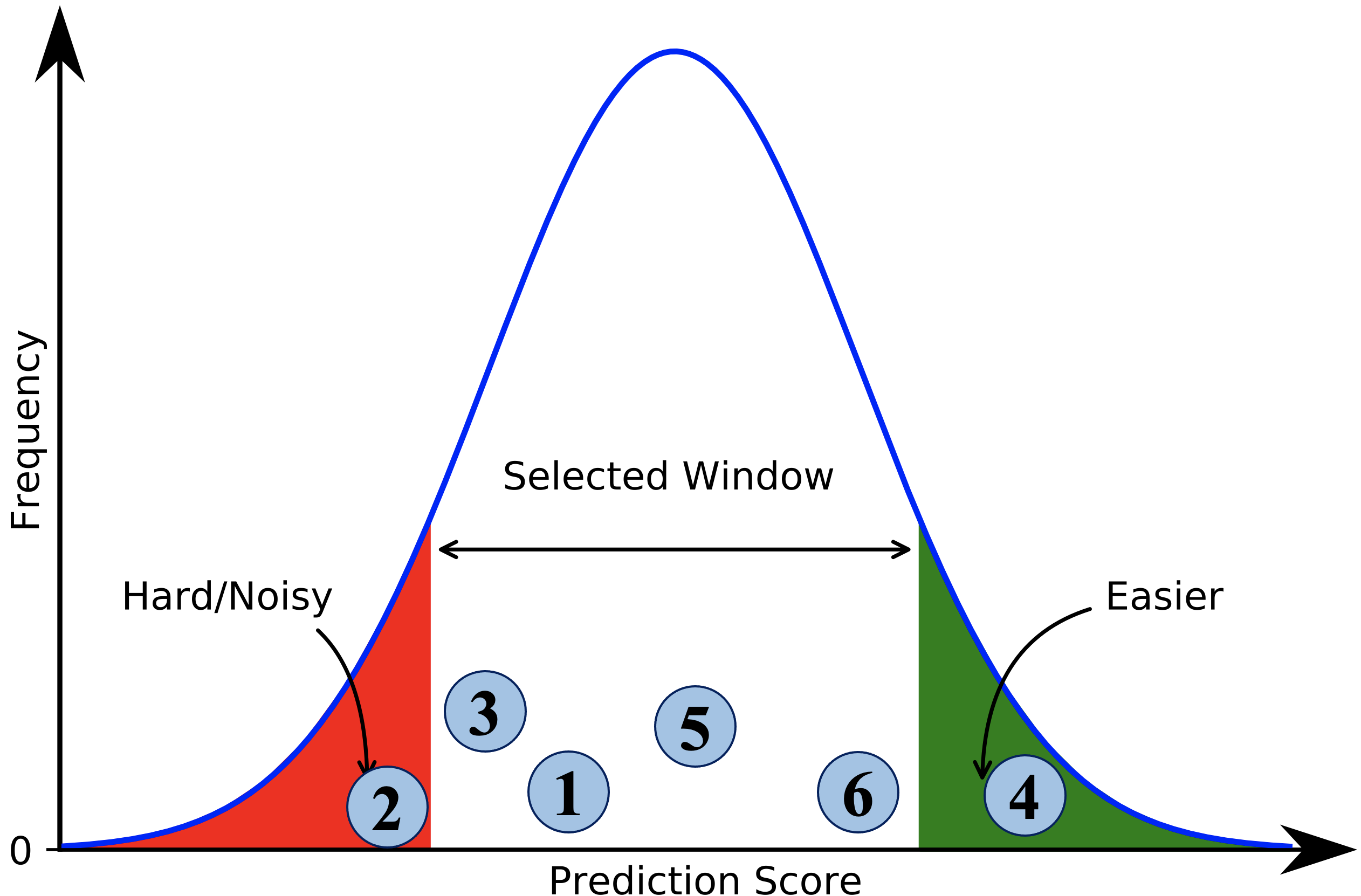}
  \caption{Next epoch.}
\end{subfigure}

}

\caption{Illustrative example of how data samples varies in \textit{static data-selection window} approach in online curriculum. Even though the size of data-selection window is fixed throughout the model fine-tuning stage, the samples in the selected subsets {vary} from epoch-to-epoch due to the change in their prediction scores by the emerging model.}
\label{fig:illust-static}
\end{figure*}

\section{Schedulers in Dynamic Data-selection Window}
\label{app:sched-dynamic}

To change the data-selection window size in dynamic approach, we use schedulers which controls how the size of the window will grow in subsequent epochs (\S\ref{subsec:online}). Apart from the linear scheduler (Eq. \ref{eq:lin-scheduler}), we also experiment with two other schedulers:

\paragraph{$\bullet$ Exponential Scheduler} We find the data-selection window size (for window expansion and shrink) at epoch $t$ using the following formula: 

\begin{equation}
\small
\begin{split}
    \lambda_{\text{exp}}^{expn}(t) &= 
    \begin{cases}
    \lambda_{init} * {E^{t}_{inc}},& \text{if } \lambda_{\text{exp}}^{expn}(t) < \lambda_{max}\\
    \lambda_{max},              & \text{otherwise}
    \end{cases} \\
    \lambda_{\text{shr}}^{expn}(t) &= 
    \begin{cases}
    \lambda_{init} * {E^{-t}_{dec}},& \text{if } \lambda_{\text{shr}}^{expn}(t) > \lambda_{min}\\
    \lambda_{min},              & \text{otherwise}
    \end{cases}
\end{split}
\label{eq:expn-scheduler}
\end{equation}
\normalsize
\noindent Where $\lambda_{init}$ is the initial window size which is smaller for \textit{expansion} and larger for \textit{shrink}, and ${E_{inc}}$, ${E_{dec}}$ are the hyperparameters of the exponential schedulers.

\paragraph{$\bullet$ Square-Root Scheduler} We find the data-selection window size (for window expansion and shrink) at epoch $t$ using the following formula: 

\begin{equation}
\small
\begin{split}
    \lambda_{\text{exp}}^{sqrt}(t) &= 
    \begin{cases}
    \sqrt{\lambda^2_{init} + (C_1 - \lambda^2_{init})  * \frac{t}{S_{inc}}},& \text{if } \lambda_{\text{exp}}^{sqrt}(t) < \lambda_{max}\\
    \lambda_{max},              & \text{otherwise}
    \end{cases} \\
    \lambda_{\text{shr}}^{sqrt}(t) &= 
    \begin{cases}
    \sqrt{\lambda^2_{init} + (C_2 - \lambda^2_{init})  * \frac{t}{S_{dec}}},& \text{if } \lambda_{\text{shr}}^{sqrt}(t) > \lambda_{min}\\
    \lambda_{min},              & \text{otherwise}
    \end{cases}
\end{split}
\label{eq:sqrt-scheduler}
\end{equation}
\normalsize
\noindent Where $\lambda_{init}$ is the initial window size which is smaller for \textit{expansion} and larger for \textit{shrink}, and $C_1$, $C_2$, ${S_{inc}}$, ${S_{dec}}$ are the hyperparameters of the square-root schedulers.



In our initial experiments, we explore the three schedulers --- linear, exponential, and square-root. We found that linear scheduler performs better compared to the other schedulers. We present the results in Table \ref{tab:schedulers-dyn-window-results}.


\begin{table}[t!]
\centering
\scalebox{1}{
\begin{tabular}{l|cc|cc}
\toprule
\textbf{Scheduler} &  \multicolumn{2}{c}{\textbf{En-Ha}}  &        \multicolumn{2}{c}{\textbf{En-De}}
\\

&  \textbf{$\rightarrow$} & \textbf{$\leftarrow$} & \textbf{$\rightarrow$} & \textbf{$\leftarrow$}  
\\   
\toprule

Linear & \textbf{14.9} & \textbf{16.6} & \textbf{37.3} & \textbf{41.6} \\
Exponential & 14.5 & 16.3 & 36.7 & 41.0 \\
Square-Root & 14.4 & 16.0 & 36.9 & 40.9 \\

\bottomrule
\end{tabular}}
\caption{Results for En$\leftrightarrow$\{Ha, De\} for online curriculum with dynamic data-selection window (expansion) using different schedulers. We keep $\lambda_{init}$ the same. }
\label{tab:schedulers-dyn-window-results} 
\end{table}

\section{Overlap of Selected Data Subsets}
\label{app:overlap_analysis}

We compare the data percentage overlap of the ranked data between any two methods of \S \ref{subsec:deterministic-curriculum} in Figure \ref{fig:det-overlap}. From the plots, we see that the overlaps between the data subsets are quite low. Let us consider En-De for an example: if we take the top 40\% data ranked by both \text{LASER} and \text{DCCE} methods, the overlap between these two subsets is 47\%. Nevertheless, both of the subsets perform pretty well compared to the converged model and traditional fine-tuned model (Table \ref{tab:high-res-main-results}). We observe the similar phenomena in almost all the cases (Figure \ref{fig:det-cur-diff-perc}, \ref{fig:det-overlap}). These observations suggest that there can be multiple subsets of data for each language pair, fine-tuning the base model on which exhibits better performance compared to the traditional fine-tuning that uses all the data.





\begin{figure*}[t!]
\centering
\begin{subfigure}{.33\textwidth}
  \centering
  \includegraphics[width=1\linewidth]{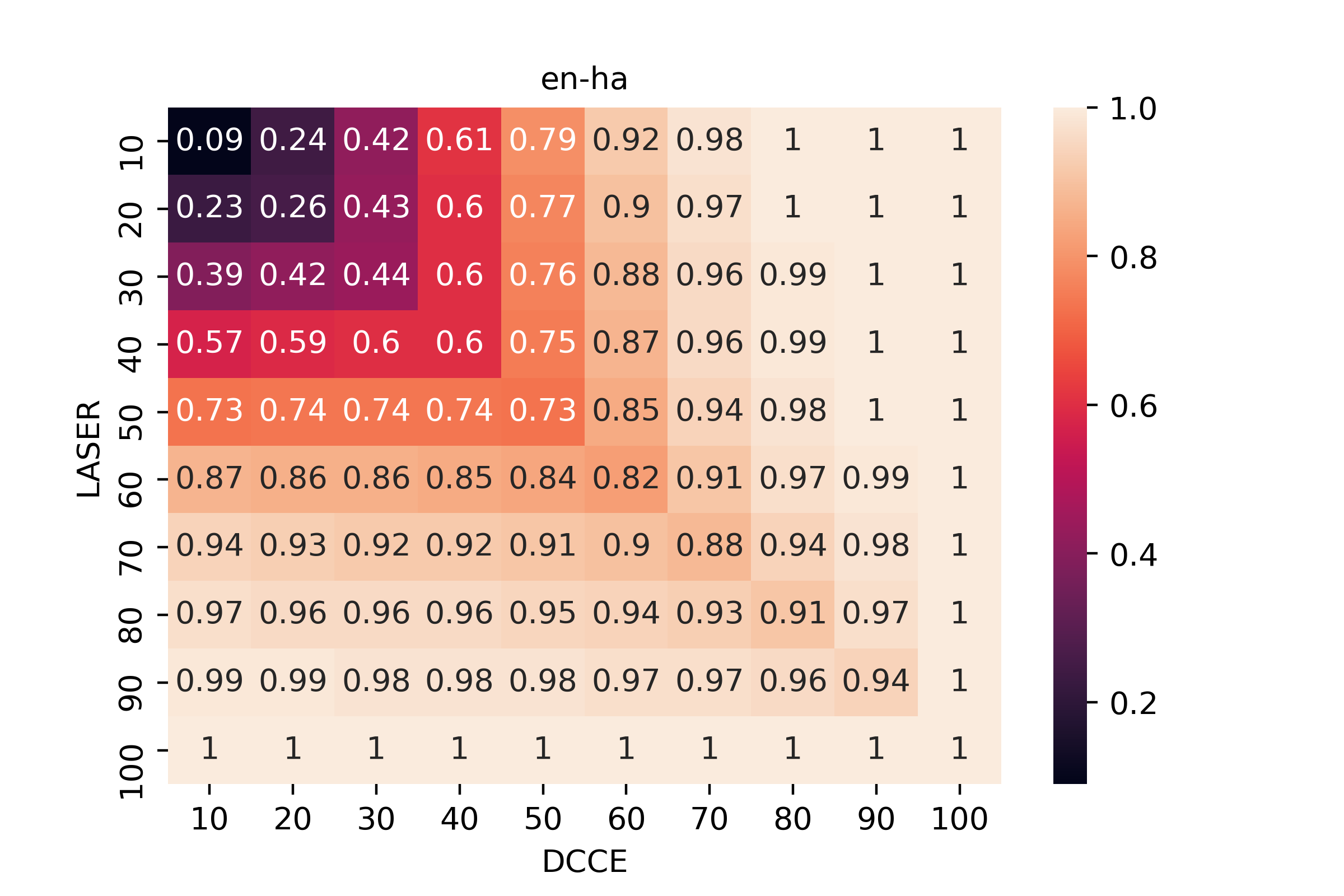}
\end{subfigure}%
\begin{subfigure}{.33\textwidth}
  \centering
  \includegraphics[width=1\linewidth]{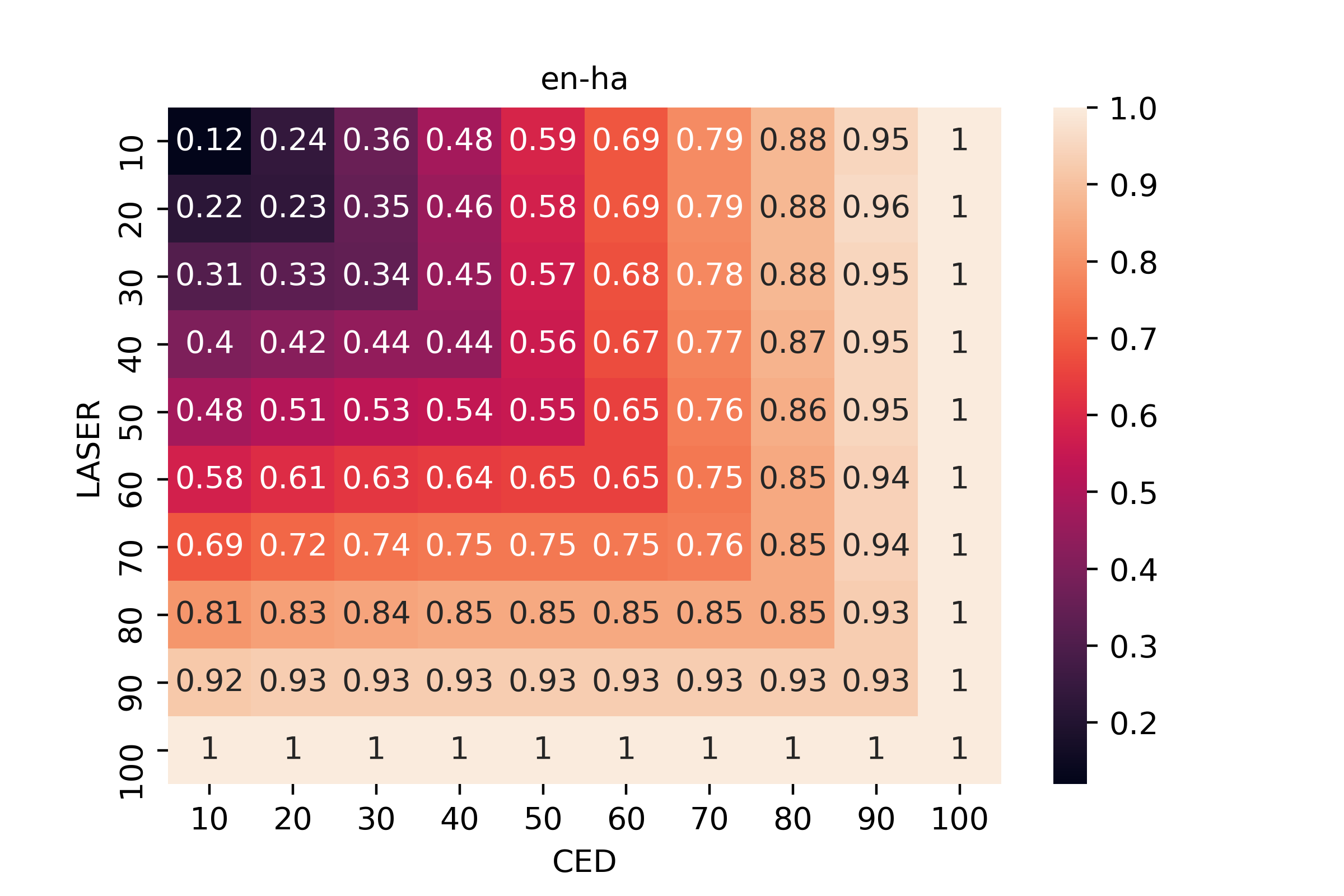}
\end{subfigure}
\begin{subfigure}{.33\textwidth}
  \centering
  \includegraphics[width=1\linewidth]{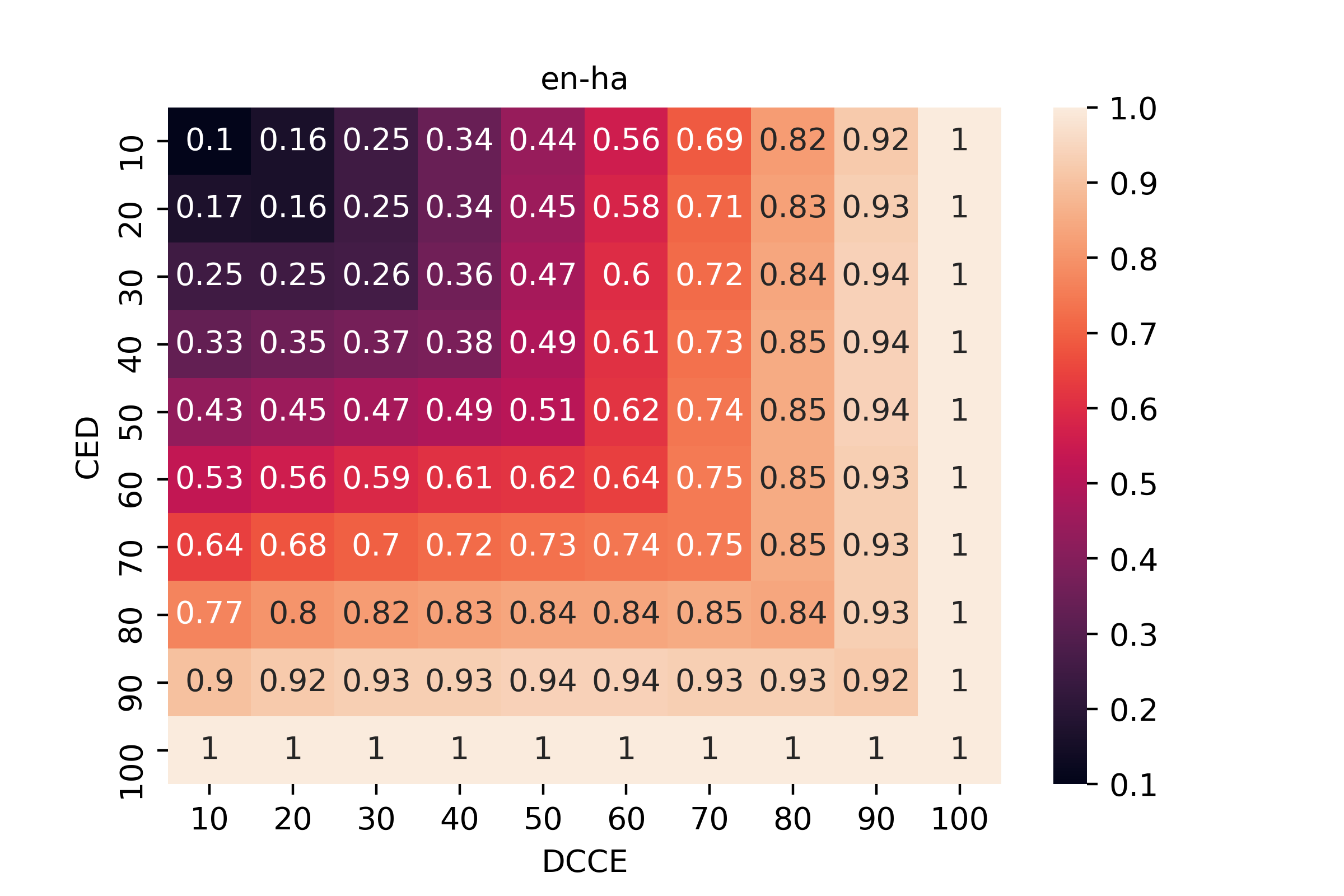}
\end{subfigure}

\begin{subfigure}{.33\textwidth}
  \centering
  \includegraphics[width=1\linewidth]{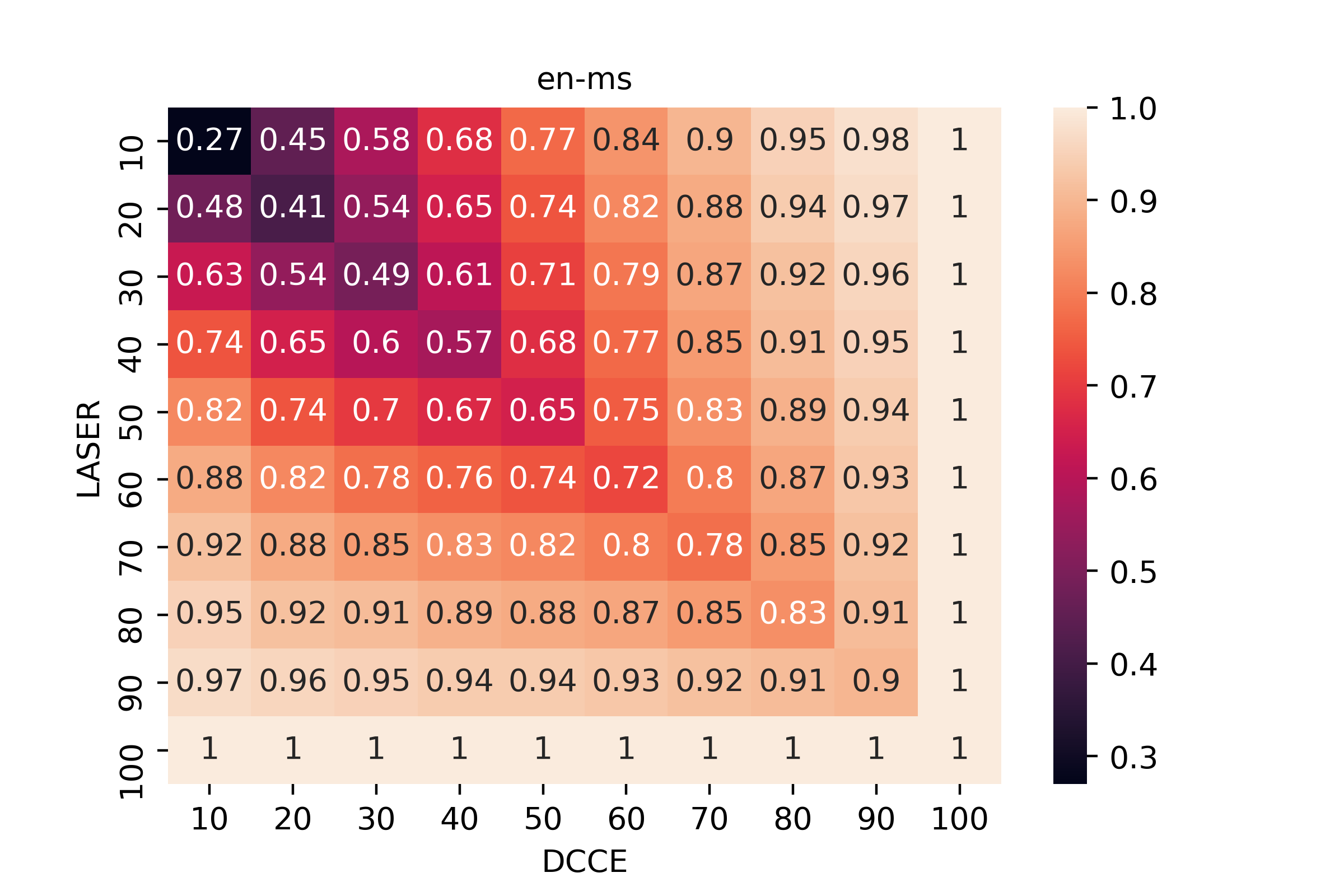}
\end{subfigure}%
\begin{subfigure}{.33\textwidth}
  \centering
  \includegraphics[width=1\linewidth]{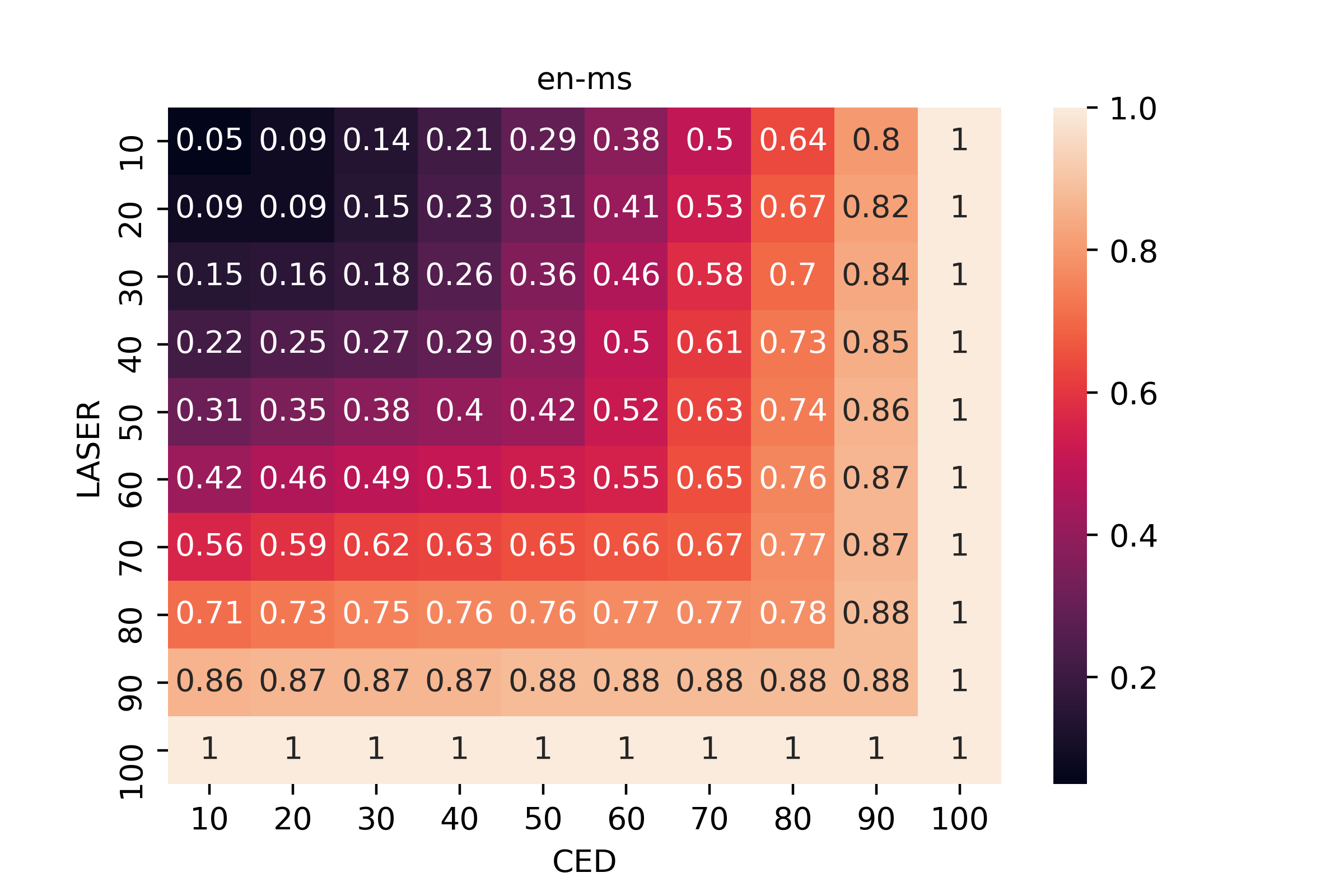}
\end{subfigure}
\begin{subfigure}{.33\textwidth}
  \centering
  \includegraphics[width=1\linewidth]{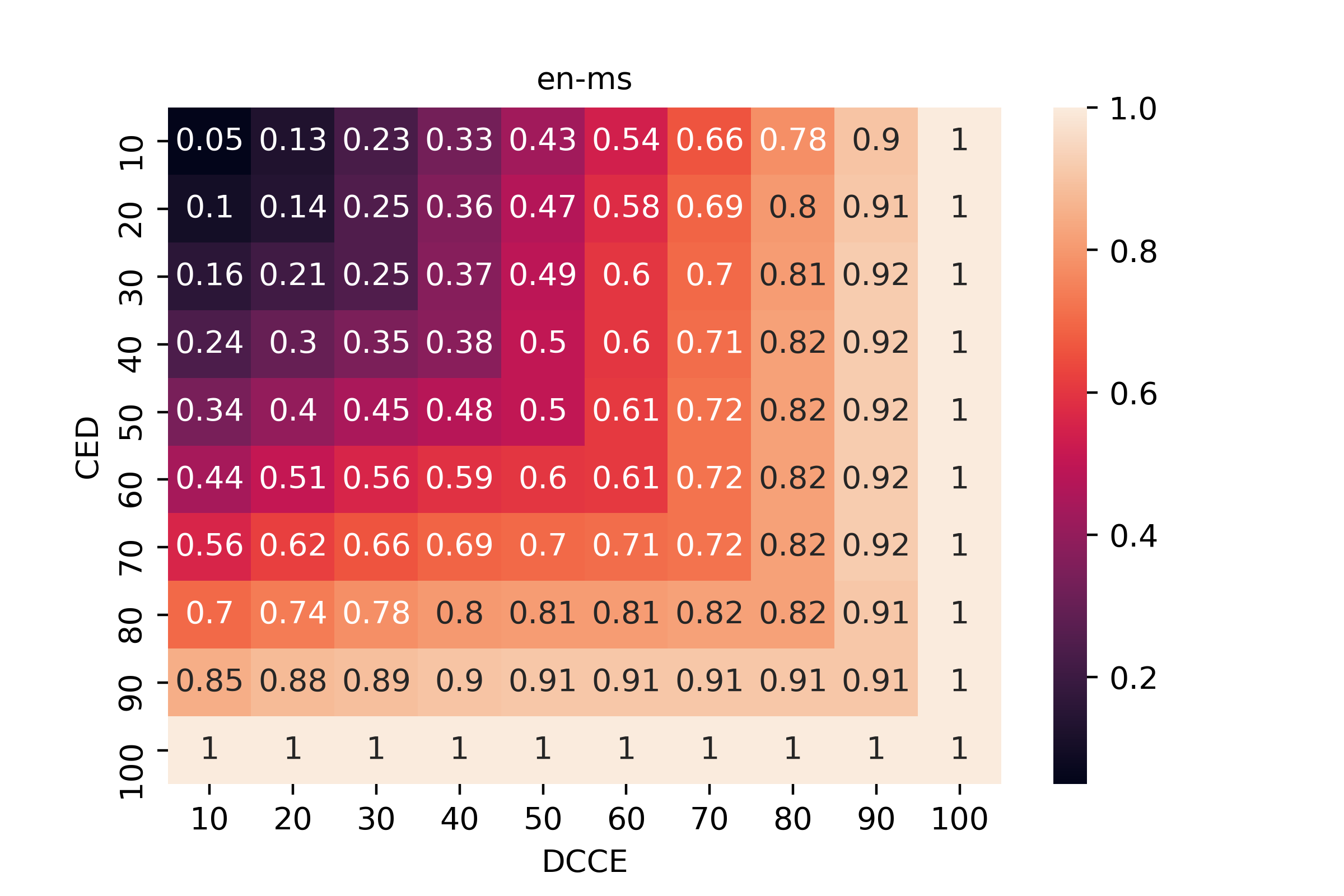}
\end{subfigure}

\begin{subfigure}{.33\textwidth}
  \centering
  \includegraphics[width=1\linewidth]{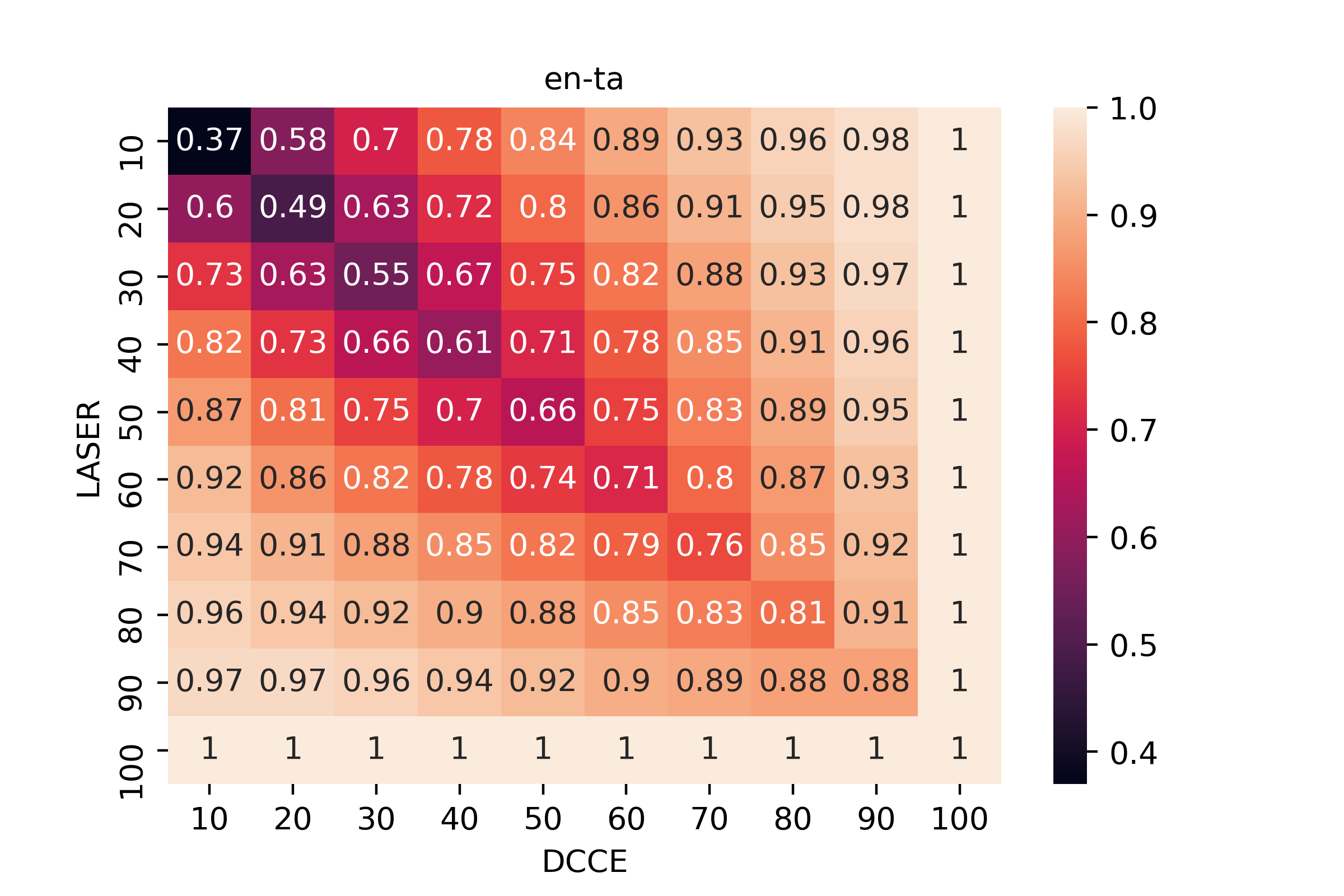}
\end{subfigure}%
\begin{subfigure}{.33\textwidth}
  \centering
  \includegraphics[width=1\linewidth]{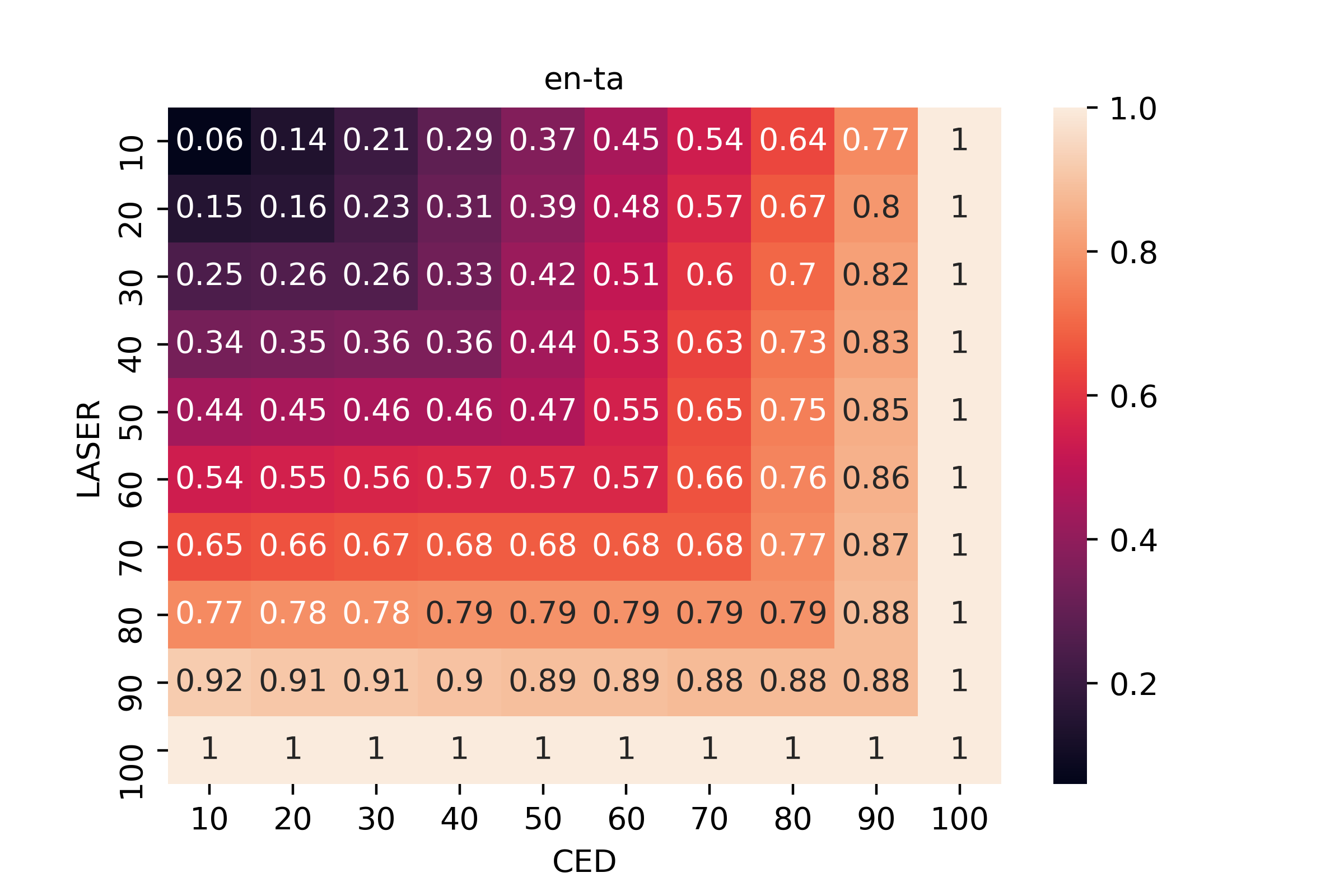}
\end{subfigure}
\begin{subfigure}{.33\textwidth}
  \centering
  \includegraphics[width=1\linewidth]{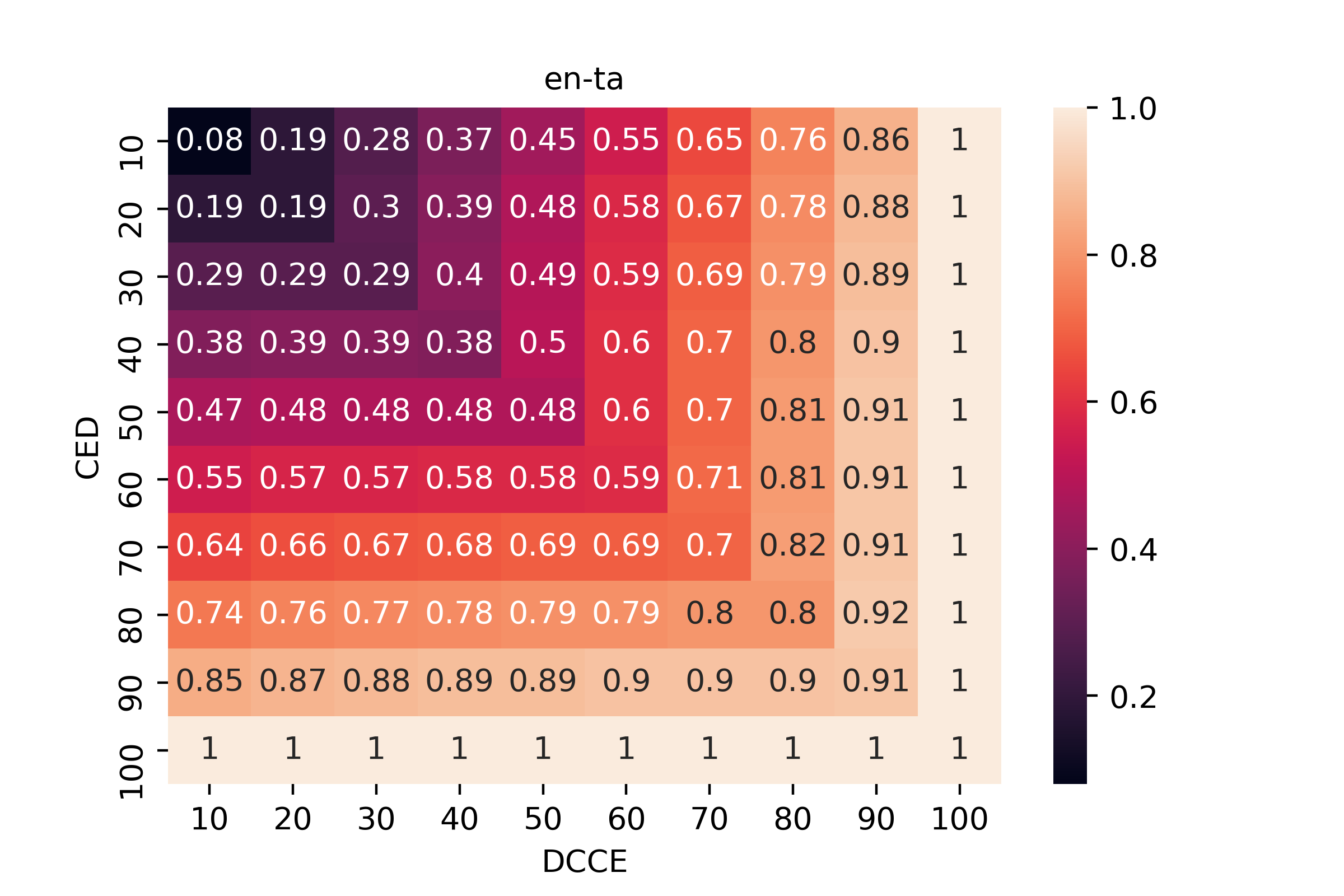}
\end{subfigure}

\begin{subfigure}{.33\textwidth}
  \centering
  \includegraphics[width=1\linewidth]{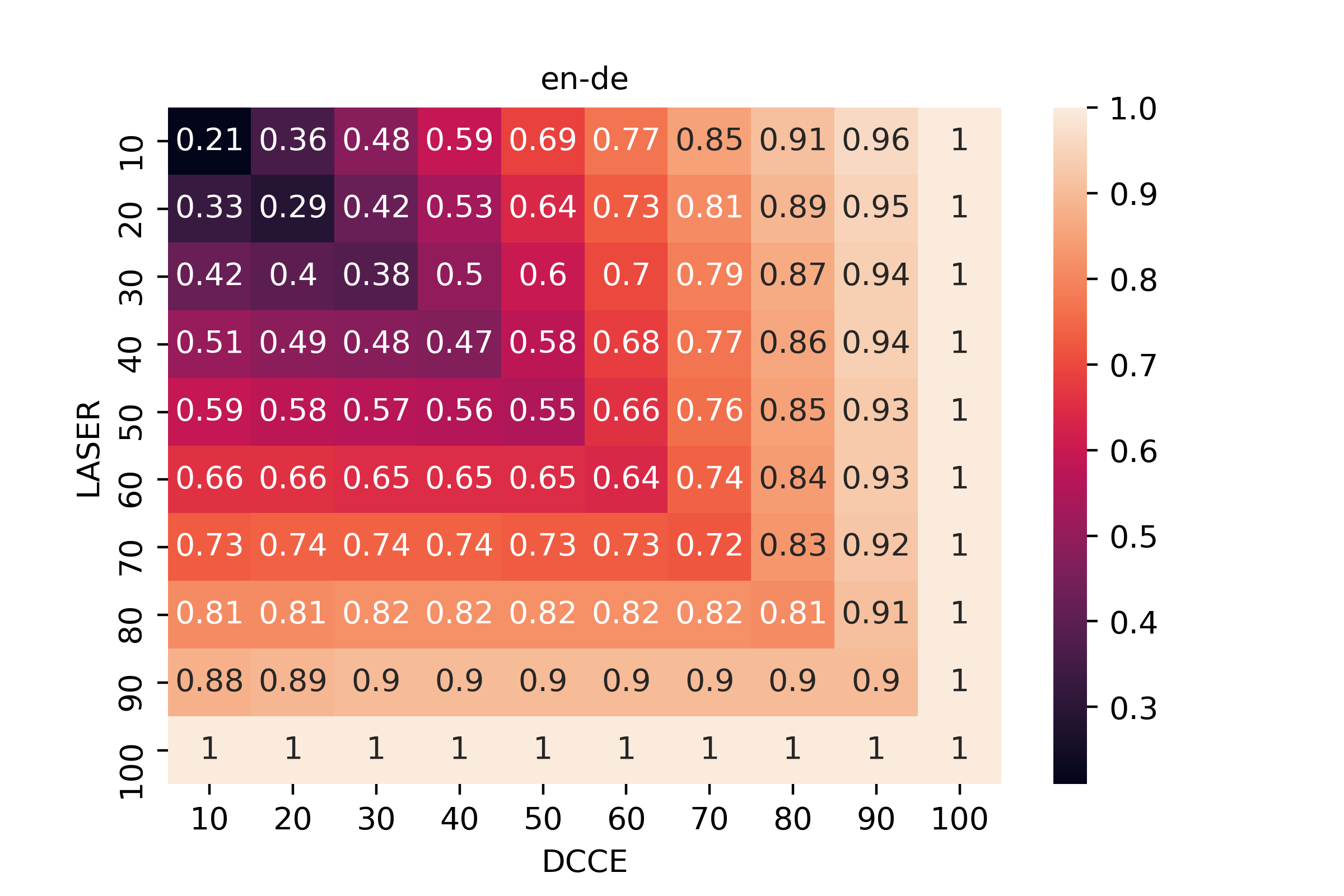}
\end{subfigure}%
\begin{subfigure}{.33\textwidth}
  \centering
  \includegraphics[width=1\linewidth]{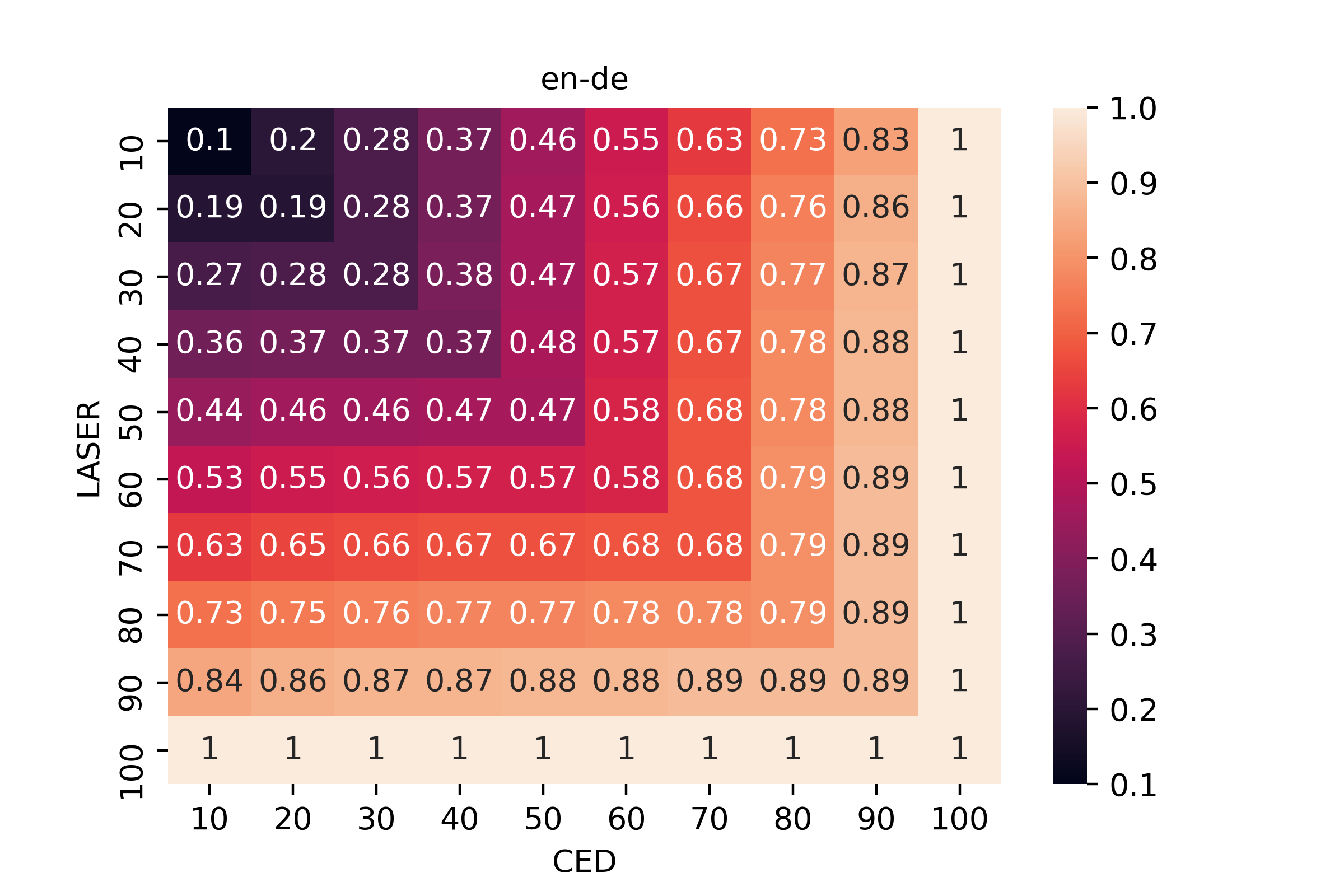}
\end{subfigure}
\begin{subfigure}{.33\textwidth}
  \centering
  \includegraphics[width=1\linewidth]{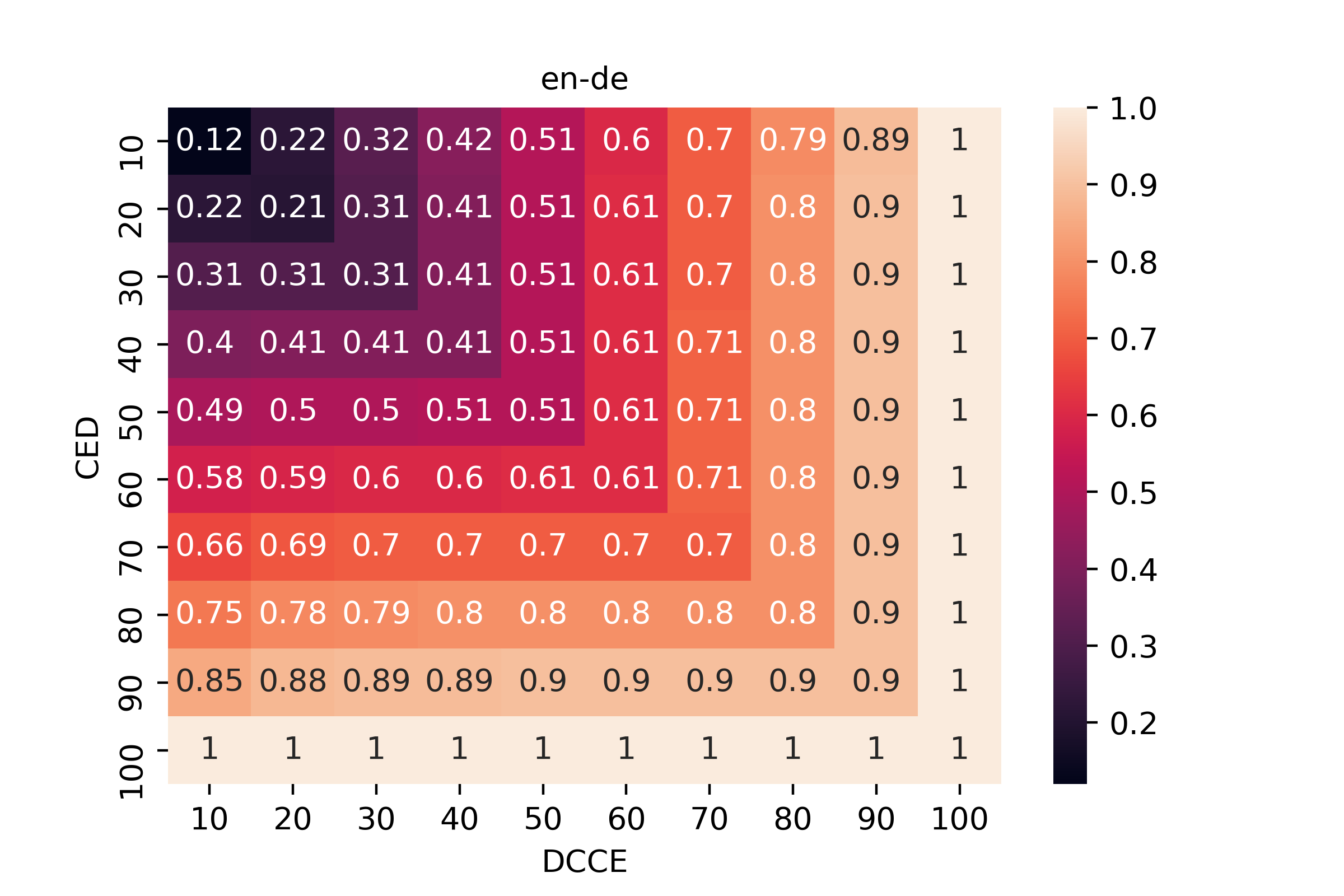}
\end{subfigure}

\begin{subfigure}{.33\textwidth}
  \centering
  \includegraphics[width=1\linewidth]{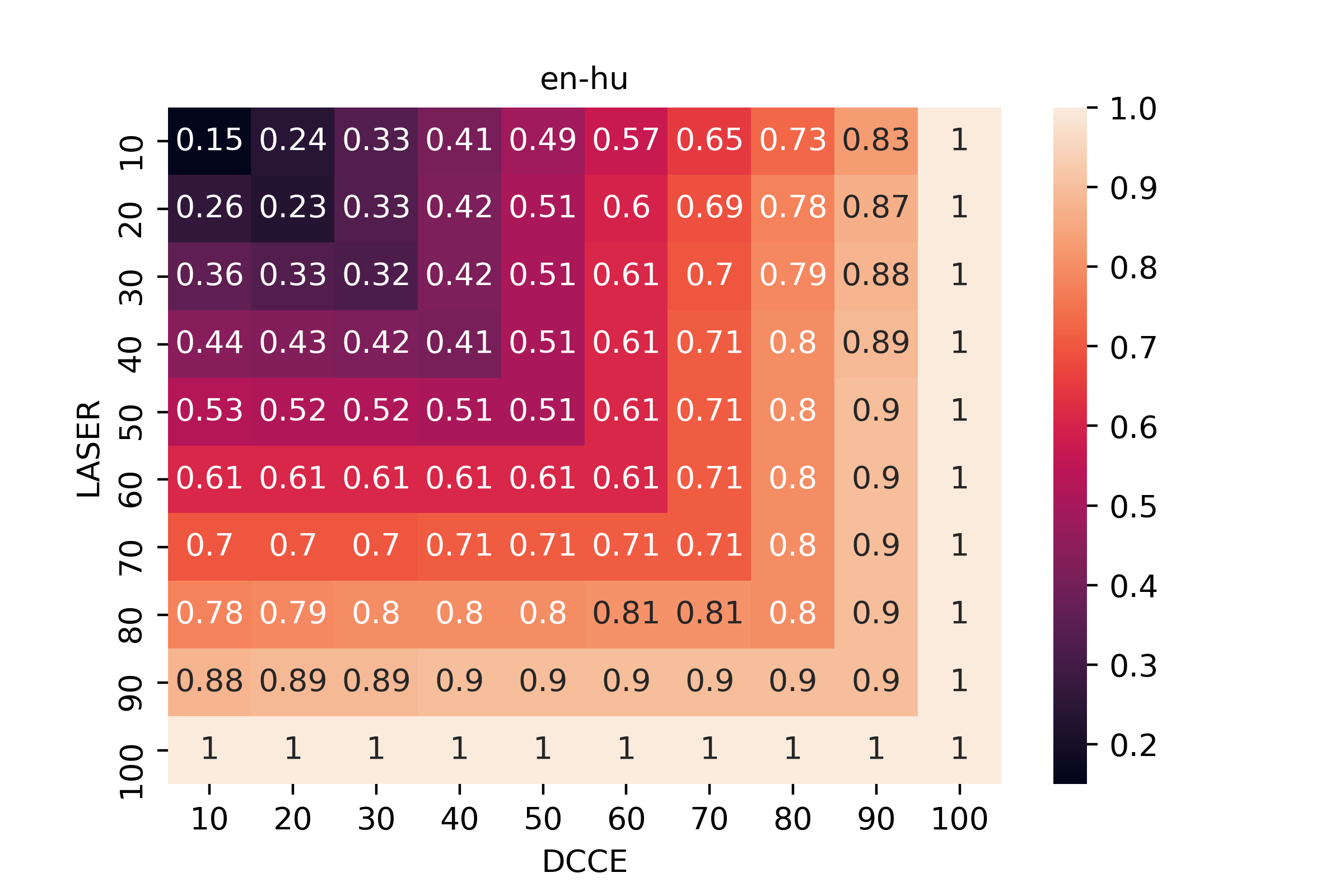}
\end{subfigure}%
\begin{subfigure}{.33\textwidth}
  \centering
  \includegraphics[width=1\linewidth]{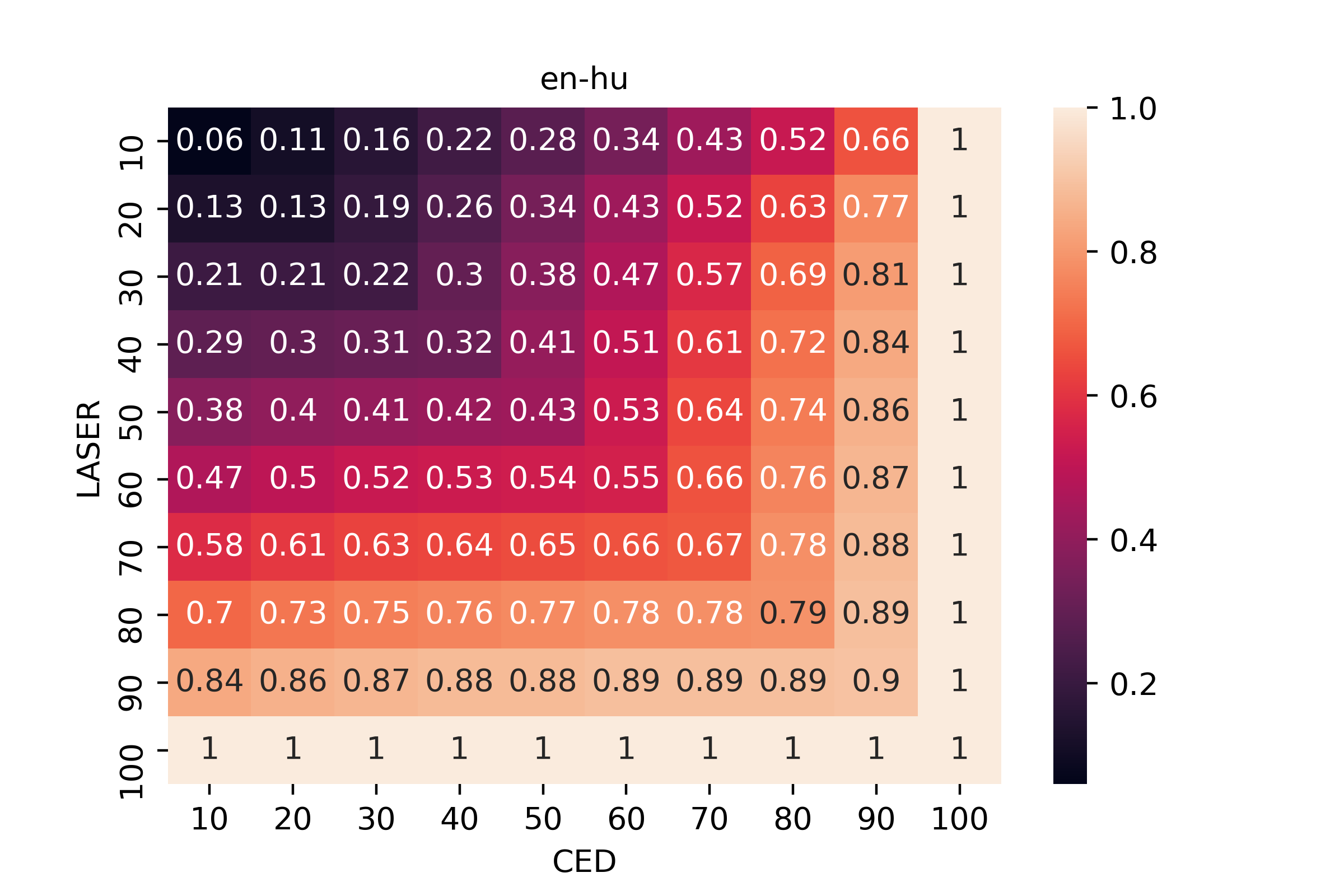}
\end{subfigure}
\begin{subfigure}{.33\textwidth}
  \centering
  \includegraphics[width=1\linewidth]{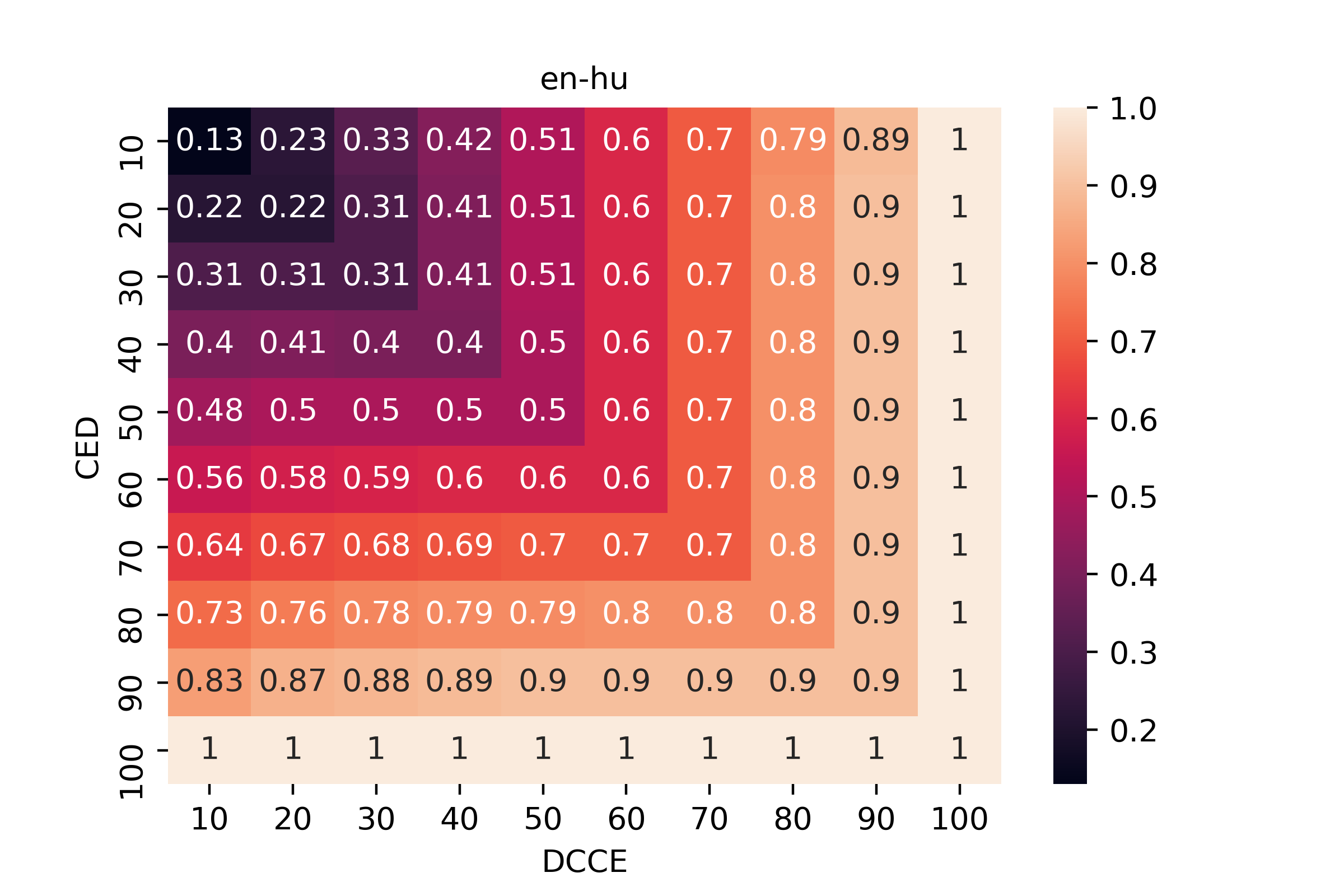}
\end{subfigure}

\begin{subfigure}{.33\textwidth}
  \centering
  \includegraphics[width=1\linewidth]{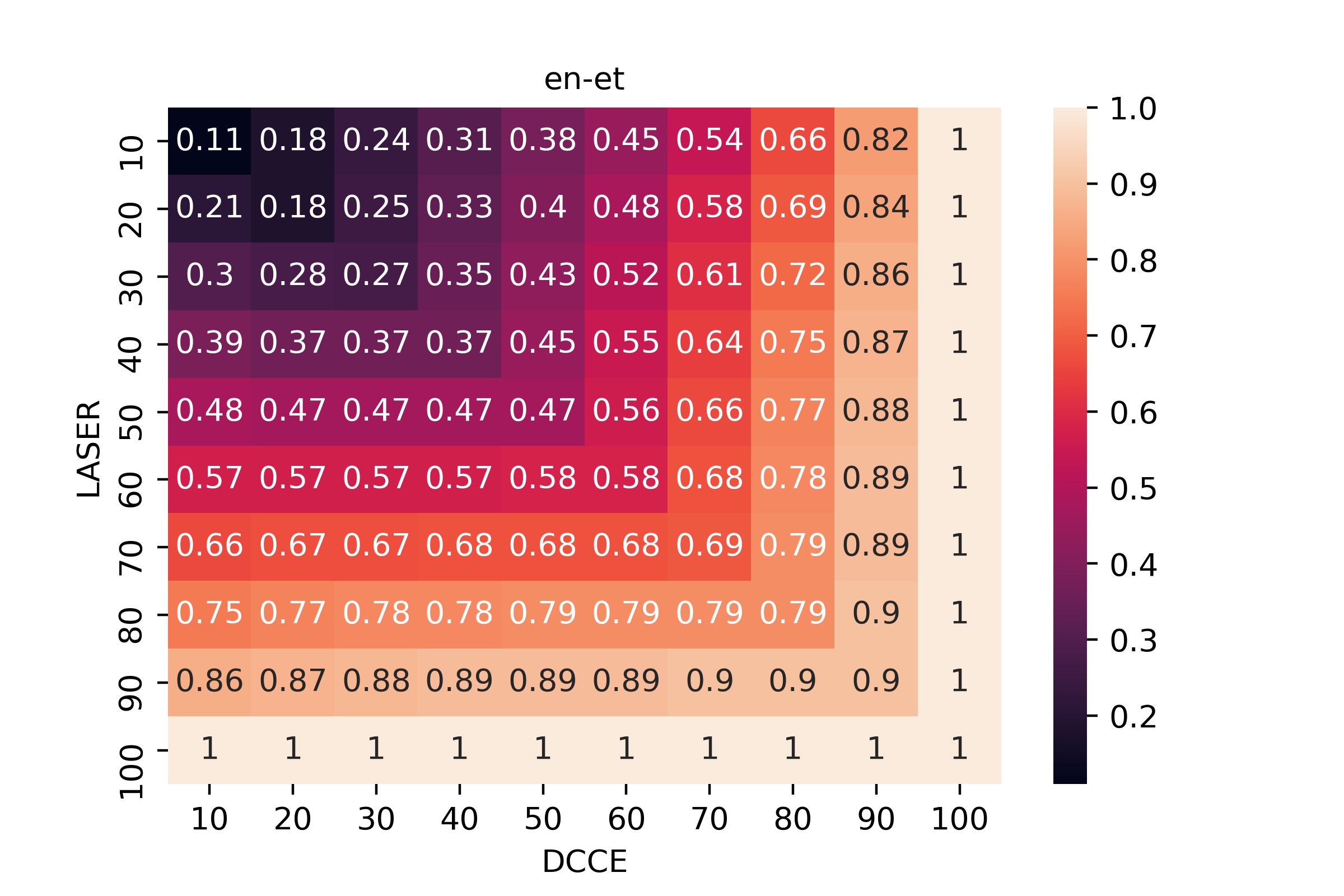}
\end{subfigure}%
\begin{subfigure}{.33\textwidth}
  \centering
  \includegraphics[width=1\linewidth]{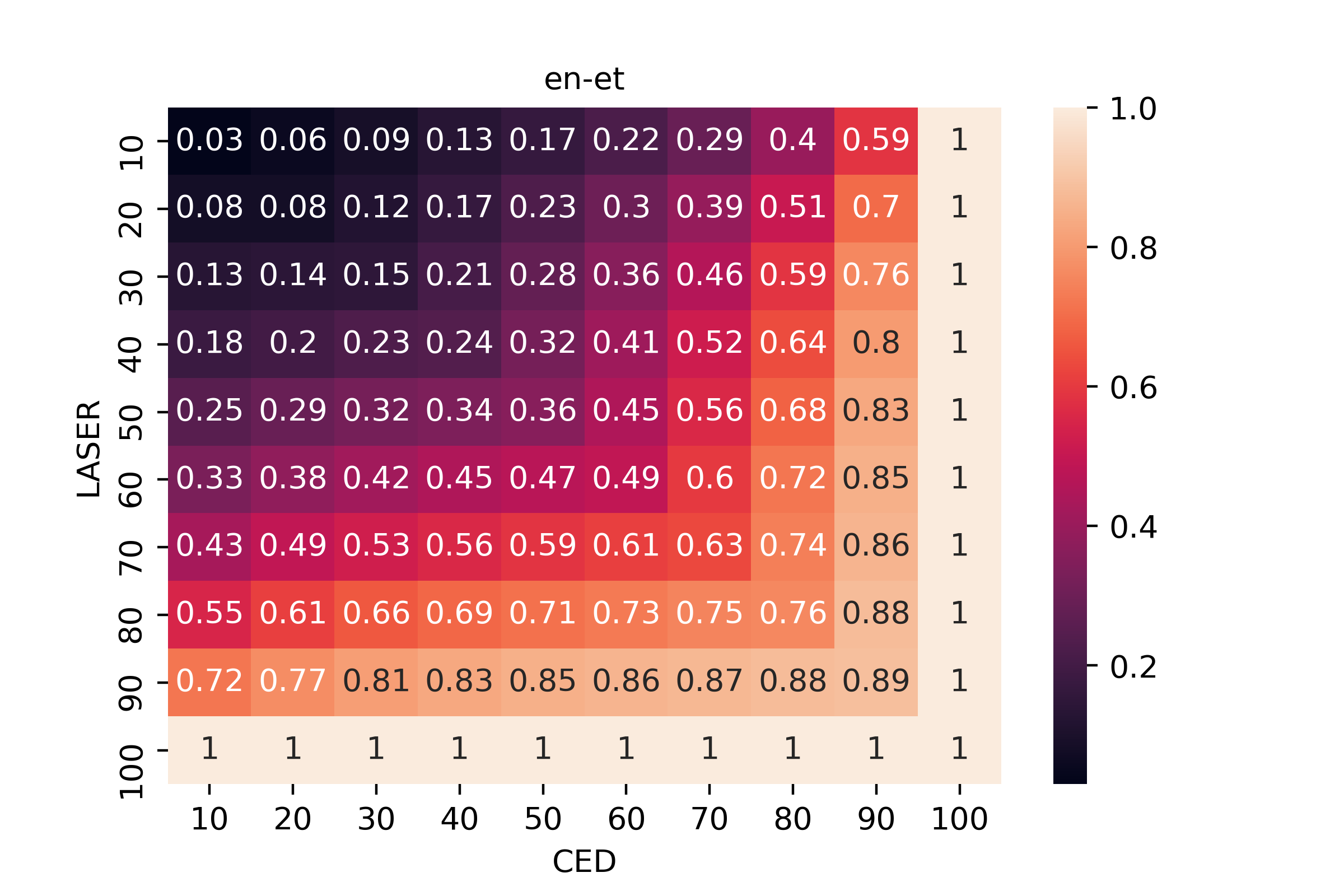}
\end{subfigure}
\begin{subfigure}{.33\textwidth}
  \centering
  \includegraphics[width=1\linewidth]{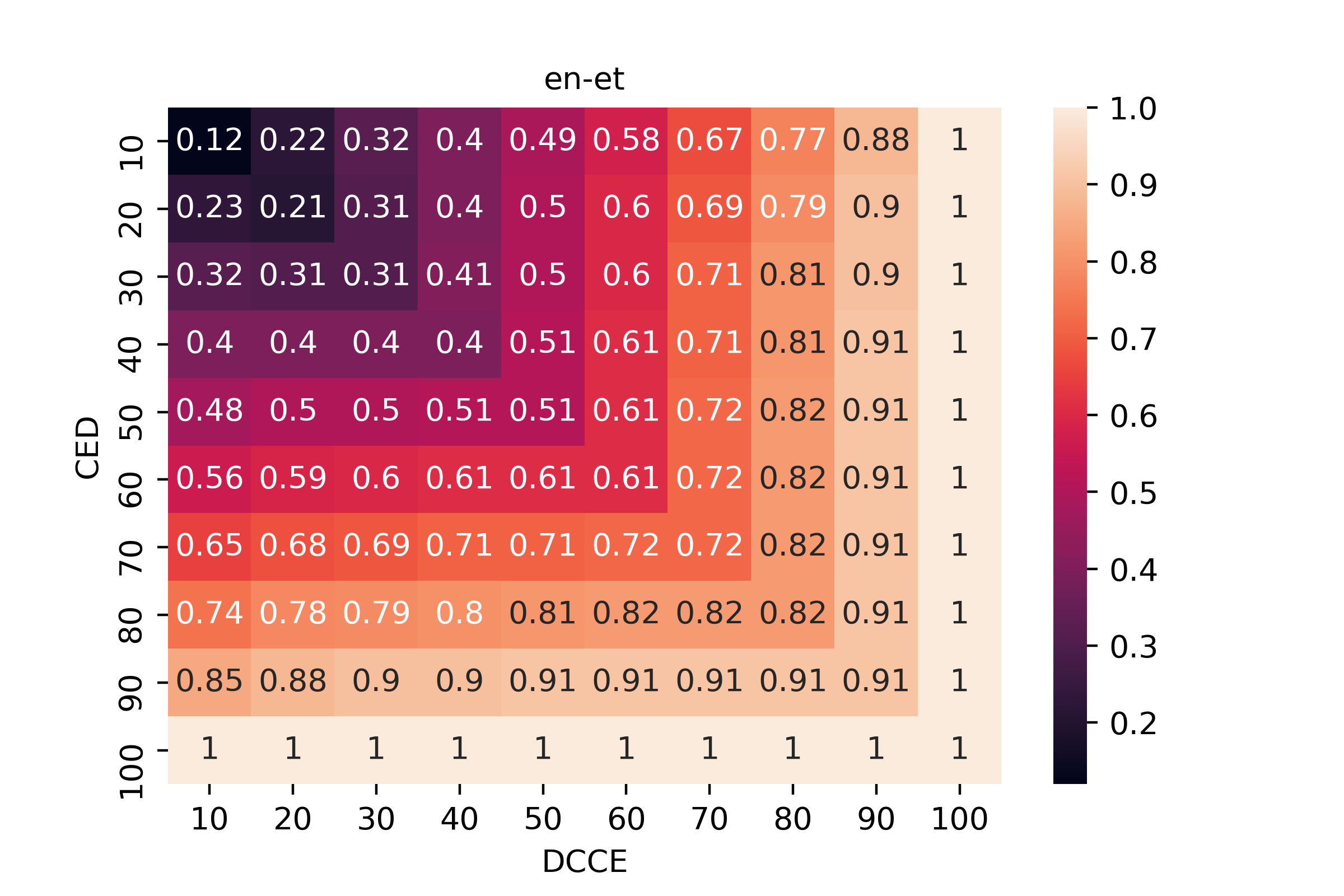}
\end{subfigure}

\caption{\textbf{Overlap percentage} of \textit{ranked data} between any two methods \{\text{LASER}, \text{DCCE}, \text{MML}\}.}
\label{fig:det-overlap}
\end{figure*}

\end{document}